\documentclass[11pt,fleqn,twoside]{article}

\usepackage[german,english]{babel}
\usepackage{isreport}

\usepackage{latexsym}
\usepackage{times}
\usepackage{theorem}
\usepackage{xspace}

\foreignreport

\newcommand{\nop}[1]{}

\newif\ifmakebbl

\newtheorem{definition}{Definition}
\newtheorem{theorem}{Theorem}
\newtheorem{proposition}{Proposition}
\newtheorem{corollary}{Corollary}
\newtheorem{lemma}{Lemma}
\newtheorem{example}{Example}

\newcounter{myenumctr}

\newenvironment{myitemize}{\begin{list}{$\bullet$}{\setlength{\leftmargin}{0pt}
\setlength{\itemindent}{\labelwidth}}}
{\end{list}}

\newcommand{\OR}{\vee}
\newcommand{\AND}{\wedge}

\newcommand{\card}[1]{\mathit{card}(#1)}
\newcommand{\atm}[1]{\mathit{Atm}(#1)}
\newcommand{\var}[1]{{\atm{#1}}}
\newcommand{\at}{{\mathcal A}}
\newcommand{\Lit}{{\mathit Lit}}
\newcommand{\extat}{\Lit_{\at}}

\newcommand{\DLP}{\mbox{DLP}}
\newcommand{\NLP}{\mbox{NLP}}
\newcommand{\head}[1]{H(#1)}

\newcommand{\body}[1]{B(#1)}
\newcommand{\bodyp}[1]{B^+(#1)}
\newcommand{\bodyn}[1]{B^-(#1)}
\newcommand{\la}{\leftarrow}
\newcommand{\ra}{\rightarrow}
 
\newcommand{\naf}{{\it not}\,}
\newcommand{\SM}{\mathcal{SM}}
\newcommand{\AS}{\mathcal{AS}}
\renewcommand{\SM}{\AS}
\newcommand{\SE}{\mathit{SE}}
\newcommand{\UE}{\mathit{UE}}
\newcommand{\SErel}[1]{\SE^{#1}}
\newcommand{\UErel}[1]{\UE^{#1}}
\newcommand{\SEA}{\SErel{A}}
\newcommand{\UEA}{\UErel{A}}
\newcommand{\equivu}{\equiv_{u}}
\newcommand{\equivs}{\equiv_{s}}
\newcommand{\equive}{\equiv_{e}}
\newcommand{\modelss}{\models_{s}}
\newcommand{\modelsu}{\models_{u}}
\newcommand{\modelse}{\models_{e}}

\newcommand{\Pol}{\ensuremath{\mathrm{P}}}

\newcommand{\NP}{\ensuremath{\mathrm{NP}}}

\newcommand{\coNP}{\ensuremath{\mathrm{coNP}}}
\newcommand{\CONP}{\ensuremath{\mathrm{coNP}}}
\newcommand{\SigmaP}[1]{{\Sigma}_{#1}^{P}}
\newcommand{\PiP}[1]{{\Pi}_{#1}^{P}}
\newcommand{\DeltaP}[1]{{\Delta}_{#1}^{P}}

\def\endproof{\ifhmode\nobreak\qed\par\fi\medskip}

\newcommand{\qed}{\hspace*{0mm}\hfill $\Box $\vspace{1.3mm}}

\newcommand{\iec}[0]{i.e.,\ }
\newcommand{\egc}[0]{e.g.,\ }
\newcommand{\commadots}[0]{,\ldots ,}

\title{Semantical Characterizations and Complexity of 
Equivalences in Answer Set Programming}

\authornames{
Thomas Eiter
\and 
Michael Fink
\and 
Stefan Woltran
}

\innertitle{
Semantical Characterizations and Computational Aspects of 
Equivalences in Stable Logic Programming}

\author{
Thomas Eiter\affiliation{
Institute of Information Systems, Knowledge-Based Systems Group,
TU Vienna, Favoritenstra\ss{}e\ 9-11,  A-1040 Vienna, Austria.
\mbox{E-mail: eiter@kr.tuwien.ac.at.}} 
and Michael Fink\affiliation{
Institute of Information Systems, Knowledge-Based Systems Group,
TU Vienna, Favoritenstra\ss{}e\ 9-11,  A-1040 Vienna, Austria.
\mbox{E-mail: michael@kr.tuwien.ac.at.}} 
and Stefan Woltran\affiliation{
Institute of Information Systems, Knowledge-Based Systems Group,
TU Vienna, Favoritenstra\ss{}e\ 9-11,  A-1040 Vienna, Austria.
\mbox{E-mail: stefan@kr.tuwien.ac.at.}} 
}

\published{This report extends INFSYS RR-1843-03-08.
The contents were partially published in
Proceedings 19th International Conference on Logic Programming 
(ICLP 2003), LNCS, Springer, 2003; and 
Proceedings 9th
European Conference on Logics in Artificial Intelligence (JELIA 2004), 
LNCS, Springer, 2004; see~\cite{eite-fink-03,Woltran04}.
}

\acknowledgement{This work was partially supported by the Austrian Science Fund (FWF) under 
project  
P18019-N04, as well as by the European Commission under
projects FET-2001-37004 WASP,
IST-2001-33570 COLOGNET,
and IST-2001-33570 INFOMIX.}

\abstract{ 
In recent research on non-monotonic logic programming, repeatedly
strong equivalence of logic programs $P$ and $Q$ has been considered,
which holds if the programs $P \cup R$ and $Q\cup R$ have the same 
answer sets for any other program $R$. This property strengthens
equivalence of $P$ and $Q$ with respect to 
answer sets (which is the
particular case for $R=\emptyset$), and has its applications in program
optimization, verification, and modular logic programming.
In this paper, we consider more liberal notions of
strong equivalence, 
in which the actual form of $R$ may be syntactically restricted.
On the one hand, we consider uniform equivalence, where $R$ 
is a set of facts rather than a set of rules. 
This notion, which is well known
in the area of deductive databases, is
particularly useful for assessing whether programs $P$ and $Q$ are
equivalent as components of a logic program which is modularly
structured. 
On the other hand, we consider relativized notions of equivalence,
where $R$ ranges over rules over a fixed alphabet, 
and thus generalize our results to
relativized notions of strong and uniform equivalence.
For all these notions, we 
consider disjunctive logic programs 
in the propositional (ground) case, 
as well as some restricted classes,
provide semantical characterizations and 
analyze the computational complexity.
Our results, which naturally extend to answer set semantics for
programs with strong negation, complement the results on strong
equivalence of logic programs and pave the way for optimizations in
answer set solvers as a tool for input-based problem solving.
\\*[\parskip]
{\bf Keywords:} \raisebox{.2cm}{\phantom{$|$}} 
answer set semantics, 
stable models, 
computational complexity, 
program optimization,
uniform equivalence, strong equivalence.
\\*[\parskip]
~\vspace*{-.65cm}}

\infsysrr{1843-05-01}
\date{February 2005}

\begin{document}

\maketitle

\newpage 
\pagenumbering{Roman}

\tableofcontents

\newpage
\pagenumbering{arabic}

\section{Introduction} 

In the last decade, the approach to reduce finding solutions of a
problem to finding ``models'' of a logical theory has gained
increasing importance as a declarative problem solving method. The
idea is that a problem at hand is encoded to a logical theory, such
that the models of this theory correspond to the solutions of the
problem, in a way such that from an arbitrary model of the theory, the
corresponding solution can be extracted efficiently. Given that the
mappings can be computed in polynomial time, this facilitates
polynomial time problem solving modulo the computation of a model of
the constructed logical theory, for which an efficient solver may be
used. An example of a fruitful application of this approach is
\cite{kaut-selm-92}, which showed that planning problems can be
competitively solved by encodings to the classical propositional
satisfiability problem (SAT) and running efficient SAT solvers.
Encodings of planning problems to nonclassical logics, in particular
to non-monotonic logic programs, have later been given in
\cite{subr-zani-96,dimo-etal-97,lifs-99,eite-etal-2001d}. Because of
the features of non-monotonic negation, such programs allow for a more
natural and succinct encoding of planning problems than classical
logic, and thus are attractive from a declarative point of view.

Given this potential, encoding problems to non-monotonic logic
programs under the 
answer set semantics~\cite{gelf-lifs-88,gelf-lifs-91}, which is now known as
\emph{Answer-Set Programming (ASP)} \cite{asp-2001}, has been
considered in the recent years for a broad range of other applications
including knowledge-base
updates~\cite{zhan-foo-98,inou-saka-99,alfe-etal-2000,eite-etal-01},
linguistics~\cite{erde-etal-03}, security requirements engineering
\cite{gior-etal-04}, or symbolic model checking~\cite{Heljanko01} as
well, to mention some of them. Many of these applications are realized
via dedicated languages (see, for instance, \cite{eite-etal-01e})
using ASP solvers as back-ends in which a specified reasoning task is
translated into a corresponding logic program.  Thus, an ever growing
number of programs is {\em automatically generated}, leaving the
burden of optimizations to the underlying ASP system.

Despite the high sophistication of current ASP-solvers like
\cite{simo-etal-2002,leon-etal-02-dlv,lin-zhao-2002,ange-etal-2001},
their current support for optimizing the programs is restricted in the
sense that optimizations are mainly geared towards on-the-fly model
generation. In an ad-hoc manner, program optimization aims at
simplifying an input program in a way such that the resulting program
has the same answer sets. This is heavily exploited in the systems
Smodels \cite{simo-etal-2002} and DLV \cite{leon-etal-02-dlv}, for
instance, when variables are eliminated from programs via grounding.

However, such optimization can only be applied to the entire program.
Local simplifications in parts of the program may not be correct at
the global level, since by the non-monotonicity of answer set 
semantics, adding the same rules to equivalent programs may lead to
programs with different models. This in particular hampers an offline
optimization of programs to which at run-time further rules are added,
which is important in different respects. Regarding code reuse, for
instance, a program may be used as a ``subprogram'' or ``expanded
macro'' within the context of another program (for example, to
nondeterministically choose an element from a set), and thus be
utilized in many applications. On the other hand, a problem encoding
in ASP usually consists of two parts: a generic problem specification
and instance-specific input (for example, 3-colorability of a graph in
general and a particular graph); here, an offline simplification of
the generic part is desirable, regardless of the concrete input at
run-time.

As pointed out by several
authors~\cite{Lifschitz01,eite-fink-03,osor-etal-01}, this calls for
stronger notions of equivalence.  As discussed below, there are
different ways to access this problem, depending on the actual context
of application and optimization. Accordingly, different notions of
equivalence may serve as a theoretical basis for optimization
procedures. In this paper, we present a first systematic and thorough
exploration of different notions of equivalence for answer set 
semantics with respect to semantical characterizations and
computational complexity. It provides a theoretical underpinning for
advanced methods of program optimization and for enhanced ASP
application development, as well as a potential basis for the
development of ASP debugging tools. In the following, we recall some
notions of equivalence that have been considered for answer set 
semantics, illustrated with some examples.

\paragraph{Notions of Equivalence.}
A notion of equivalence which is feasible for the issues discussed
above is {\em strong equivalence}~\cite{Lifschitz01,Turner01}: Two
logic programs $P_1$ and $P_2$ are strongly equivalent, if by adding
any set of rules $R$ to both $P_1$ and $P_2$, the resulting programs
$P_1 \cup R$ and $P_2 \cup R$ are equivalent under the answer set 
semantics, i.e., have the same answer sets. Thus, if a
program $P$ contains a subprogram $Q$ which is strongly equivalent to
a program $Q'$, then we may replace $Q$ by $Q'$, in particular if the
resulting program is simpler to evaluate than the original one.
\begin{example}\label{exa:hcf}
The programs $P_1 = \{ a \lor b \}$ and $Q_1 = \{ a \lor b;\; a \la \naf
b\}$ are strongly equivalent. Intuitively, the rule $a\la\naf b$ in
$Q$ is redundant since under 
answer set semantics, $a$ will be
derived from the disjunction $a\lor b$ if $b$ is false. On the other
hand, the programs $P_2 = \{ a \lor b \}$ and $Q_2 = \{ a \la \naf b;\; b
\la \naf a\}$ are not strongly equivalent: $P_2 \cup \{ a \la b;\; b
\la a\}$ has the  
answer set $\{a,b\}$, which is not an  
answer set of
$Q_2 \cup \{ a \la b;\; b \la a\}$.
\end{example}
Note that strong equivalence is, in general, suitable as a theoretical 
basis for local optimization. 
However,  
it is a very restrictive concept. 
There are two fundamental options to weaken it and obtain less restrictive 
notions. On the one hand, one can restrict the syntax of possible program 
extensions $R$, or one can restrict the set of atoms occurring in $R$.

The first approach leads us to 
to the well known
notion of {\em uniform
equivalence}~\cite{sagi-88,mahe-88}. 
Two logic programs $P_1$ and $P_2$ are uniformly equivalent,
if by adding any set of {\em facts} $F$ to both $P_1$ and $P_2$, the
resulting programs $P_1 \cup F$ and $P_2 \cup F$ have the same set of
answer sets.
That strong equivalence and uniform equivalence are different
concepts is illustrated by the following simple example.  
\begin{example}\label{exa:excl}
It can be checked that the programs $P_2$ and $Q_2$ from
Example~\ref{exa:hcf}, while not strongly equivalent, are uniformly
equivalent. We note that by adding the constraint $~\la a,b$ to them,
the resulting programs $P_3 = \{ a \lor b;\; \la a,b\}$ and $Q_3 = \{
a \la \naf b;\; b \la \naf a;\; \la a,b \}$, which both express exclusive
disjunction of $a$ and $b$, are strongly equivalent (and hence also uniformly
equivalent).
\end{example}

This example may suggest that disjunction is an essential feature
to make a difference between strong and uniform equivalence. 
In fact this is not the case, as shown by the following example.

\begin{example}\label{ex:1}
Let $P_4=\{a\la \naf b ;\  a\la b\}$ and 
$Q_4=\{a\la \naf c ;\  a\la c\}$. Then, it is easily verified that 
$P_4$ and $Q_4$ are uniformly equivalent. However, they are not strongly 
equivalent: For $P_4 \cup\{b\la a\}$ and  $Q_4 \cup\{b\la a\}$, we have
that $S=\{ a, b\}$  is a  
answer set of $Q_4 \cup\{b\la a\}$ but not of 
$P_4 \cup\{b\la a\}$. 
\end{example}

As for program optimization, compared to strong equivalence, uniform
equivalence is more sensitive to a modular structure of logic programs
which naturally emerges by splitting them into layered {\em
components} that receive input from lower layers by facts and in turn
may output facts to a higher
layer~\cite{lifs-turn-94,eite-etal-97f}. In particular, the applies to
the typical ASP setting outlined above, in which a generic problem
specification component receives problem-specific input as a set of
facts.

However, as mentioned before, a 
different way to obtain weaker equivalence notions than strong 
equivalence is to restrict the alphabet of possible program extensions.
This is of particular interest, whenever
one wants to \emph{exclude} dedicated atoms from  
program extensions. 
Such atoms may play the role 
of internal atoms in  
program  
components  
and are 
considered not to appear anywhere else in the complete program $P$.
This notion
of equivalence 
was originally suggested by Lin in~\cite{Lin02} but not further investigated. 
We will formally define 
\emph{strong equivalence relative to a given set of atoms $A$}
of two programs $P$ and~$Q$
as the test whether, for all sets of rules $S$ over a given set of atoms $A$, 
$P\cup S$ and $Q\cup S$ have the same  
answer sets.

Finally, we introduce the notion of \emph{uniform equivalence relative
to a given set of atoms $A$}, as the property that for two programs
$P$ and~$Q$ and for all sets $F\subseteq A$ of facts, $P\cup F$ and
$Q\cup F$ have the same   
answer sets. Note that relativized uniform
equivalence generalizes the notion of equivalence of DATALOG programs
in deductive databases~\cite{Shmueli93}. There, DATALOG
programs are called equivalent, if it holds that they compute
the same outputs on any set of external atoms (which are atoms that do
not occur in any rule head) given as input. The next example
illustrates that relativization weakens corresponding notions of
equivalence.

\begin{example}\label{exa:rel}
Let $P_5 = \{ a \lor b \}$ and $Q_5 = \{ a \la \naf b;\; b \la \naf
a;\; c \la a,b;\; \la c\}$.  The programs $P_5$ and $Q_5$ have the
same  
answer set, but are neither uniformly equivalent nor strongly
equivalent.  In particular, it is sufficient to add the fact
$c$. Then, $P_5\cup \{c\}$ has $\{a,b,c\}$ as an answer set,  
while
$Q_5\cup\{c\}$ has no   
answer set. 
However, if we exclude $c$ from
the alphabet of possible program extensions, uniform equivalence
holds. More specifically, $P$ and~$Q$ are uniformly equivalent
relative to for any set of atoms $A$ such that $c\notin A$. On the
other hand, $P$ and $Q$ are not strongly equivalent relative to any
$A$ which includes both $a$ and $b$.  The reason is that adding $a\la
b$ and $b\la a$ leads to different 
answer sets (cf.\
Example~\ref{exa:hcf}).
\end{example}

\paragraph{Main Contributions.} 
In this paper, we study semantical and complexity properties of the
above notions of equivalence, where we focus on the propositional case
(to which first-order logic programs reduce by instantiation).  Our
main contributions are briefly summarized as follows.

\begin{myitemize}
 \item We provide characterizations of uniform equivalence of logic
 programs.  To this aim, we build on the concept of {\em
 strong-equivalence models (SE-models)}, which have been introduced
 for characterizing strong equivalence \cite{Turner01,Turner03} in
 logic programming terms, resembling an earlier characterization of
 strong equivalence in terms of equilibrium logic which builds on the
 intuitionistic logic of here and there \cite{Lifschitz01}. A strong
 equivalence model of a program $P$ is a pair $(X,Y)$ of (Herbrand)
 interpretations such that $X\subseteq Y$, $Y$ is a classical model of
 $P$, and $X$ is a model of the Gelfond-Lifschitz reduct $P^Y$ of $P$
with respect to $Y$ \cite{gelf-lifs-88,gelf-lifs-91}. Our
characterizations of uniform equivalence will elucidate the
differences between strong and uniform equivalence, as illustrated in
the examples above, such that they immediately become apparent.

\item For the finitary case, we provide a mathematical simple and
appealing characterization of a logic program with respect to uniform
equivalence in terms of its {\em uniform equivalence models
(UE-models)}, which is a special class of SE-models. Informally, those
SE-models $(X,Y)$ of a program $P$ are UE-models, such that either $X$
equals $Y$ or is a maximal proper subset of $Y$.  On the other hand,
we show that uniform equivalence of infinite programs cannot be
captured by any class of SE-models in general.  Furthermore, the
notion of logical consequence from UE-models, $P\modelsu Q$, turns out
to be interesting since programs $P$ and $Q$ are uniformly equivalent
if and only if $P\modelsu Q$ and $Q\modelsu P$ holds. Therefore,
logical consequence (relative to UE-models) can be fruitfully used to
determine redundancies under uniform equivalence.

\item By suitably generalizing the characterizations of strong and
  uniform equivalence, and in particular SE-models and UE-models, we
  also provide suitable {\em semantical characterizations} for both
  relativized strong and uniform equivalence.  Our new
  characterizations thus capture all considered notions of equivalence
  (including ordinary equivalence) in a uniform way.  Moreover, we
  show that relativized strong equivalence shares an important
  property with strong equivalence: constraining possible
  program extensions to sets of rules of the form $A\la B$ , where $A$
  and $B$ are atoms, does not lead to a different concept (Corollary~\ref{cor:unary}). The
  observation of Pearce and Valverde~\cite{Pearce04} that uniform and
  strong equivalence are essentially the only concepts of equivalence
  obtained by varying the \emph{logical} form of the program
  extensions therefore generalizes to relative equivalence.

 \item Besides the general case, we consider various major syntactic
subclasses of programs, in particular Horn programs, positive
programs, disjunction-free programs, and head-cycle free
programs~\cite{bene-dech-94}, and consider how these notions of
equivalence relate among each other.  For instance, we establish that
for positive programs, all these notions coincide, and therefore only
the classical models of the programs have to be taken into account for
equivalence testing. Interestingly, for head-cycle free programs,
eliminating disjunctions by shifting atoms from rule heads to the
respective rule bodies preserves (relativized) uniform equivalence,
while it affects (relativized) strong equivalence in general.

\item We thoroughly analyze the computational complexity of deciding
(relativized) uniform equivalence and relativized strong equivalence,
as well as the complexity of model checking for the corresponding
model-theoretic characterizations. We show that deciding uniform
equivalence of programs $P$ and $Q$ is $\PiP{2}$-complete in the
general propositional case, and thus harder than deciding strong
equivalence of $P$ and $Q$, which is $\coNP$-complete
\cite{Pearce01,Lin02,Turner03}. The relativized notions of equivalence
have the same complexity as uniform equivalence in general
($\PiP{2}$-completeness). These results reflect the intuitive
complexity of equivalence checking using the characterizations we
provide. Furthermore, we consider the problems for subclasses and
establish $\coNP$-completeness results for important fragments,
including positive and head-cycle free programs, and thus obtain a
complete picture of the complexity-landscape, which is summarized in
Table~\ref{tab:newR}.
Some of the results obtained
are surprising; for example, checking relativized uniform equivalence
of head-cycle free programs, is \emph{easier} than deciding
relativized strong equivalence. For an overview and discussion of the
complexity results, we refer to Section~\ref{sec:complexity}.

\item 
Finally, we address extensions of our results w.r.t.\
modifications in the language of propositional programs, viz.\
addition of strong negation or nested expressions, as well as 
disallowing constraints. Moreover, we briefly 
discuss the general DATALOG-case.
\end{myitemize}

Our results extend recent results on strong equivalence of logic
programs, and pave the way for optimization of logic programs under  
answer set semantics by exploiting either strong equivalence, uniform
equivalence, or relativized notions thereof.

\paragraph{Related Work.} 

While strong equivalence of logic programs under  
answer set semantics has
been considered in a number of papers
\cite{Cabalar02,dejo-hend-03,Lin02,Lifschitz01,osor-etal-01,Pearce01,Turner01,Turner03,Pearce04b,Pearce04c},
investigations on uniform equivalence just started with preliminary
parts of this work~\cite{eite-fink-03}.  Recent papers on program
transformations~\cite{Eiter03a,Eiter04a} already take both notions
into account.  In the case of DATALOG, uniform equivalence is a
well-known concept, however.  Sagiv~\cite{sagi-88}, who coined the
name, has studied the property in the context of definite Horn DATALOG
programs, where he showed decidability of uniform equivalence testing,
which contrasts the undecidability of equivalence testing for DATALOG
programs~\cite{Shmueli93}.  
Also Maher~\cite{mahe-88} considered uniform equivalence
for definite general Horn programs (with function symbols), and
reported undecidability.  Moreover, both~\cite{sagi-88,mahe-88} showed
that uniform equivalence coincides for the respective programs with
Herbrand logical equivalence. Maher also pointed out that for DATALOG
programs, this result has been independently established by Cosmadakis
and Kanellakis \cite{cosm-kane-86}.  Finally, a general notion of
equivalence has also been introduced by Inoue and
Sakama~\cite{Inoue04}. In their framework, called \emph{update
equivalence}, one can exactly specify a set of arbitrary rules which
may be added to the programs under consideration and, furthermore, a
set of rules which may be deleted.  However, for such an explicit
enumeration of rules for program extension, respectively modification,
it seems to be much more complicated to obtain simple semantical
characterizations.

The mentioned papers on strong equivalence
mostly concern logical characterizations. 
In particular, the seminal work by Lifschitz {\em et\ al.}~\cite{Lifschitz01}
showed that strong equivalence corresponds to equivalence in the non-classical
logic of here-and-there. 
De Jongh and Hendriks~\cite{dejo-hend-03} generalized 
this result
by showing
that strong equivalence is characterized
by equivalence in all intermediate logics lying between
here-and-there (upper bound) and the logic KC of weak excluded
middle~\cite{Kowalski68} (lower
bound) which is axiomatized by intuitionistic logic together with
the schema $\neg \varphi \vee \neg \neg \varphi$.
In addition, \cite{Cabalar02} presents another multi-valued
logic known as $L_3$ which can be 
employed to decide strong equivalence in the same
manner.
However, the most popular semantical characterization was introduced
by Turner~\cite{Turner01,Turner03}. He abstracts from the
Kripke-semantics
as used in the logic of here-and-there, resulting in the above
mentioned
\emph{SE-models}.
Approaches to 
implement strong equivalence can be found 
in~\cite{Eiter03a,Janhunen04,Pearce01}.  
Complexity characterizations of strong equivalence 
were given by several authors~\cite{Pearce01,Lin02,Turner03}.
Our work refines and generalizes this work by 
considering (relativized) strong equivalence also for syntactic fragments,
which previous work did not pay much attention to.
As well, we present a new syntactical 
criterion to retain strong equivalence when transforming  
head-cycle free programs to 
disjunction-free ones, complementing 
work on program 
transformations~\cite{Eiter04a,Eiter03a,osor-etal-01,Pearce04}.
The recent work by 
Pearce and Valverde~\cite{Pearce04}
addresses strong equivalence of programs
over disjoint alphabets which are synonymous under structurally
defined mappings.

\paragraph{Structure of the paper.} 
The remainder of this paper is organized as follows. The next section
recalls important concepts and fixes notation. After that, in
Section~\ref{sec:char}, we present our characterizations of uniform
equivalence.  We also introduce the notions of UE-model and
UE-consequence and relate the latter to other notions of consequence.
Then, Section~\ref{sec:rel} introduces the relativized notions of
equivalence, and we present our generalized characterizations in
model-theoretic terms.  Section~\ref{sec:classes} considers two
important classes of programs, in particular positive and head-cycle
free logic programs, which include Horn and normal logic programs,
respectively. The subsequent Section~\ref{sec:complexity} is devoted
to a detailed analysis of complexity issues, while
Section~\ref{sec:extensions} considers possible extensions of our
results to nested logic programs and answer set semantics for programs
with strong negation (also allowing for inconsistent answer sets), as
well as to DATALOG programs.  The final Section~\ref{sec:conclusion}
concludes the paper and outlines issues for further research.

\section{Preliminaries}
\label{sec:prelim}

We deal with disjunctive logic programs, which allow the use of default 
negation $\naf$ in rules. A rule $r$ is a triple 
$\langle \head{r},\bodyp{r},\bodyn{r}\rangle$, where 
$\head{r}=\{A_1\commadots A_l\}$, $\bodyp{r}=\{A_{l+1}\commadots A_m\}$, 
$\bodyn{r}=\{A_{m+1}\commadots A_n\}$, where $0\leq l \leq m \leq n$ and $A_i$, $1\leq i\leq n$, 
are atoms from a first-order language. 
Throughout, we use the traditional representation of a rule as an expression 
of the form 
\medskip

\centerline{$A_1\vee \dots \vee A_l \la A_{l+1},\dots,A_m,\naf
  A_{m+1},\dots,\naf A_n$.}

\medskip

We call $\head{r}$ the \emph{head} of $r$, 
and $\body{r}=\{A_{l+1}\commadots A_m,\naf A_{m+1},$ $\ldots, \naf A_n\}$ the 
\emph{body} of $r$. 
If $\head{r}=\emptyset$, then $r$ is a \emph{constraint}.
As usual, 
$r$ is a \emph{disjunctive fact} if $\body{r}=\emptyset$, 
and $r$ is a (non-disjunctive) \emph{fact} if $\body{r}=\emptyset$ and 
$l=1$,
both also represented by 
$\head{r}$ if it is nonempty, and by $\bot$ (falsity) otherwise.
A rule $r$ is \emph{normal} (or non-disjunctive), if $l\leq1$;
\emph{definite}, if $l=1$; and \emph{positive}, if $n=m$. A rule 
is \emph{Horn} if it is normal and positive. A definite Horn rule is called
\emph{unary} iff its body contains at most one atom. 

A \emph{disjunctive logic program} ($\DLP$) $P$ is a (possibly
infinite) set of rules. A program $P$ is a \emph{normal logic program} ($\NLP$)
(resp., definite, positive, Horn, or unary), if all rules in $P$
are normal (resp., definite, positive, Horn, unary).  Furthermore, a
program $P$ is \emph{head-cycle free $($HCF$)$}~\cite{bene-dech-94},
if each each $r\in P$ is head-cycle free (in $P$), \iec
if the dependency graph of $P$ (which is defined as usual)
where literals of form $\naf A$ are disregarded, has no directed
cycle that contains two atoms belonging to $H(r)$. 

{In the rest of this paper, we focus on propositional
programs over a set of atoms $\at$ -- programs with variables reduce to their ground
(propositional) versions as usual.}
The set of all atoms occurring in a program $P$ is denoted by $\var{P}$.

We shall deal with further variations of the syntax, 
where either \emph{strong} negation is available or constraints are 
disallowed in Section~\ref{sec:extensions}. There we shall also briefly discuss
how to apply our results to programs with nested expressions~\cite{Lifschitz99}
or to non-ground programs directly.

We recall the answer set semantics for $\DLP$s~\cite{gelf-lifs-91}, 
which generalizes the answer set 
semantics for $\NLP$s~\cite{gelf-lifs-88}. An \emph{interpretation}
$I$, viewed as subset of $\at$, models the head of a rule $r$, denoted
$I\models \head{r}$, iff $A\in I$ for some $A\in \head{r}$. It models
$\body{r}$, \iec $I\models \body{r}$ iff ($i$) each $A\in\bodyp{r}$ is
true in $I$, \iec $A\in I$, and ($ii$) each $A\in\bodyn{r}$ is false
in $I$, \iec $A\not\in I$. Furthermore, $I$ models rule $r$, \iec $I\models r$ 
iff
$I\models \head{r}$ whenever $I\models\body{r}$, and
$I$ is a model of a program $P$, denoted $I\models P$, iff $I\models r$,
for all $r\in P$. If $I\models P$ (resp.\ $I\models r)$, $I$ is called
a \emph{model} of $P$ (resp.\ $r$).

The {\em reduct} of a rule $r$ {\em relative to\/}
a set of atoms $I$, denoted $r^I$, is the positive rule $r'$ such that 
$\head{r'}= \head{r}$ and $\bodyp{r'}=\bodyp{r}$ if $I\cap
\bodyn{r} = \emptyset$; otherwise $r^I$ is void. 
Note that a void rule has any interpretation as its model. 
The {\em Gelfond-Lifschitz reduct} $P^I$, of a program $P$ 
is $P^I=\{ r^I \mid r\in P \}$.
An interpretation $I$ is an \emph{answer set} 
(or a \emph{stable model}~\cite{przy-91})   of a program $P$
iff $I$ is a minimal model (under inclusion $\subseteq$) of
$P^I$. By $\SM(P)$ we denote the set of all answer sets of $P$.

Several notions for equivalence of logic programs 
have been considered, cf.\ \cite{Lifschitz01,mahe-88,sagi-88}. 
In answer set programming, two $\DLP$s $P$ and $Q$ are regarded as equivalent,
denoted $P\equiv Q$, iff $\SM(P)=\SM(Q)$. 

The more restrictive form of strong equivalence~\cite{Lifschitz01} 
is as follows.
\begin{definition}
Let $P$ and $Q$ be two $\DLP$s. Then,
$P$ and $Q$ are strongly equivalent, denoted $P\equivs Q$, 
iff for any rule set $R$, the programs $P\cup R$ and $Q\cup R$ are equivalent, 
\iec $P\cup R\equiv Q\cup R$. 
\end{definition}

One of the main results of~\cite{Lifschitz01} is a semantical
characterization of strong equivalence in terms of the non-classical
logic HT. For characterizing strong equivalence in
logic programming terms, Turner introduced the
following notion of SE-models~\cite{Turner01,Turner03}:

\begin{definition}
Let $P$ be a $\DLP$, 
and let $X, Y$ be sets of atoms such that $X\subseteq Y$. The pair $(X,Y)$ 
is an SE-model of $P$, if $Y\models P$ and 
$X\models P^Y$. By $\SE(P)$ we denote the set of all SE-models of
                $P$. For a single rule $r$, we write $\SE(r)$ instead
of $\SE(\{r\})$. 
\end{definition}

Strong equivalence can be characterized as follows.

\begin{proposition}[\cite{Turner01,Turner03}]\label{prop:turner}
For every $\DLP$s $P$ and $Q$, $P \equivs Q$ iff $\SE(P) = \SE(Q)$.
\end{proposition}

To check strong equivalence of two programs $P$ and $Q$, 
it is obviously sufficient to consider SE-interpretations $(X,Y)$ over
$\var{P\cup Q}$, \iec
with
$X\subseteq Y\subseteq \var{P\cup Q}$. We
implicitly
make use of this
simplification when convenient.

\begin{example}
Reconsider the examples from the introduction. First take programs
$P=\{a\OR b\}$ and $Q=\{a\la \naf b;\; b\la \naf{a}\}$.  We
have\footnote{To ease notation, we write $abc$ instead of $\{a,b,c\}$,
$a$ instead of $\{a\}$, etc.}
\begin{eqnarray*}
\SE(P) & = & \{ (a,a);\; (b,b);\; (a,ab);\; (b,ab);\; (ab,ab)\}; \\
\SE(Q) & = & \{ (\emptyset,ab);\; (a,a);\; (b,b);\; (a,ab);\; (b,ab);\; (ab,ab)\}.
\end{eqnarray*}
Thus, $(\emptyset,ab)$ is SE-model of $Q$ but not of $P$. This is 
due to the fact that $P^{\{a,b\}}=\{a\OR b\}$ and $Q^{\{a,b\}}$ is the empty program.
The latter is modelled by the empty interpretation, while the former is not. 
Hence, we derive $P\not\equivs Q$. 
\end{example}

\begin{example}
For the second example,
$P=\{a\la \naf b ;\  a\la b\}$ and 
$Q=\{a\la \naf c ;\  a\la c\}$,  
we also get
$P \not\equivs Q$. 
In this case, we have:
\begin{eqnarray*}
\SE(P) & = & \{ (\emptyset,ab);\; (\emptyset,abc);\; (c,abc)\} \cup S; \\
\SE(Q) & = & \{ (\emptyset,ac);\; (\emptyset,abc);\; (b,abc)\} \cup S; 
\end{eqnarray*}
with $S= \{ (X,Y) \mid \{a\}\subseteq X\subseteq Y\subseteq \{a,b,c\}\}$.
This shows $P\not\equivs Q$.
\end{example}

Note that from the proofs of the results in \cite{Lifschitz01,Turner03}, it appears
that for strong equivalence, only the addition of 
unary rules is crucial. That is, by constraining the rules in
the set $R$ in the definition of strong equivalence to 
normal rules having at most 
one positive atom in the body does not lead to a different concept. 
This
is encountered by restriction to facts (i.e., empty rule bodies), however.

As well, 
answer sets of a program can be characterized via its SE-models as follows:

\begin{proposition}\label{prop:sm}
For any $\DLP$ $P$, $Y\in \SM(P)$ iff $(Y,Y)\in\SE(P)$ and $(X,Y)\in\SE(P)$ implies $X=Y$, for any $X$.
\end{proposition}

Finally, we define a consequence relation associated to SE-models. 

\begin{definition}\label{def:secons}
Let $P$ be a $\DLP$ and $r$ a rule.
Then, $r$ is a SE-consequence of $P$, denoted $P\modelss r$, iff for each $(X,Y)\in\SE(P)$, 
it holds that $(X,Y)\in\SE(r)$.
Furthermore,
we write
$P \modelss Q$
iff
$P \modelss r$, for every $r\in Q$.
\end{definition}
\begin{proposition}
\label{prop:equive-modelse}
For any $\DLP$ $P$ and $Q$, $P \equivs Q$ iff $P \modelss Q$ and $Q \modelss  P$. 
\end{proposition}

Thus, the notion of SE-consequence captures strong equivalence of
logic programs. 

\section{Uniform Equivalence} 
\label{sec:char}

After the preliminary definitions, 
we now turn to the issue of uniform equivalence of logic programs.
We follow the definitions of 
uniform equivalence in~\cite{sagi-88,mahe-88}. 
\begin{definition}
Let $P$ and $Q$ be two $\DLP$s. Then,
$P$ and $Q$ are \emph{uniformly equivalent}, denoted $P\equivu Q$, iff for any set of $($non-disjunctive$)$
facts $F$, the programs $P\cup F$ and $Q \cup F$ are equivalent, 
\iec $P\cup F\equiv Q\cup F$.
\end{definition}

\subsection{A Characterization for Uniform Equivalence}

We proceed by 
characterizing uniform equivalence of logic programs in
model-theoretic terms. As restated above, strong equivalence can be
captured by the notion of SE-model (equivalently, HT-model~\cite{Lifschitz01}) 
for a logic program. The weaker notion of uniform
equivalence can be characterized in terms of SE-models as well, by
imposing further conditions.

We start with a seminal lemma, which allows us to derive simple
characterizations of uniform equivalence.

\begin{lemma}\label{theo-equivu-1} 
Two $\DLP$s $P$ and $Q$ are uniformly equivalent, i.e.\ $P\equivu Q$, iff 
for every SE-model $(X,Y)$, such that $(X,Y)$ is an SE-model of exactly one 
of the programs $P$ and $Q$,  it holds that ($i$) $Y\models P\cup Q$, and  
($ii$) there exists an SE-model $(X',Y)$, $X\subset X'\subset Y$, of the other 
program.
\end{lemma}

\begin{proof}
For the only-if direction, suppose $P\equivu Q$. If $Y$ neither models
$P$, nor $Q$, then $(X,Y)$ is not an SE-model of any of the programs
$P$ and $Q$.  Without loss of generality, assume $Y\models P$ and
$Y\not\models Q$. Then, since in this case $Y\models P^Y$ and no
strict subset of $Y$ models $P\cup Y$, $Y\in\SM(P\cup Y)$, while 
$Y\not\in\SM(Q\cup Y)$. This contradicts
our assumption $P\equivu Q$.  Hence,  
($i$) must hold.

To show ($ii$), assume first that $(X,Y)$ is an SE-model of $P$ but
not of $Q$.  In view of ($i$), it is clear that $X\subset Y$ must
hold. Suppose now that for every set $X'$, $X\subset X'\subset Y$, it
holds that $(X',Y)$ is not an SE-model of $Q$. Then, since no subset of
$X$ models $Q^Y\cup X$, $(Y,Y)$ is the only SE-model of $Q\cup X$ of
form $(\cdot,Y)$. Thus, $Y\in\SM(Q\cup X)$ in this
case, while $Y\not\in\SM(P\cup X)$ ($X\models P^Y$
implies $X\models (P\cup X)^Y$, so $(X,Y)$ is an SE-model of $P\cup
X$).  However, this contradicts $P\equivu Q$. Thus, it follows that
for some $X'$ such that $X \subset X' \subset Y$, $(X,Y)$ is an SE-model of $Q$.
The argument in the case where $(X,Y)$ is an SE-model of $Q$ but
not of $P$ is analogous. This proves  
($ii$).

For the if direction, assume that ($i$) and ($ii$) hold for every
SE-model $(X,Y)$ which is an SE-model of exactly one of $P$ and $Q$.
Suppose that there exist sets of atoms $F$ and $X$, such that w.l.o.g., 
$X\in\SM(P\cup F)\setminus\SM(Q\cup F)$. Since $X\in\SM(P\cup F)$, we have 
that $F\subseteq X$, and, moreover, 
$X\models P$. Consequently, $(X,X)$ is an SE-model of $P$.  Since 
$X\not\in\SM(Q\cup F)$, either $X\not\models (Q\cup F)^X$,
or there exists $Z\subset X$ such that $Z\models
(Q\cup F)^X$.

Let us first assume $X\not\models (Q\cup F)^X$. Then, since 
$(Q\cup F)^X = Q^X\cup F$ and $F\subseteq X$, it follows that 
$X\not\models Q^X$. This implies $X\not\models Q$ and hence, $(X,X)$ is not 
an SE-model of $Q$. Thus, $(X,X)$ is an SE-model of exactly one program, $P$, 
but $(X,X)$ violates ($i$) since $X\not\models Q$; this is a contradiction.

It follows that $X\models (Q\cup F)^X$ must hold, and that there must exist 
$Z\subset X$ such that $Z\models (Q\cup F)^X = Q^X\cup F$. 
So we can conclude $X\models Q$ and that $(Z,X)$ is an SE-model of 
$Q$ but not of $P$. To see the latter, note that $F\subseteq Z$ must hold. 
So if $(Z,X)$ were an SE-model of $P$, then it would also be an SE-model 
of $P\cup F$, contradicting the assumption that $X\in\SM(P\cup F)$. 
Again we get an SE-model, $(Z,X)$, of exactly one of the 
programs, $Q$ in this case. Hence, according to ($ii$), there exists an 
SE-model $(X',X)$ of $P$, $Z\subset X'\subset X$. However, because of 
$F\subset Z$, it follows that $(X',X)$ is also an SE-model of 
$P\cup F$, contradicting our assumption that $X\in\SM(P\cup F)$.

This proves that, given ($i$) and ($ii$) for every SE-model $(X,Y)$ 
such that $(X,Y)$ is an SE-model of exactly one of  $P$ and $Q$, 
no sets of atoms $F$ and $Z$ exists such that $Z$ is  
an answer set of 
exactly one of $P\cup F$ and $Q\cup F$. That is,
$P\equivu Q$ holds.
\end{proof}

From Lemma~\ref{theo-equivu-1} we immediately obtain the following
characterization of uniform equivalence of logic programs.

\begin{theorem}\label{theo-equivu-2} 
Two $\DLP$s, $P$ and $Q$ are uniformly equivalent, $P\equivu Q$, iff, 
for  
interpretations $X$, $Y$, 

\begin{itemize}
\item[($i$)] $(X,X)$ is an SE-model of $P$ iff it is an SE-model of $Q$, and 
\item[($ii$)] $(X,Y)$, where $X\subset Y$, is an SE-model of $P$ 
(respectively $Q$) iff there exists a set $X'$, such that 
$X\subseteq X' \subset Y$, and $(X',Y)$ is an SE-model of $Q$ (respectively $P$).
\end{itemize}
\end{theorem}

\begin{example}
Reconsider the programs $P = \{ a \lor b \}$ and $Q = \{ a \la \naf
b;\; b \la \naf a\}$. By Theorem~\ref{theo-equivu-2}, we can easily
verify that $P$ and $Q$ are uniformly equivalent: Their SE-models
differ only in  
$(\emptyset,ab)$,
which is an SE-model of $Q$ but
not of $P$. Thus, items ($i$) and ($ii$) clearly hold for all other
SE-models. 
Moreover, 
$(a,ab)$
is an SE-model of $P$, and thus
item ($ii$) also holds for 
$(\emptyset,ab)$.
\end{example}

Recall that $P$ and $Q$ are strongly equivalent after 
adding the constraint $~\la a,b$, which enforces exclusive disjunction
(see Example~\ref{exa:excl}).
Uniform equivalence does not require such an addition. 

From Theorem~\ref{theo-equivu-2} 
we can derive 
the following characterization of uniform equivalence. 

\begin{theorem}\label{theo-equivu-3} 
Two $\DLP$s $P$ and $Q$, such that at least one of them is finite, are uniformly 
equivalent, \iec $P\equivu Q$, iff the following conditions hold:
\begin{itemize}
\item[($i$)] for every $X$, 
$(X,X)$ is an SE-model of $P$ iff it is an SE-model of
$Q$, 
and 
\item[($ii$)] for every SE-model $(X,Y) \in \SE(P)\cup \SE(Q)$ such 
 that $X\subset Y$, there exists an SE-model $(X',Y)\in \SE(P)\cap\SE(Q)$ 
 $($=$\SE(P \cup Q)$$)$ such that $X \subseteq X' \subset Y$.
\end{itemize}
\end{theorem}

\begin{proof}
Since ($i$) holds by virtue of Theorem~\ref{theo-equivu-2}, we only need to 
show ($ii$). Assume $(X,Y)$, where $X\subset Y$, is in $\SE(P)\cup\SE(Q)$. 

If $(X,Y)\in \SE(P)\cap\SE(Q)$, then the statement holds. Otherwise, by virtue of Theorem~\ref{theo-equivu-2}, there exists $(X_1,Y)$, 
$X\subseteq X_1\subset Y$, such that $(X_1,Y)$ is in $\SE(P)\cup\SE(Q)$. 
By repeating this argument, we obtain a chain of SE-models 
$(X,Y) = (X_0,Y)$,  $(X_1,Y)$, \ldots , $(X_i,Y)$, \ldots\ such that
$(X_i,Y) \in \SE(P)\cup \SE(Q)$ and $X_i \subseteq X_{i+1}$, 
for all $i \geq 0$. Furthermore, we may
choose $X_1$ such that $X_1$ coincides with $Y$ on all atoms which do
not occur in $P \cup Q$ (and hence all $X_i$, $i\geq 1$, do so). 
Since one of $P$ and $Q$ is finite, 
it follows that $X_i=X_{i+1}$ must hold for some $i \geq 0$ and hence
$(X_i,Y)\in \SE(P)\cap\SE(Q)$ must hold. This proves the result. 
\end{proof}

\subsection{Introducing UE-Models}

In the light of this result, we can capture uniform equivalence of
finite programs by the notion of UE-models defined as follows.  
\begin{definition}[UE-model]\label{def:UE}
Let $P$ be a $\DLP$. Then, 
$(X,Y)\in \SE(P)$ is a {\em
uniform equivalence} $($UE$)$ model of $P$, if for every $(X',Y)\in
\SE(P)$ it holds that $X\subset X'$ implies $X'=Y$.
By $\UE(P)$ we denote the set of all UE-models of $P$.
\end{definition}

That is, the UE-models comprise all total SE-models $(Y,Y)$ of a $\DLP$
plus all its {\em maximal} non-total SE-models $(X,Y)$, with $X\subset Y$.
Formally, 
$$
\UE(P)\quad =\quad \{(Y,Y)\in \SE(P)\}\;\;\cup\;\; \mathit{max}_{\geq} \{(X,Y)\in \SE(P)\mid X\subset Y\};
$$
where $(X',Y') \geq (X,Y)$ iff jointly $Y'=Y$  and $X\subseteq X'$.

By means of UE-models, we then can characterize uniform equivalence of
finite logic programs by the following simple condition. 

\begin{theorem}\label{theo-equivu-4} 
Let $P$ and $Q$ be $\DLP$s. Then,
\begin{enumerate}
\item[$(a)$] $P\equivu Q$ implies $\UE(P)=\UE(Q)$;
\item[$(b)$] $\UE(P)=\UE(Q)$ implies $P\equivu Q$, whenever at least one of the programs $P$, $Q$ is finite.
\end{enumerate}
\end{theorem}

\begin{proof}
For proving $(a)$, let $P\equivu Q$. Then, by 
Theorem~\ref{theo-equivu-2} 
($i$), $\UE(P)$ and $\UE(Q)$ coincide on models $(X,X)$. 
Assume w.l.o.g.~that $(X,Y)$, 
$X\subset Y$, is in $\UE(P)$, but not in $\UE(Q)$. By 
Theorem~\ref{theo-equivu-2} 
($ii$), there exists $(X',Y)$, $X\subseteq X'\subset Y$, which is an 
SE-model of 
$Q$, and by a further application, the existence of $(X'',Y)$, 
$X'\subseteq X''\subset Y$, which is an SE-model of $P$ follows.
Since $X\subset X''$ contradicts $(X,Y)\in \UE(P)$, let $X''=X'=X$, 
\iec $(X,Y)$ is an SE-model of $Q$ as well, but it is not in $\UE(Q)$. 
Hence, there exists $(Z,Y)\in \SE(Q)$, $X\subset Z \subset Y$ and, 
again by Theorem~\ref{theo-equivu-2} ($ii$),  
there exists $(Z',Y)$, $Z\subseteq Z'\subset Y$, which is an SE-model 
of $P$. This again contradicts $(X,Y)\in \UE(P)$. Hence, $\UE(P)=\UE(Q)$ must 
hold.

For $(b)$, assume $\UE(P)=\UE(Q)$, 
and w.l.o.g.~let $P$ be finite. 
Since $\UE(P)=\UE(Q)$ implies Theorem~\ref{theo-equivu-2} ($i$), towards a 
contradiction, suppose that Theorem~\ref{theo-equivu-2} ($ii$) is 
not satisfied, \iec there exists $X\subset Y$, such that either (1) 
$(X,Y)\in \SE(P)$ and not exists $X\subseteq X'\subset Y$, $(X',Y)\in \SE(Q)$, 
or vice versa (2) $(X,Y)\in \SE(Q)$ and not exists $X\subseteq X'\subset Y$, 
$(X',Y)\in \SE(P)$. 

\noindent 
Case (1): We show the existence of a set $Z$, $X\subseteq Z\subset Y$, such that 
$(Z,Y)\in \UE(P)$. 
If $(X,Y) \in\UE(P)$, or $Y$ is finite, this is trivial. So let $(X,Y)\not\in\UE(P)$ 
and $Y$ infinite. Then $Y_P=Y\cap\atm{P}$ and $X_P=X\cap\atm{P}$ are finite, 
$(X_P,Y_P)\in\SE(P)$, and $X_P\subset Y_P$. (To see the latter, observe that 
otherwise we end up in a contradiction by the fact that then $X_P\models P$, hence 
$X\models P$, and thus $(X,X)\in \UE(P)=\UE(Q)$, which implies $(X,Y)\in \SE(Q)$, since $(Y,Y)\in\UE(Q)=\UE(P)$ holds.)  
Since $Y_P$ is finite, there exists a set $Z_P$, $X_P\subseteq Z_P\subset Y_P$, such 
that $(Z_P,Y_P)\in \UE(P)$.
Now, let $Z=Z_P\cup (Y\setminus Y_P)$. Then $X\subseteq Z\subset Y$ holds by 
construction. Furthermore $(Z,Y)\in \UE(P)$, since $Y\setminus Z=Y_P\setminus Z_P$,  
$P^Y=P^{Y_P}$, and $(Z_P,Y_P)\in \UE(P)$. 
By our assumption $(Z,Y)\in \UE(Q)$ follows. Contradiction.

\noindent 
Case (2):  We show the existence of a set $Z$, $X\subseteq Z\subset Y$, such that 
$(Z,Y)\in \UE(Q)$. 
If $(X,Y) \in\UE(Q)$, or $Y$ is finite, this is trivial. So let $(X,Y)\not\in\UE(Q)$, 
and $Y$ infinite. Futhermore, $Y\setminus X\subseteq\atm{P}$ must hold. (To see the 
latter, observe that otherwise we end up in a contradiction by taking any atom 
$a\in Y\setminus X$, such that $a\not\in\atm{P}$, and considering  
$Z=Y\setminus\{a\}$. Then $X\subseteq Z\subset Y$ holds by construction and since 
$(Y,Y)\in\UE(P)=\UE(Q)$, $Y\models P$ and so does 
$Z$, \iec $(Z,Y)\in \SE(P)$, a contradiction.)
However, since $\atm{P}$ is finite, this means that $Y\setminus X$ is finite, \iec 
there cannot exist an infinite chain of $\SE$-models 
$(X,Y)=(X_0,Y), (X_1,Y), \ldots, (X_i,Y),\ldots$, such that $X_i\subset X_j\subset Y$, 
for $i<j$, and $(X_i,Y)\in\SE(Q)$. 
Thus, there exists a maximal model $(Z,Y)\in \UE(Q)$.
By our assumption $(Z,Y)\in \UE(P)$ follows. Contradiction.
Thus, Theorem~\ref{theo-equivu-2} ($ii$) holds as well, proving $P\equivu Q$ in Case (b).
\end{proof}

This result shows that UE-models capture the notion of uniform
equivalence for finite logic programs, in the same manner as SE-models capture strong
equivalence. That is, the essence of a program $P$ with respect to
uniform equivalence is expressed by a semantic condition on $P$ alone.

\begin{corollary}\label{cor-equivu-4} 
Two finite $\DLP$s $P$ and $Q$ are uniformly equivalent, \iec $P\equivu
Q$, if and only if $\UE(P)=\UE(Q)$.
\end{corollary}

\begin{example}
Each SE-model of the program $P=\{a\vee b\}$ satisfies the
condition of an UE-model, and thus $\UE(P) = \SE(P)$. 
The program $Q = \{a\ \la \ \naf b;\; b\ \la\ \naf a\}$ 
has the additional SE-model 
$(\emptyset,ab)$,
and all of its SE-models except this one are UE-models of $Q$. 
Thus, 
$$
\UE(P) = \UE(Q)  =  \{ (a,a);\; (b,b);\; (a,ab);\; (b,ab);\; (ab,ab)\}.
$$
Note that the strong equivalence of $P$ and $Q$ fails because 
$(\emptyset,ab)$
is not an SE-model of $P$. This SE-model is
enforced by the intersection property 
($(X_1,Y)$ and $(X_2,Y)$ in $\SE(P)$ implies $(X_1\cap X_2,Y) \in \SE(P)$). 
This intersection property is satisfied by the Horn program $Q^Y$, 
but violated by the disjunctive program $P^Y$ (=$P$). The
maximality condition of UE-models eliminates this intersection
property.
\end{example}

\begin{example}
Reconsider $P=\{a\la \naf b ;\ a\la b\}$, which has classical models (over
$\{a,b,c\})$ of form 
$\{a\} \subseteq Y \subseteq \{a, b,c\}$. Its UE-models are 
$(X,Y)$ where $X \in \{Y, Y\setminus\{ b\},
Y\setminus\{ c\}\}$.  
Note that the atoms $b$ and $c$ have symmetric
roles in $\UE(P)$. Consequently, the program obtained by exchanging the
roles of $b$ and $c$, $Q=\{a\la \naf c ;\ a\la c\}$ has the same UE
models. Hence, $P$ and $Q$ are uniformly equivalent.
\end{example}

The following example shows 
why the characterization via UE-models fails if both 
compared programs are infinite.
The crucial issue here is the expression of an
``infinite chain'' resulting in an infinite number of 
non-total SE-models. In this case, 
the concept of maximal non-total SE-models
does not capture the general characterization from Theorem~\ref{theo-equivu-2}.

\begin{example}
\label{ex:10}
Consider the programs $P$ and $Q$ over 
$\at=\{a_i \mid i\geq 1\}$, defined by 
\begin{eqnarray*}
P = \{ a_i \la\ \mid i \geq 1 \}, & \textrm{ and } &  
Q = \{ a_i \la \naf a_i,\ a_i \la a_{i+1}
\mid i \geq 1 \}. 
\end{eqnarray*}
Both $P$ and $Q$ have the single classical model $\at = \{ a_i \mid i \geq
1 \}$. Furthermore, $P$ has no ``incomplete'' SE-model $(X,\at)$ such
that $X\subset \at$, while $Q$ has the incomplete SE-models $(X_i,\at)$, where 
$X_i = \{a_1,\ldots,a_i\}$ for $i \geq 0$.
Both $P$ and $Q$ have the same maximal incomplete SE-models (namely
none), and hence they have the same UE-models. 

However, $P\not\equivu Q$, since e.g.\ $P$ has 
an answer set while $Q$ has obviously not. Note that this is caught by our 
Theorem~\ref{theo-equivu-2}, item ($ii$): for $(X_0,\at)$, which
is an SE-model of $Q$ but not of $P$, we cannot find an SE-model $(X,\at)$ of
$P$ between $(X_0,\at)$ and $(\at,\at)$. 
\end{example}

In fact, uniform equivalence of infinite programs 
$P$ and $Q$ cannot be captured by a selection 
of SE-models: 

\begin{theorem}\label{thm:selection}
Let $P$ and $Q$ be infinite $\DLP$s. There is no selection of SE-models, 
$\sigma(\SE(\cdot))$, such that $P$ and $Q$ are uniformly equivalent, 
$P\equivu Q$, if and only if $\sigma(\SE(P)) = \sigma(\SE(Q))$.
\end{theorem}

\begin{proof}
Consider programs over 
$\at=\{a_i \mid i\geq 1\}$ as follows.  
The program $P = \{ a_i \la\ \mid i \geq 1 \}$ 
in Example~\ref{ex:10}, as well as  
\begin{eqnarray*}
Q & = & \{ a_i \la \naf a_i,\ a_i \la a_{i+1},\ a_{2i} \la a_{2i-1}
\mid i \geq 1 \}, \\
R & = & \{ a_i \la \naf a_i,\ a_i \la a_{i+1},\ a_{2i+1} \la a_{2i},\ a_1 \la
\mid i \geq 1 \}, \textrm{ and } \\
S & = & \{ a_i \la \ ,\ \la a_1
\mid i \geq 1 \}.
\end{eqnarray*}
Considering corresponding SE-models, it is easily verified that 
$\SE(P)=\{(\at,\at)\}$, $\SE(S)=\emptyset$, as well as
\begin{eqnarray*}
\SE(Q) & = & \{ (\emptyset,\at), (a_1a_2,\at),\ldots,
(a_1a_2\cdots a_{2i},\at), \ldots, (\at,\at) 
\mid i \geq 0 \}, \textrm{ and } \\
\SE(R) & = & \{ (a_1,\at), (a_1a_2a_3,\at),\ldots,
(a_1a_2\cdots a_{2i+1},\at), \ldots, (\at,\at) 
\mid i \geq 0 \}.
\end{eqnarray*}
Hence, we have that $\SE(Q)\cap\SE(R)=\{(\at,\at)\}$. Observe also that 
$Q\cup X$ and $R\cup X$ do not have an answer set 
for any proper subset 
$X\subset \at$, while $\at$ is (the only) 
answer set of both $Q\cup \at$ 
and $R\cup \at$. Thus, $Q\equivu R$. However, $S\cup \at$ does not have 
an answer set
and we get $Q\not\equivu S$ and $R\not\equivu S$. Since $P$ has 
the answer set $\at$, we finally conclude that $P\not\equivu Q$, 
$P\not\equivu R$, and $P\not\equivu S$. 

Towards a contradiction, let us assume that there exists a selection function 
$\sigma(\SE(\cdot))$, such that $P_i\equivu P_j$ iff 
$\sigma(\SE(P_i))=\sigma(\SE(P_j))$, for $P_i,P_j \in \{P,Q,R,S\}$.
Then, $\sigma(\SE(S))=\emptyset$ and, since $P\not\equivu S$, 
$\sigma(\SE(P))=\{(\at,\at)\}$. Furthermore, $Q\equivu R$ implies 
$\sigma(\SE(Q))=\sigma(\SE(R))$ and by $\SE(Q)\cap\SE(R)=\{(\at,\at)\}$ we 
conclude either $\sigma(\SE(Q))=\sigma(\SE(R))=\emptyset$, or 
$\sigma(\SE(Q))=\sigma(\SE(R))=\{(\at,\at)\}$. From $P\not\equivu Q$, the 
former follows, \iec $\sigma(\SE(Q))=\sigma(\SE(R))=\emptyset$. However, 
then $\sigma(\SE(Q))=\sigma(\SE(S))$ while $Q\not\equivu S$, which is a contradiction.
\end{proof}

\subsection{Consequence under Uniform Equivalence}

Based on  UE-models, we define an associated notion
of consequence under {\em uniform equivalence}. 

\begin{definition}[UE-consequence] 
\label{def:uecons}
A rule, $r$, is an \emph{UE-consequence} of a 
program $P$, denoted $P\modelsu r$, if 
$(X,Y)\in\SE(r)$,
for all $(X,Y)\in \UE(P)$.
\end{definition}

Clearly, $P \modelsu r$ for all $r \in P$, and $\emptyset \models r$
iff $r$ is a classical tautology. The next result shows that the UE-models
of a program remain invariant under addition of UE-consequences. 

\begin{proposition}\label{prop-equivu-rule}
For any program $P$ and rule $r$, if $P\modelsu r$ then 
$\UE(P)=\UE(P\cup\{r\})$.
\end{proposition} 
\begin{proof}
Let $P\modelsu r$, we show that $\UE(P)=\UE(P\cup\{r\})$.

\noindent``$\subseteq$'': Let $(X,Y)\in\UE(P)$. Then, by hypothesis $Y\models r$ and $X\models
r^Y$. Hence, $Y\models P\cup\{r\}$ and $X\models (P\cup\{r\})^Y$. Suppose
$(X,Y)\not\in\UE(P\cup\{r\})$. Then there exists a set $X'$, $X\subset
X'\subset Y$, such that $(X',Y)\models (P\cup\{r\})^Y$. But then
$X'\models P^Y$, which contradicts $(X,Y)\in\UE(P)$. It follows that
$(X,Y) \in \UE(P\cup\{r\})$ . 

\noindent``$\supseteq$'': Let $(X,Y)\in\UE(P\cup\{r\})$. Then $X\models P^Y$ and $Y\models P$. Suppose 
$(X,Y) \notin \UE(P)$. Then, some $(X',Y)\in \UE(P)$ exists such that $X\subset X'\subset Y$. By hypothesis, $(X',Y)\in\SE(r)$ 
(otherwise $P\not\modelsu r$), hence $X' \models (P\cup \{r\})^Y$. But
then $(X,Y) \in \UE(P\cup\{r\})$, which is a contradiction. It follows
$(X,Y) \in \UE(P)$.
\end{proof}

As usual, we write $P\modelsu R$ for any set of rules $R$ if $P\modelsu r$ 
for all $r\in R$. As a corollary, taking Theorem~\ref{theo-equivu-4} $(b)$ into account, 
we get the following.

\begin{corollary}\label{cor-equivu-rule} 
For any finite program $P$ and set of rules $R$, if $P\modelsu R$ then $P\cup R\equivu P$.
\end{corollary}

From this proposition, we also obtain an alternative characterization of
uniform equivalence in terms of UE-consequence.

\begin{theorem}\label{prop-equivu-prog}
Let $P$ and $Q$ be $\DLP$s. Then,
\begin{enumerate}
\item[$(a)$] $P\equivu Q$ implies $P\modelsu Q$ and $Q\modelsu P$;
\item[$(b)$] $P\modelsu Q$ and $Q\modelsu P$ implies $P\equivu Q$, whenever at least one of 
the programs $P$, $Q$ is finite.
\end{enumerate}
\end{theorem} 

\begin{proof}
In Case $(a)$, we have $\UE(P)=\UE(Q)$ if $P\equivu
Q$ by Theorem~\ref{theo-equivu-4} $(a)$, and thus $P$ and $Q$ have the same
UE-consequences. Since $(X,Y)\models P$ (resp.\ $(X,Y) \models Q$),
for all $(X,Y)\in \UE(P)$ (resp.\ $(X,Y) \in \UE(Q)$), it follows $Q
\modelsu P$ and $P \modelsu Q$. 
For $(b)$, we apply Proposition~\ref{prop-equivu-rule} repeatedly 
and obtain $\UE(P)=\UE(P\cup Q)=\UE(Q)$. By Theorem~\ref{theo-equivu-4} $(b)$
$P\equivu Q$.
\end{proof}

Rewriting this result in terms of SE- and UE-models gives the 
following characterization
(which has also been derived for finite programs in~\cite{Eiter04a}; Proposition~5).

\begin{proposition}\label{prop:subset}
Let $P$ and $Q$ be 
$\DLP$s. Then, 
\begin{enumerate}
\item[$(a)$] $P\equivu Q$ implies $\UE(P)\subseteq \SE(Q)$ and $\UE(Q)\subseteq \SE(P)$;
\item[$(b)$] $\UE(P)\subseteq \SE(Q)$ and $\UE(Q)\subseteq \SE(P)$ implies $P\equivu Q$, 
whenever at least one of the programs $P$, $Q$ is finite.
\end{enumerate}
\end{proposition}

We note that with respect to uniform equivalence, every program $P$
has a canonical normal form, $P^{*}$, given by its UE-consequences, \iec
$P^{*} = \{r \mid P\modelsu r\}$. 
Clearly, $P \subseteq P^{*}$ holds for every program $P$, and
$P^{*}$ has exponential size. Applying optimization methods 
built on UE-consequence, $P$ resp.\ $P^{*}$ may be transformed into
smaller uniformally equivalent programs; we leave this for further
study.

As for the relationship of UE-consequence to classical consequence and
cautious consequence under  
answer set semantics, we note the following
hierarchy. Let $\models_c$ denote consequence from the 
answer sets,
\iec $P \models_c r$ iff $M \models r$ for every $M \in \SM(P)$.

\begin{proposition}\label{prop:cons}
For any finite program $P$ and rule $r$, ($i$) $P \modelsu r$ implies 
$P \cup F \models_c r$, for each set of facts $F$; ($ii$) $P \cup F
\models_c r$, for each set of facts $F$, implies $P\models_c r$;
and ($iii$) $P \models_c r$ implies $P \models r$.
\end{proposition}
\begin{proof}
Since each  
answer set is a classical model, it remains to show ($i$).
Suppose $P \modelsu r$. Then, $P \equivu P \cup \{r\}$ by 
Corollary~\ref{cor-equivu-rule}, i.e., $\SM(P \cup F) = \SM(P \cup
\{r\} \cup F)$, for each set of facts $F$. Since $X \models r$ for
each $X \in \SM(P\cup\{r\}\cup F)$, it follows that $P\cup F \models_c
r$, for each set of facts $F$. 
\end{proof}

This hierarchy is strict, \iec none of the implications holds in the
converse direction. (For ($i$), note that $\{ a \la \naf a\}
\models_c a$ but $\{ a \la \naf a\} \not\modelsu a$, since the
UE-model $(\emptyset,\{a\})$ violates $a$.)

We next present a semantic characterization in terms of UE-models, under
which UE- and classical consequence and thus all four 
notions of consequence coincide.

\begin{lemma}\label{lemm-A}
Let $P$ be a $\DLP$. Suppose that $(X,Y)\in\UE(P)$ implies $X\models
P$ $($\iec $X$ is a model of $P$$)$. Then, $P\models r$ implies $P\modelsu
r$, for every rule $r$.
\end{lemma}

\begin{proof}
Consider $(X,Y)\in\UE(P)$. By hypothesis, $X\models P$ and $P\models
r$, thus $X\models r$, which implies $X\models r^X$. Furthermore,
$Y\models r$ since $Y\models P$. We need to show that $X\models
r^Y$. Note that either $r^Y$ is void, or, since $X\subseteq Y$, we have
$r^Y=r^X$. In both cases $X\models r^Y$ follows, which proves
$(X,Y)\in\SE(r)$. Thus, $P\modelsu r$.
\end{proof}

\begin{theorem}\label{theo-C}
Let $P$ be any $\DLP$. Then the following conditions are equivalent:\\[-3ex]
\begin{enumerate}
 \item[($i$)] 
$P\modelsu r$ iff $P\models r$, for every rule $r$.
 \item[($ii$)] 
For every $(X,Y)\in\UE(P)$, it holds that $X\models P$. 
\end{enumerate}
\end{theorem}

\begin{proof}

\noindent($ii$) $\Rightarrow$ ($i$). 
Suppose ($ii$) holds. The only-if direction in ($i$) holds immediatly 
by Lemma~\ref{lemm-A}. The if direction in ($i$) holds in gerenal, 
since $P\modelsu r$ iff $\UE(P)\subseteq \SE(r)$. The latter clearly
implies that each total SE-model of $P$ is a total SE-model of $r$. 
Consequently, $P\models r$.

\noindent($i$) $\Rightarrow$ ($ii$). Suppose $P\modelsu r$ iff
$P\models r$, for every rule $r$, but there exists some UE-model
$(X,Y)$ of $P$ such that $X\not\models P$. Hence $X\not\models r$ for
some rule $r \in P$. Let $r'$ be the rule which results from $r$ by
shifting the negative literals to the head, i.e., $\head{r'} =
\head{r}\cup \bodyn{r}$, $\bodyp{r'}=\bodyp{r}$, and
$\bodyn{r'}=\emptyset$. Then, $X \not\models r'$. On the other hand,
$r \in P$ implies $(X,Y) \models r$. Hence, $Y\models  r$ and thus $Y
                \models r'$. Moreover,
$\bodyn{r'}=\emptyset$ implies that  $r' \in P^Y$, and hence $X\models
r'$. This is a contradiction. It follows that $X\models P$ for each
UE-model  $(X,Y)$ of $P$. 
\end{proof}

An immediate corollary to this result is that for finite 
{\em positive} programs, 
UE-consequence collapses with classical consequence, and
hence uniform equivalence of finite  positive programs amounts to
classical equivalence. We shall obtain these results as corollaries of
more general results in Section~\ref{sec:pos}, though.

\section{Relativized Notions of Strong and Uniform Equivalence}\label{sec:rel}

In what follows, we formally introduce the 
notions of relativized strong equivalence (RSE)
and 
relativized uniform equivalence (RUE).
\begin{definition}\label{def:seue}
Let $P$ and $Q$ be programs and let $A$ be a set of atoms.
Then,
\begin{enumerate}
\item[(i)]
$P$ and $Q$ are {\em strongly equivalent relative to $A$},
denoted
$P \equivs^A Q$, iff 
$P\cup R\equiv Q\cup R$,
for all 
programs $R$ over $A$;
\item[(ii)]
$P$ and $Q$ are {\em uniformly equivalent relative to $A$},
denoted
$P \equivu^A Q$, iff 
$P\cup F \equiv Q\cup F$,
for all $($non-disjunctive$)$ facts 
$F\subseteq A$.
\end{enumerate}
\end{definition}

Observe that the range of applicability of these notions covers 
ordinary equivalence (by setting $A=\emptyset$) 
of two programs $P$, $Q$,
and
{\em general} strong (resp.\ uniform) equivalence 
(whenever $\var{P\cup Q}\subseteq A$). 
Also the following relation holds:
For any set $A$ of atoms, let 
$A'=A\cap \var{P\cup Q}$.  Then,
$P \equive^A Q$ holds, iff 
$P \equive^{A'} Q$ holds, for $e\in\{s,u\}$.

Our first 
main
result lists some properties for relativized strong equivalence.
Among them, we
show
that 
RSE
shares an important property 
with
general strong equivalence:
In particular,
from the proofs of the results in \cite{Lifschitz01,Turner03}, 
it appears
that for strong equivalence, 
only the addition of unary rules 
is crucial. 
That is, by constraining the rules in
the set $R$ in Definition~\ref{def:seue}
to unary ones
does not lead to a different concept. 

\begin{lemma}\label{thm:central}
For programs $P$, $Q$, and a set of atoms $A$, 
the following statements are equivalent:
\begin{itemize}
\item[$(1)$] 
there exists a program  $R$ over $A$, such
that $\SM(P\cup R)\not\subseteq \SM(Q\cup R)$;
\item[$(2)$] there exists a unary program $U$ over $A$, such
that $\SM(P\cup U)\not\subseteq \SM(Q\cup U)$;
\item[$(3)$] there exists an interpretation $Y$, such that (a)
$Y\models P$; (b) for each $Y'\subset Y$ with $(Y'\cap A)=(Y\cap A)$, $Y'\not\models P^Y$ holds; and (c) $Y\models Q$ implies existence of an $X\subset Y$, such that $X\models Q^Y$ and, for each $X'\subset Y$ with $(X'\cap A)=(X\cap A)$, $X'\not\models P^Y$ holds.
\end{itemize}
\end{lemma}
\begin{proof}
(1) $\Rightarrow$ (3):
Suppose an interpretation $Y$ and 
a set $R$ of rules over $A$, such
that $Y\in\SM(P\cup R)$ and $Y\notin \SM(Q\cup R)$. 
From $Y\in\SM(P\cup R)$, we get $Y\models P\cup R$ and, for each 
$Z\subset Y$, $Z\not\models P^Y \cup R^Y$. Thus (a) holds, and since 
$Y'\models R^Y$ holds, for each $Y'$ with $(Y'\cap A)=(Y\cap A)$, 
(b) holds as well.
From $Y\notin \SM(Q\cup R)$, we get that either $Y\not\models Q\cup R$ or
there exists an interpretation $X\subset Y$, such that $X\models Q^Y\cup R^Y$.
Note that $Y\not\models Q\cup R$ implies $Y\not\models Q$, since from above, 
we have $Y\models R$. Thus, in the case of  $Y\not\models Q\cup R$, (c) holds; otherwise
we get that $X\models Q^Y$. Now since $X\models R^Y$, we know that, for
each $X'\subset Y$ with $(X'\cap A) = (X\cap A)$, $X'\not\models P^Y$ has to hold, 
otherwise $Y\notin \SM(P\cup R)$. Hence, (c) is satisfied.

(3) $\Rightarrow$ (2):
Suppose an interpretation $Y$, such that Conditions (a--c) hold.
We have two cases: First, if $Y\not\models Q$, 
consider
the unary program $U=(Y\cap A)$. 
By 
Conditions (a) and (b), it is easily seen that $Y\in\SM(P\cup U)$,
and from $Y\not\models Q$, $Y\notin \SM(Q\cup U)$ follows.
So suppose, $Y\models Q$. 
By (c), there exists an $X\subset Y$, such that
$X\models Q^Y$. 
Consider the program
$U= (X\cap A)\cup \{ p\la q \mid p,q\in (Y\setminus X)\cap A\}$. Again, $U$ is unary over $A$.
Clearly, $Y\models Q\cup U$ and $X\models Q^Y\cup U$. Thus
$Y\notin \SM(Q\cup U)$. It remains to show that $Y\in\SM(P\cup U)$. 
We have $Y\models P\cup U$. Towards a contradiction, suppose a $Z\subset Y$,
such that $Z\models P^Y\cup U$. By definition of $U$, $Z\supseteq (X\cap A)$.
If $(Z\cap A)=(X\cap A)$, Condition (c) is violated; if 
$(Z\cap A) = (Y\cap A)$, Condition (b) is violated. 
Thus, $(X\cap A)\subset (Z\cap A)\subset (Y\cap A)$.
But then, $Z\not\models U$, since 
there exists at least one rule $p\la q$ in $U$, such that 
$q\in Z$ and $p\notin Z$. 
Contradiction.  

(2) $\Rightarrow$ (1) is obvious.
\end{proof}

The next result is an immediate consequence of the fact that 
Propositions (1) and (2) from above result are equivalent.

\begin{corollary}\label{cor:unary}
For programs $P$, $Q$, and a set of atoms $A$,
$P \equivs^A Q$ holds iff, for each unary program $U$ over $A$,
$P\cup U \equiv Q \cup U$ holds.
\end{corollary}

We emphasize that therefore also for relatived equivalences, 
it holds that restricting the syntax of the added rules, 
RSE and RUE are the only concepts which differ. 
Note that this generalizes an observation reported in~\cite{Pearce04} to 
relativized notions of equivalence, namely
that uniform and strong equivalence are the only 
forms of equivalence obtained by varying the 
logical form of expressions in the extension.

\subsection{A Characterization for Relativized Strong Equivalence}

In this section, 
we 
provide a semantical characterization of 
RSE
by 
generalizing the notion of SE-models. Hence, 
our aim is to 
capture the problem $P\equivs^A Q$
in model-like terms. 
We emphasize that the forthcoming results are also applicable 
to infinite programs.
Moreover, having found a suitable notion of \emph{relativized SE-models}, 
we expect
that a corresponding pendant for 
relativized uniform equivalence
can be derived
in the same manner as general UE-models 
are defined over general SE-models. 
As in the case of UE-models, we need some restrictions concerning 
the infinite case, \iec if infinite programs are considered.

We introduce 
the following notion.

\begin{definition}\label{def:relSE}
Let $A$ be a set of atoms. A pair of interpretations $(X,Y)$
is a \emph{(relativized) $A$-SE-interpretation} iff either
$X=Y$ or $X\subset (Y\cap A)$.
The former are called total and the latter non-total $A$-SE-interpretations.

Moreover,
an $A$-SE-interpretation $(X,Y)$ is 
a \emph{(relativized) $A$-SE-model} of  a program $P$
iff
\begin{enumerate}
\item[(i)] $Y\models P$;
\item[(ii)] for all $Y'\subset Y$ with $(Y'\cap A)=(Y\cap A)$, $Y'\not\models P^Y$; and
\item[(iii)] 
$X\subset Y$ implies
existence of a $X'\subseteq Y$ with $(X'\cap A)=X$, 
such that $X'\models P^Y$ holds.
\end{enumerate}
The set of $A$-SE-models of $P$ is given by $\SEA(P)$.
\end{definition}

Compared to SE-models, this definition is more involved. 
This is due to the fact, that we have to take care of two different
effects when relativizing strong equivalence. 
The first one is as follows:
Suppose a program $P$ has among its SE-models the pairs 
$(Y,Y)$ and $(Y',Y)$ with 
$(Y'\cap A)=(Y\cap A)$ and $Y'\subset Y$. 
Then, $Y$ never becomes 
an answer set of a program $P\cup R$,
regardless of the rules $R$ over $A$ we add to $P$. This is due
to the fact that either $Y'\models (P\cup R)^Y$ still holds for some $Y'\subset Y$, 
or, $Y\not\models (P\cup R)^Y$
(the latter is a consequence of finding an $R$ such that 
$Y'\not\models (P\cup R)^Y$,
for $(Y'\cap A)=(Y\cap A)$, $Y'\subset Y$ modelling $P$).
In other words, for the construction of a program $R$ over $A$, such 
that $\SM(P\cup R)\neq \SM(Q\cup R)$, it is not worth to 
to pay attention to any original SE-model 
of $P$ of the form $(\cdot,Y)$,
whenever there exists a $(Y',Y)\in\SE(P)$ with $(Y'\cap A)=(Y\cap A)$.
This motivates Condition~(ii).
Condition~(iii) deals with 
a different effect:
Suppose $P$ has SE-models $(X,Y)$ and $(X',Y)$, with $(X\cap A)=(X'\cap A)\subset (Y\cap A)$.
Here, it is not possible to eliminate just one of these two SE-models by
adding rules over $A$. Such SE-models which do not differ
with respect to $A$, are collected into a single $A$-SE-model $((X\cap A),Y)$.

The different role of these two independent conditions becomes even 
more apparent
in the following cases.
On the one hand, setting $A=\emptyset$, 
the $A$-SE-models of a program $P$ collapse with the 
answer sets 
of $P$. 
More precisely, all such $\emptyset$-SE-models have to be of the
form $(Y,Y)$, 
and it holds that $(Y,Y)$ is an $\emptyset$-SE-model
of a $\DLP$ $P$ iff $Y$ is 
an answer set of 
$P$. This is easily seen by the fact that under $A=\emptyset$,
Conditions (i) and (ii) in Definition~\ref{def:relSE}
exactly coincide with 
the characterization of 
answer sets, following Proposition~\ref{prop:sm}.
Therefore, $A$-SE-model-checking for $\DLP$s is not possible in polynomial 
time in the general case;
otherwise we get that checking whether a $\DLP$ has some 
answer set is $\NP$-complete; which is in contradiction to known results~\cite{eite-gott-95},
provided the polynomial hierarchy does not collapse.  
On the other hand, if each atom from $P$ is contained in $A$, 
then the $A$-SE-models of $P$ coincide with the 
SE-models (over $A$) of $P$.
The conditions in Definition~\ref{def:relSE} are 
hereby instantiated as follows: 
A pair $(X,Y)$ is an $A$-SE-interpretation iff $X\subseteq Y$, and
by (i) we get $Y\models P$,
(ii) is trivially satisfied, and (iii) states $X\models P^Y$. 

The central result is as follows.
In particular, we 
show that $A$-SE-models capture the notion of $\equivs^A$
in the same manner as SE-models capture $\equivs$.

\begin{theorem}\label{thm:A}
For programs $P$, $Q$, and a set of atoms $A$,
$P\equivs^A Q$ holds iff 
$\SEA(P)=\SEA(Q)$.
\end{theorem}

\begin{proof}
First suppose $P\not \equivs^A Q$ and wlog consider 
for some $R$ over $A$, $\SM(P\cup R)\not\subseteq \SM(Q\cup R)$.
By Lemma~\ref{thm:central}, 
there exists an interpretation $Y$,
such that  
(a) $Y\models P$;
(b) for each $Y'\subset Y$ with $(Y'\cap A)=(Y\cap A)$, $Y'\not\models P^Y$; and  
(c) $Y\not\models Q$ or there exists an interpretation 
$X\subset Y$, such that $X\models Q^Y$ and, 
for each $X'\subset Y$ with $(X'\cap A)=(X\cap A)$, $X'\not\models P^Y$.
First suppose $Y\not\models Q$, 
or $Y\models Q$ and $(X\cap A)=(Y\cap A)$.
Then $(Y,Y)$ is an 
$A$-SE-model of $P$ but not of $Q$.
Otherwise, \iec $Y\models Q$ and $(X\cap A)\subset (Y\cap A)$,
$((X\cap A), Y)$ is an $A$-SE-model of $Q$. 
But, by Condition~(c), $((X\cap A), Y)$ is not an $A$-SE-model of $P$.

For the converse direction of the theorem, suppose a pair $(Z,Y)$, such 
that wlog $(Z,Y)$ is an $A$-SE-model of $P$ but not of $Q$.
First, let $Z=Y$. 
We show that $\SM(P\cup R)\not\subseteq \SM(Q\cup R)$ for some program 
$R$ over $A$.
Since $(Y,Y)$ is an $A$-SE-model of $P$, we get 
from Definition~\ref{def:relSE},  
that $Y\models P$ and, for each $Y'\subset Y$ with 
$(Y\cap A)=(Y'\cap A)$, $Y'\not\models P^Y$.
Thus, 
Conditions~(a) and (b) 
in Part~(3) of 
Lemma~\ref{thm:central} are
satisfied 
for $P$ by
$Y$. 
On the other hand, $(Y,Y)$ is not an $A$-SE-model of $Q$. 
By Definition~\ref{def:relSE}, 
either $Y\not\models Q$, or there exists a 
$Y'\subset Y$, with $(Y'\cap A)=(Y\cap A)$, such that
$Y'\models Q^Y$.   
Therefore, Condition~(c) from Lemma~\ref{thm:central} 
is satisfied by either $Y\not\models Q$ or,
if $Y\models Q$, by setting $X=Y'$.
We apply Lemma~\ref{thm:central}
and get 
the desired result. Consequently, $P\not\equivs^A Q$.
So suppose, $Z\neq Y$. 
We show that then $\SM(Q\cup R)\not\subseteq \SM(P\cup R)$ holds, 
for some program $R$ over $A$.
First, observe that whenever $(Z,Y)$ is an $A$-SE-model of $P$, then 
also $(Y,Y)$ is an $A$-SE-model of $P$. Hence, the case where 
$(Y,Y)$ is not an $A$-SE-model of $Q$ is already shown.
So, suppose $(Y,Y)$ is an $A$-SE-model of $Q$.
We have $Y\models Q$ and, for each $Y'\subset Y$ with $(Y'\cap A)=(Y\cap A)$,
$Y'\not\models Q^Y$.
This satisfies Conditions~(a) and (b) in Lemma~\ref{thm:central} for $Q$.
However, since $(Z,Y)$ is not an $A$-SE-model of $Q$, for each 
$X'\subset Y$ with $(X'\cap A)=Z$, $X'\not\models Q^Y$ holds. 
Since $(Z,Y)$ in turn is an $A$-SE-model of $P$, there exists
an $X\subset Y$ with $(X\cap A)=Z$, such that  $X\models P^Y$.
These observations 
imply that (c) holds in Lemma~\ref{thm:central}. 
We apply the lemma 
and finally get
$P\not\equivs^{A} Q$.
\end{proof}
Altough $A$-SE-models are more involved than SE-models, they 
share some fundamental properties with general SE-models. 
On the other hand, some properties do not generalize to 
$A$-SE-models. We shall discuss these issues in detail in Section~\ref{sec:properties}.
For the moment, we list some 
observations, concerning the relation between SE-models and $A$-SE-models, 
in order to present some examples.

\begin{lemma}\label{lemma:setoase}
Let $P$ be a program and $A$ be a set of atoms. We have the following 
relations between $A$-SE-models and SE-models.
\begin{itemize}
\item[$(i)$]
If $(Y,Y)\in\SEA(P)$, then $(Y,Y)\in\SE(P)$.
\item[$(ii)$]
If $(X,Y)\in\SEA(P)$, then $(X',Y)\in\SE(P)$, for some $X'\subseteq Y$ with $(X'\cap A)=X$.
\end{itemize}
\end{lemma}

\begin{table}[t!]
\begin{center}
\begin{tabular}{l|l|l}
$A$ & $A$-SE-models of $Q$ & $A$-SE-models of $Q'$ \\
\hline
$\{a,b,c\}$ & $(abc,abc), (a,abc), (b,abc)$ & $(abc,abc), (a,abc), (b,abc), (\emptyset,abc)$ \\
$\{a,b\}$ & $(abc,abc), (a,abc), (b,abc)$ & $(abc,abc), (a,abc), (b,abc), (\emptyset,abc)$ \\
$\{a,c\}$ & $(abc,abc), (a,abc), (\emptyset,abc)$ & $(abc,abc), (a,abc), (\emptyset,abc)$ \\
$\{b,c\}$ & $(abc,abc), (\emptyset,abc), (b,abc)$ & $(abc,abc), (b,abc), (\emptyset,abc)$ \\
$\{a\}$ & - & - \\ 
$\{b\}$ & - & - \\ 
$\{c\}$ & $(abc,abc)$, $(\emptyset, abc)$ & $(abc,abc)$, $(\emptyset, abc)$ \\
$\emptyset$  & -  & - \\
\end{tabular}
\end{center}
\caption{Comparing the $A$-SE-models for Example Programs $Q$ and $Q'$.}\label{tab:1}
\end{table}

\begin{example}
Consider  the programs
\begin{eqnarray*}
Q  & = &  \{ a \OR b\la ;  \,
a \la c;\,
b \la c;\,
\la \naf c;\,
c \la a, b\};\\
Q' & = & \{ 
a\la \naf b;\, b\la \naf a;\,
a \la c;\,
b \la c;\,
\la \naf c;\,
c \la a, b\}.
\end{eqnarray*}
Thus,
$Q'$ results from $Q$ by replacing 
the disjunctive rule
$a\OR b \la$ by the two rules $a\la \naf b;\, b\la \naf a$.

Table~\ref{tab:1} lists, for each $A\subseteq \{a,b,c\}$,
the $A$-SE-models of $Q$ and $Q'$, respectively.
The first row of 
the table
gives the SE-models 
(over $\{a,b,c\}$)
for $Q$ and $Q'$. 
From this row, we can by Definition~\ref{def:relSE} and 
Lemma~\ref{lemma:setoase}, obtain 
the other rows quite easily.
Observe that we have $Q\not\equivs Q'$. 
The second row shows that, for $A=\{a,b\}$, 
$Q\not\equivs^A Q'$, as well.
Indeed, adding $R=\{a \la b;\, b\la a\}$ yields $\{a,b,c\}$ as 
answer set 
of $Q\cup R$, whereas $Q'\cup R$ has no 
answer set.
For all other $A\subset \{a,b,c\}$, the $A$-SE-models of $Q$ and $Q'$ 
coincide.
Basically, there are two different reasons. 
First, for $A=\{a,c\}$, $A=\{b,c\}$, or $A=\{c\}$, Condition~(iii)
from Definition~\ref{def:relSE} comes into play. 
In those cases, at least one of the SE-interpretations $(a,abc)$ or $(b,abc)$
is ``switched'' to $(\emptyset,abc)$, and thus the original difference between
the SE-models disappears when considering $A$-SE-models.
In the remaining cases, \iec $A\subset \{a,b\}$, 
Condition~(ii) prevents any $(\cdot,abc)$ to be an $A$-SE-model of $Q$ or $Q'$.
Then, neither $Q$ nor $Q'$ possesses any $A$-SE-model. 
\end{example}

\subsection{A Characterization for Relativized Uniform Equivalence}

In what follows, we consider the problem of checking relativized uniform 
equivalence. Therefore, we shall make use of the newly introduced $A$-SE-models
in the same manner  as 
Section~\ref{sec:char} provided characterizations for uniform equivalence
using SE-models.\footnote{For a slightly different way to prove the main
results on RUE, we refer to~\cite{Woltran04}.} 

We start with a generalization of Lemma~\ref{theo-equivu-1}.
The proof is similar to the proof of Lemma~\ref{theo-equivu-1} and thus
relegated to the Appendix.

\begin{lemma}\label{theo-equivru-1} 
Two $\DLP$s $P$ and $Q$ are uniformly equivalent wrt to a set of atoms $A$, 
i.e.\ $P\equivu^A Q$, iff 
for every $A$-SE-model $(X,Y)$, such that $(X,Y)$ is an $A$-SE-model 
of exactly one of the programs $P$ and $Q$,  
it holds that ($i$) $(Y,Y)\in \SE^A(P)\cap \SE^A(Q)$, and  
($ii$) there exists an $A$-SE-model $(X',Y)$, $X\subset X'\subset Y$, 
of the other program.
\end{lemma}

From Lemma~\ref{theo-equivru-1} we immediately obtain the following
characterization of relativized uniform equivalence. 

\begin{theorem}\label{theo-equivru-2} 
Two programs, $P$ and $Q$ are uniformly equivalent wrt to a set of atoms
$A$, $P\equivu^A Q$, iff
\begin{itemize}
\item[($i$)] for each $Y$, 
$(Y,Y)\in\SE^A(P)$ iff $(Y,Y)\in\SE^A(Q)$, \iec
the total $A$-SE-models of $P$ and $Q$ coincide; 
\item[($ii$)] for each $(X,Y)$, where $X\subset Y$, $(X,Y)$ 
is an $A$-SE-model of $P$ 
(respectively $Q$) iff there exists a set $X'$, such that 
$X\subseteq X'\subset Y$, and $(X',Y)$ is an $A$-SE-model of $Q$ (respectively $P$).
\end{itemize}
\end{theorem}

In contrast to uniform equivalence, we can obtain further characterizations
for $\equivu^A$ also for infinite programs, provided that $A$ is finite.

\begin{theorem}\label{theo-equivru-3} 
Let $P$ and $Q$ be programs, $A$ a set of atoms, such that 
$P$, $Q$, or $A$ is finite.
Then $P\equivu^A Q$, iff the following conditions hold:
\begin{itemize}
\item[($i$)] for each $Y$, $(Y,Y)\in\SE^A(P)$ iff $(Y,Y)\in\SE^A(Q)$, \iec
the total $A$-SE-models of $P$ and $Q$ coincide; 
\item[($ii$)] 
for each $(X,Y) \in \SE^A(P)\cup \SE^A(Q)$ such 
that $X\subset Y$, there exists an $(X',Y)\in \SE^A(P)\cap\SE^A(Q)$ 
such that $X \subseteq X' \subset Y$.
\end{itemize}
\end{theorem}

The result is proved by
the same argumentation as used in the proof of Theorem~\ref{theo-equivu-3}. 
The only additional argumentation is needed for the cases 
that $P$ and $Q$ are both infinite, but $A$ is finite. Recall
that in this case there is also only a finite number of non-total
$A$-SE-interpretations $(X,Y)$ for fixed $Y$, since $X\subseteq A$ holds by
definition of $A$-SE-interpretation.
Therefore, any chain (as used in the proof of Theorem~\ref{theo-equivu-3})
of different $A$-SE-models $(X,Y)$ with fixed $Y$ is finite.

As mentioned before,
we aim at defining
relativized $A$-UE-models over
$A$-SE-models in the same manner as
general UE-models are defined over general SE-models, 
following Definition~\ref{def:UE}.

\begin{definition}\label{def:relUE}
Let $A$ be a set of atoms and $P$ be a program.
A pair $(X,Y)$ is a {\em (relativized) $A$-UE-model} of $P$
iff it is an $A$-SE-model of $P$ and,
for every $A$-SE-model $(X',Y)$ of $P$, $X\subset X'$ implies $X'=Y$.
The set of $A$-UE-models of $P$ is given by $\UEA(P)$.
\end{definition}

An alternative characterization of $A$-UE-models, which will be 
useful later, is immediately obtained from Definitions~\ref{def:relSE} 
and~\ref{def:relUE}
as follows.

\begin{proposition}\label{prop:rel}
An $A$-SE-interpretation  $(X,Y)$ is an $A$-UE-model of a program $P$ iff
\begin{enumerate}
\item[(i)] $Y\models P$;
\item[(ii)] for each $X''\subset Y$ with
either $(X\cap A)\subset (X''\cap A)$ or $(X''\cap A)=(Y\cap A)$, $X''\not\models P^Y$; and
\item[(iii)] if $X\subset Y$, there exists a $X'\subseteq Y$ with $(X'\cap A)=(X\cap A)$,
such that $X'\models P^Y$.
\end{enumerate}
\end{proposition}

Next, we derive the desired characterization for relativized
uniform equivalence, generalizing the results in Theorem~\ref{theo-equivu-4}.

\begin{theorem}\label{thm:RUE} 
Let $P$ and $Q$ be $\DLP$s, and $A$ a set of atoms. Then,
\begin{enumerate}
\item[$(a)$] $P\equivu^A Q$ implies $\UEA(P)=\UEA(Q)$;
\item[$(b)$] $\UEA(P)=\UEA(Q)$ implies $P\equivu^A Q$, whenever at least one of $P$, $Q$, or $A$ is finite.
\end{enumerate}
\end{theorem}
\begin{proof}
Proving $(a)$ is basically done 
as for Theorem~\ref{theo-equivu-4}, 
applying Theorem~\ref{theo-equivru-2} instead of Theorem~\ref{theo-equivu-2}.

We proceed with the more interesting part $(b)$.
First assume that $P$ or $A$ is finite. The case where 
$Q$ (or $A$) is finite is 
analogous. 
Assume $\UEA(P)=\UEA(Q)$. 
Then Property~($i$) of Theorem~\ref{theo-equivru-2} holds, and towards a 
contradiction, suppose that Theorem~\ref{theo-equivru-2} ($ii$) is 
not satisfied, \iec there exists $X\subset Y$, such that either (1) 
$(X,Y)\in \SEA(P)$ and not exists $X\subseteq X'\subset Y$, 
$(X',Y)\in \SEA(Q)$, 
or vice versa (2) $(X,Y)\in \SEA(Q)$ and not exists $X\subseteq X'\subset Y$, 
$(X',Y)\in \SEA(P)$. 

\noindent 
Case (1): We show the existence of a set $Z$, 
$X\subseteq Z\subset Y$, such that 
$(Z,Y)\in \UEA(P)$. 
If $(X,Y) \in\UEA(P)$, or either $Y$ or $A$ is finite, this is trivial. 
So let $(X,Y)\not\in\UEA(P)$ 
and both $Y$ and $A$ be infinite. 
Then $Y_P=Y\cap\atm{P}$ and $X_P=X\cap\atm{P}$ are finite, 
and $(X_P,Y_P)\in\SEA(P)$. 
The latter holds by the observations that 
(i) $Y\models P$ implies $Y_P\models P$;
(ii) for each $Y'\subset Y$ with $(Y'\cap A)=(Y\cap A)$, $Y'\not\models P^Y$ implies that, for each $Y''\subset Y_P$ with $(Y''\cap A)=(Y_P\cap A)$, $Y''\not\models P^Y$; and 
(iii) $X'\models P^Y$ for some $(X'\cap A)=X$ implies that $(X'\cap \atm{P})\models P^Y=P^{Y_P}$.
Moreover, $X_P\subset Y_P$, 
otherwise we end up in a contradiction by the fact that then 
$(X'',X'')\in\UEA(P)=\UEA(Q)$ for some $(X'' \cap A)=X_P$, 
implying 
$(X,Y)\in \SEA(Q)$, since 
$(Y,Y)\in\UEA(Q)=\UEA(P)$ holds.  
Since $Y_P$ is finite, there exists a set $Z_P$, $X_P\subseteq Z_P\subset Y_P$, such 
that $(Z_P,Y_P)\in \UEA(P)$.
Now, let $Z=A\cap (Z_P\cup (Y\setminus Y_P))$. 
Then $X\subseteq Z\subset Y$ holds by 
construction. Furthermore $(Z,Y)\in \UEA(P)$, since 
$Y\setminus Z=Y_P\setminus Z_P$,  
$P^Y=P^{Y_P}$, and $(Z_P,Y_P)\in \UEA(P)$. 
By our assumption $(Z,Y)\in \UEA(Q)$ follows. Contradiction.

\noindent 
Case (2):  
We show the existence of a set $Z$, $X\subseteq Z\subset Y$, such that 
$(Z,Y)\in \UEA(Q)$. 
If $(X,Y) \in\UEA(Q)$, or one of $A$, $Y$ is finite, this is trivial. 
So let $(X,Y)\not\in\UE(Q)$, 
and both $Y$ and $A$ infinite. 
If $(X,Y)\not\in\UEA(Q)$, $((Y\cap A)\setminus X)\subseteq\atm{P}$ must hold;
otherwise we end up in a contradiction by taking any atom 
$a\in (Y\cap A)\setminus X$. 
(Consider $Z=(Y\cap A)\setminus\{a\}$. 
Then $X\subseteq Z\subset Y$ holds by construction and since 
$(Y,Y)\in\UEA(P)=\UEA(Q)$, 
as well as some $Z'$ with $(Z'\cap A)=Z$ models $P^{Z'}=P^{Y}$ 
we get $(Z,Y)\in \SEA(P)$, a contradiction). 
Now, since $\atm{P}$ is finite, 
this means that $(Y\cap A)\setminus X$ is finite, \iec 
there cannot exist an infinite chain of $\SE$-models 
$(X,Y)=(X_0,Y), (X_1,Y), \ldots, (X_i,Y),\ldots$, such that $X_i\subset X_j\subset (Y\cap A)$, 
for $i<j$, and $(X_i,Y)\in\SEA(Q)$. 
Thus, there exists a maximal model $(Z,Y)\in \UEA(Q)$.
By our assumption $(Z,Y)\in \UEA(P)$ follows. Contradiction.
Thus, Theorem~\ref{theo-equivru-2} ($ii$) holds as well, 
proving $P\equivu^A Q$ in Case ($b$).
\end{proof}

\begin{example}
Recall our example programs $Q$ and $Q'$ from above. 
Via the first row in the table (\iec for $A=\{a,b,c\}$,
yielding the respective SE-models), it is easily checked
by Proposition~\ref{theo-equivu-4}
that $Q$ and $Q'$ are uniformly equivalent. 
In fact, the 
SE-model $(\emptyset,abc)$ of $Q'$ 
is not a UE-model of $Q'$, due to the 
presence of the SE-model $(a,abc)$, or alternatively because of $(b,abc)$. 
Note that $Q\equivu Q'$ implies $Q\equivu^A Q'$ for any $A$.
Inspecting the remaining rows in the table, it can be seen that 
for any $A$,  
the sets of $A$-UE-models of $Q$ and $Q'$ are equal, as expected.
\end{example}

We conclude this section, with remarking that
we do not have a directly corresponding result to 
Theorem~\ref{prop-equivu-prog} for relativized uniform equivalence 
(see also next subsection).
A generalization of
Proposition~\ref{prop:subset} is possible, however.
The proof is in the Appendix.

\begin{theorem}\label{thm:subsetUE}
Let $P$ and $Q$ be 
$\DLP$s, 
and $A$ a set of atoms.
Then, 
\begin{enumerate}
\item[$(a)$] $P\equivu^A Q$ implies $\UEA(P)\subseteq \SEA(Q)$ and $\UEA(Q)\subseteq \SEA(P)$;
\item[$(b)$] $\UEA(P)\subseteq \SEA(Q)$ and $\UEA(Q)\subseteq \SEA(P)$ implies $P\equivu^A Q$, 
whenever at least one of $P$, $Q$, or $A$  is finite.
\end{enumerate}
\end{theorem}

\subsection{Properties of Relativized Equivalences}\label{sec:properties}

This section collects a number of properties of 
$A$-SE-models and $A$-UE-models, respectively. 
Note that there are situations where $A$-SE-models and 
$A$-UE-models are the same concepts.

\begin{proposition}\label{prop:a1}
For any program $P$, and a set of atoms $A$ with $\card{A}<2$, 
$\SEA(P)=\UEA(P)$ holds.
\end{proposition}

\begin{corollary}\label{cor:uiss}
For programs $P$,$Q$  and a set of atoms $A$ with $\card{A}<2$, 
$P\equivs^A Q$ iff $P\equivu^A Q$.
\end{corollary}

The following results are only given in terms of $A$-SE-models;
the impact of the results on properties of $A$-UE-models is 
in most cases obvious, and thus not explicitly mentioned.

First, we are able to generalize  
Proposition~\ref{prop:sm} to relativized SE-models.

\begin{lemma}
An interpretation $Y$ is 
an answer set of a program $P$ iff
$(Y,Y)\in\SE^A(P)$ and, for each $X\subset Y$, 
$(X,Y)\not\in\SE^A(P)$.
\end{lemma}

One drawback of $A$-SE-models 
is that they are not closed
under program composition. 
Formally, 
$\SEA(P\cup Q)= \SEA(P)\cap\SEA(Q)$ does not hold in general;
however, it holds whenever $A$ contains all atoms occurring in $P$ or $Q$.
However, 
the fact that, in general, $\SEA(P\cup Q)\neq\SEA(P)\cap\SEA(Q)$,  
is not a surprise, since for $A=\emptyset$, 
$A$-SE-models capture 
answer sets;
and if this closure property would
hold, 
answer set semantics would be monotonic. 

\begin{proposition}\label{prop:notclosed}
For programs $P$, $Q$, and a set of atoms $A$, we have the following relations:
\begin{enumerate}
\item[$(i)$] $(Y,Y)\in \SEA(P)\cap\SEA(Q)$ implies $(Y,Y)\in\SEA(P\cup Q)$;
\item[$(ii)$] for $X\subset Y$, $(X,Y)\in \SEA(P\cup Q)$ 
implies 
$(X,Y)\in\SEA(R)$, whenever $(Y,Y)\in \SEA(R)$, for $R\in \{P,Q\}$;
\item[$(iii)$] the converse directions of $(i)$ and $(ii)$ do not hold in general.
\end{enumerate}
\end{proposition}
\begin{proof}
ad (i): Suppose 
$(Y,Y)\notin\SEA(P\cup Q)$; then either (a) $Y\not\models P\cup Q$; or (b) there
exists a $Y'\subset Y$ with $(Y'\cap A)=(Y\cap A)$, such that $Y'\models (P\cup Q)^Y$. If $Y\not\models P\cup Q$, then either $Y\not\models P$ or $Y\not\models Q$.
Consequently, $(Y,Y)\notin\SEA(P)$ or $(Y,Y)\notin\SEA(Q)$. So, suppose
$Y\models P\cup Q$ and (b) holds. Then neither, $(Y,Y)\in\SEA(P)$ nor
$(Y,Y)\in\SEA(Q)$.

ad (ii): Let $R\in\{P,Q\}$.
Suppose $(Y,Y)\in\SEA(R)$ and $(X,Y)\notin\SEA(R)$. 
The latter implies that no $X'\subset Y$ with $(X'\cap A)=(X\cap A)$, satisfies $X'\models P^Y$. Consequently, no such $X'$ satisfies $X'\models (P\cup Q)^Y$, and thus
$(X,Y)\notin\SEA(P\cup Q)$.

ad (iii):
Take the following example programs. 
Consider programs over $V=\{a,b,c\}$ containing rules
$R=\{ ~ \la \naf a;\; ~ \la \naf b;\; ~ \la \naf c\}$.
Note that $\SE(R)=\{(X,V) \mid X\subseteq V\}$.
Let 
\begin{eqnarray*}
P_a & = & R\cup \{ a\la;\; b\la c;\;c\la b\}; \\
P_b & = & R\cup \{ b\la;\; a\la c;\;c\la a\}; \\
P_c & = & R\cup \{ c\la;\; a\la b;\;b\la a\}.
\end{eqnarray*}
Then, the SE-models 
of $P_v$ are given by $(v,abc)$ and $(abc,abc)$, for $v\in V$.

Set now, for instance, $A=\{c\}$.
Then, we have $\SEA(P_a)=\SEA(P_b)=\{ (\emptyset,abc), (abc,abc) \}$, while
$\SEA(P_c)=\emptyset$.
However,  
$\SEA(P_a\cup P_b)=\SEA(P_a\cup P_c)=\SEA(P_b\cup P_c)=\{(abc,abc)\}$.
This shows that for both, (i) and (ii) in Proposition~\ref{prop:notclosed}, the converse 
direction does not hold.
\end{proof}

The above result crucially influences the behavior
of relativized consequence operators, \iec generalizations
of $\modelse$ as introduced in Definitions~\ref{def:secons} 
and~\ref{def:uecons}, respectively,
to the relativized notions of equivalence.

To check rule redundancy in the context of relatived strong equivalence, 
we give the following result.

\begin{definition}
A rule, $r$, is an $A$-relativized \emph{SE-consequence} 
of a 
program $P$, denoted $P\modelss^A r$, if 
$(X,Y)\in\SEA(\{r\})$,
for all $(X,Y)\in \SEA(P)$.
\end{definition}

\begin{lemma}
For any set of atoms $A$, 
program $P$, and rule $r$ 
with $(B^+(r)\cup H(r))\subseteq A$, it holds that
if $P\modelss^A r$ then 
$P\cup\{r\}\equivs^A P$.
\end{lemma} 

\begin{proof}
We show $\SE^A(P\cup\{r\})=\SE^A(P)$, given $P\modelss^A r$.

\noindent``$\subseteq$'': 
Let $(X,Y)\in\SE^A(P\cup\{r\})$. 
We show $(X,Y)\in\SE^A(P)$. 
First let $X=Y$. Then, $Y\models P\cup\{r\}$ and, for each 
$Y'\subset Y$ with $(Y'\cap A)=(Y\cap A)$, $Y'\not\models (P\cup r)^Y$. 
Since $\atm{r^Y}\subseteq A$, 
for each such $Y'$, $Y'\models r^Y$, and therefore,
$Y'\not\models P^Y$. Consequently, $(Y,Y)\in\SEA(P)$.
So suppose, $X\subset Y$. Then, $(Y,Y)\in\SEA(P\cup\{r\})$. We already know that then $(Y,Y)\in\SEA(P)$. 
We apply Proposition~\ref{prop:notclosed}, and get $(X,Y)\in\SEA(P)$.

\noindent
``$\supseteq$'': 
Let $(X,Y)\in\SE^A(P)$. Then, $(Y,Y)\in\SE^A(P)$ and by assumption 
$(Y,Y)\in\SE^A(\{r\})$. 
By Proposition~\ref{prop:notclosed}, we get $(Y,Y)\in\SE^A(P\cup\{r\})$.
Moreover, from $(X,Y)\in\SE^A(P)$ and $P\modelss^A r$, we get
$(X,Y)\in\SE^A(\{r\})$.
Hence, there exist $X'$, $X''$ with $(X'\cap A)=(X''\cap A)=(X\cap A)$
such that $X'\models P^Y$ and $X''\models r^Y$. By assumption
$\var{r^Y}\subseteq A$. Since $X'$ and $X''$ agree on $A$, we 
get $X'\models r^Y$; and thus $X'\models (P\cup r)^Y$.
Consequently, $(X,Y)\in\SEA(P\cup\{r\})$.
\end{proof}

The result similarly applies to 
the notion of 
UE-consequence relative to $A$, \iec
the restriction $(H(r)\cup B^+(r))\subseteq A$ 
is also necessary in that case. 
However (as in Proposition~\ref{prop-equivu-rule}), the result has to be 
slightly rephrased for 
$A$-UE-models in order to handle 
the case of infinite programs properly.

In general, checking rule-redundancy with respect to relativized 
equivalences is a more involved task; we leave it for further study.

\section{Restricted Classes of Programs}
\label{sec:classes}

So far, we discussed several forms of equivalence for propositional
programs, in general. This section is devoted to two prominent
subclasses of disjunctive logic programs, namely positive and
head-cycle free programs. Notice that these classes include the Horn
logic programs and the disjunction-free logic programs, respectively.

\subsection{Positive Programs}\label{sec:pos}

While for programs with negation, strong equivalence and uniform
equivalence are different, the notions coincide for positive
programs, also in the relativized cases.
We start with some technical results.

\begin{lemma}\label{lemma:totnontot}
Let $P$ be a program, and $A$, $X\subset Y$ be sets of atoms. 
We have the following relations:
\begin{enumerate}
\item If $(Y,Y)\in\SE^A(P)$ and $(X,X)\in\SE^A(P)$, then $((X\cap A),Y)\in\SE^A(P)$.
\item If $(X,Y)\in\SE^A(P)$, then $(Y,Y)\in\SE^A(P)$ and, whenever $P$ is positive, there exists 
an $X'\subseteq Y$ with $(X'\cap A)=X$, such that $(X',X')\in\SE^A(P)$.
\end{enumerate}
\end{lemma}
\begin{proof}
(1)\ First, observe that 
$(X\cap A)\subset (Y\cap A)$ holds.
Otherwise, we get from $X\models P^{X}$, 
$X\models P^Y$ (since $P^Y\subseteq P^{X}$, whenever $X\subseteq Y$),  
and thus $(Y,Y)\notin\SE^A(P)$, by definition.
Moreover, since $X\models P^Y$ and $(Y,Y)\in\SE^A(P)$, we derive
$((X\cap A),Y)\in\SE^A(P)$.

(2)\ Let $(X,Y)\in\SE^A(P)$. Then, $(Y,Y)\in\SE^A(P)$ is an immediate consequence of the definition of $A$-SE-models. From $(X,Y)\in\SE^A(P)$ we get that there exists an $X'\subseteq Y$ with $(X'\cap A)=X$, such 
that $X'\models P^Y$. Take $X'$ as the minimal interpretation satisfying this condition.  For positive $P$, we have $P^{X'}=P^Y=P$ and we get $X'\models P^{X'}=P$.
Moreover, since we chose $X'$ minimal, there does not exist an $X''\subset X'$
with $(X''\cap A)=(X'\cap A)$, such that $X''\models P^{X'}=P$. 
Hence, $(X',X')\in\SE^A(P)$.
\end{proof}

In other words, the set of all $A$-SE-models of a 
positive program $P$ is determined by
its total $A$-SE-models. An important consequence of this result
is the following.

\begin{proposition}\label{prop:postotal}
Let $P$, $Q$ be programs, $P$ be positive, and suppose the
total $A$-SE-models of $P$ and $Q$ coincide. 
Then, $\SE^A(P)\subseteq\SE^A(Q)$.
\end{proposition}
\begin{proof}
Towards a contradiction, assume there exists an $A$-SE-interpretation
satisfying $(X,Y)\in\SE^A(P)$ and $(X,Y)\notin\SE^A(Q)$.
Since $P$ is positive,
by Lemma~\ref{lemma:totnontot} we get that 
there exists some $X'\subseteq Y$ with $(X'\cap A)=X$, such 
that $(X',X')\in\SE^A(P)$. By assumption, the total $A$-SE-models coincide, and
thus we have $(X',X')\in\SE^A(Q)$. 
Moreover, since $(X,Y)\in\SE^A(P)$, we get $(Y,Y)\in\SE^A(P)$ and 
furthermore $(Y,Y)\in\SE^A(Q)$.
Hence, $(X',X')\in\SE^A(Q)$ and $(Y,Y)\in\SE^A(Q)$.
By Lemma~\ref{lemma:totnontot}, 
we get that $((X'\cap A),Y)=(X,Y)$ is $A$-SE-model of $Q$, which is
in contradiction to our assumption.
\end{proof}

From this result, we get that deciding relativized strong and uniform
equivalence of positive programs collapses to checking whether total 
$A$-SE-models coincide.

\begin{theorem}\label{thm:uiss}
Let $P$ and $Q$ be positive $\DLP$s, and $A$ a set of atoms.
The following propositions are equivalent:
\begin{itemize}
\item[(i)] $P\equivs^A Q$;
\item[(ii)] $P\equivu^A Q$;
\item[(iii)] $(Y,Y)\in\SE^A(P)$ iff $(Y,Y)\in\SE^A(Q)$, for each interpretation $Y$.
\end{itemize}
\end{theorem}
\begin{proof}
(i) implies (ii) by definition; (ii) implies (iii) by Theorem~\ref{thm:RUE}.
We show (iii) implies (i).
Applying Proposition~\ref{prop:postotal} in case of two positive programs immediately
yields that (iii) implies $\SE^A(P)=\SE^A(Q)$. Hence,
$P\equivs^A Q$.
\end{proof}

Therefore, RSE and RUE are the same concepts for positive programs; 
we thus sometimes write generically $\equive$
for $\equivs$ and $\equivu$. 

An important consequence of this result, is that $A$-UE-models 
(and thus UE-models) are capable to deal with infinite programs as well, 
provided they are positive. 

\begin{corollary}
Let $A$ be a (possibly infinite) set of atoms, and $P$, $Q$ 
(possibly infinite) positive program. Then, $P\equivu^A Q$ holds
iff $\UE^A(P)=\UE^A(Q)$.
\end{corollary}

\begin{proof}
The only-if direction has already been obtained in Theorem~\ref{thm:RUE}.
For the if direction, note that $\UE^A(P)=\UE^A(Q)$ implies 
(iii) from Theorem~\ref{thm:uiss}, and since $P$ and $Q$ are positive we
derive $P\equivu^A Q$ immediately from that Theorem.
\end{proof}
Concerning strong equivalence and uniform equivalence, Lemma~\ref{lemma:totnontot}
generalizes some well known observations for positive programs.

\begin{proposition}
For any positive program $P$, and sets of atoms $X\subseteq Y$,
$(X,Y)\in\SE(P)$ iff 
$(X,X)\in\SE(P)$ and $(Y,Y)\in\SE(P)$.
\end{proposition}

In other words, the set of all SE-models of a program $P$ is determined by
its total SE-models (\iec by the classical models of $P$).
As known and easy to see from main 
results~\cite{Lifschitz01,Turner01,Turner03}, 
on the class of positive programs
classical and strong equivalence coincide. 
Using Theorem~\ref{thm:uiss}, we can extend this result:

\begin{theorem}
\label{theo:equivu-pos}
For positive programs $P$, $Q$, 
$P\equive Q$ ($e\in\{s,u\}$) iff $P$ and $Q$ have the same classical models.
\end{theorem}

Note that Sagiv~\cite{sagi-88} showed that uniform equivalence of 
DATALOG programs $\Pi$ and $\Pi'$ coincides with 
equivalence of $\Pi'$ and $\Pi$ over Herbrand models; this implies the above result for 
definite Horn programs. Maher~\cite{mahe-88} showed a generalization
of Sagiv's result for definite Horn logic programs with function
symbols.  Furthermore, Maher also pointed out that for
DATALOG programs, this result has been independently established by
Cosmadakis and Kanellakis \cite{cosm-kane-86}.

\begin{example}
Consider the positive programs $P=\{a\vee b\la a;\; b\la a\}$ and 
$Q=\{b\la a\}$. Clearly, $P\models Q$ since $Q\subset P$, but also 
$Q\models P$ holds (note that $b\la a$ is a subclause of $a\vee b\la a$). 
Hence, $P$ and $Q$ are uniformly equivalent, and even strongly equivalent 
(which is also easily verified).
\end{example}

\begin{example}
Consider the positive programs $P=\{a\vee b;\; c\la a;\; c\la b\}$ and 
$Q=\{a\vee b;\; c\}$. Their classical models are $\{a,c\}$,
$\{b,c\}$, and $\{a,b,c\}$. Hence, $P$ and $Q$ are uniformly equivalent, 
and even strongly equivalent (due to Theorem~\ref{thm:uiss}).
\end{example}

Concerning the relativized notions, a result corresponding directly to 
Theorem~\ref{theo:equivu-pos}
is not achievable. However, this is not surprising, otherwise we would
have that 
in case of empty $A$, 
$P\equivs^A Q$ (or $P\equivu^A Q$)  
collapses to classical equivalence. 
This, of course, cannot be the case since for positive programs, 
$P\equiv Q$ 
denotes the 
equivalence of the {\em minimal} 
classical models
of $P$ and $Q$, rather than 
classical equivalence.

Thus, while for strong and uniform equivalence total models 
$(Y,Y)$ for a positive program $P$ coincide with the classical 
models $Y$ of $P$,
the relativized variants 
capture a more 
specific relation, 
viz. minimal models. 
We therefore define as follows.

\begin{definition}
An $A$-minimal model of a program $P$ 
is a classical model $Y$ of $P$, 
such that,
for each $Y'\subset Y$ with
$(Y'\cap A)=(Y\cap A)$,
$Y'$ is not a classical model of $P$.
\end{definition}

Then, we can generalize Theorem~\ref{theo:equivu-pos}
in the following manner:

\begin{theorem}\label{thm:amin}
Let $P$ and $Q$ be positive  $\DLP$s, and $A$ a set of atoms. Then, 
$P\equive^A Q$ ($e\in\{s,u\}$) iff
$P$ and $Q$ have the same $A$-minimal models.
\end{theorem}

\begin{proof}
By Theorem~\ref{thm:uiss} it is sufficient to show 
that the total $A$-SE-models of a program $P$ 
equal its $A$-minimal models. 
This relation holds for positive programs, since $P^Y=P$ for any positive 
program $P$  and any interpretation $Y$. 
In this case 
the conditions for $(Y,Y)\in\SE^A(P)$ are the same as for
$Y$ being $A$-minimal for $P$.
\end{proof}

Note that for $A=\emptyset$ the theorem states that $P\equive^A Q$ iff
the minimal classical models of $P$ and $Q$
coincide, 
reflecting the minimal model semantics
of positive programs. On the other hand,
for $A=U$, the theorem states that $P\equive^A Q$ iff
all classical models of $P$ and $Q$ coincide, as stated above.

\subsection{Head-cycle free programs}

The class of head-cycle free programs generalizes the class of 
normal logic programs
by permitting a restricted form of disjunction. Still, it is capable
of expressing nondeterminism such as, \egc a guess for the value of an atom
$a$, which does not occur in the head of any other rule. For a
definition of head-cycle freeness, we refer to Section~\ref{sec:prelim}.
As shown by Ben-Eliyahu and Dechter~\cite{bene-dech-94}, 
each head-cycle free program can
be rewritten to an ordinary equivalent normal program, 
which is obtained by shifting atoms from the head to the body. 

More formally, let us
define the following notations.

\begin{definition} 
For any rule $r$, let 
\[
r^\ra = 
\left\{
  \begin{array}{lp{2mm}l}
        \big\{ a \la \bodyp{r}, \naf (\bodyn{r} \cup (H(r)\setminus\{a\})) \mid
a\in H(r)\big\}  && \mbox{if\ } \head{r}\neq \emptyset,  \\
        \{r\} && \mbox{otherwise} 
  \end{array}
\right.
\]
For any $\DLP$ $P$, let $P^\ra_r = (P\setminus \{r\})\cup r^\ra$; and
$P^\ra=\bigcup_{r\in P}r^\ra$.
\end{definition}

It is well-known that for any head-cycle free program $P$, it holds
that $P \equiv P^\ra$ (cf.\ \cite{bene-dech-94}). 
This result can be strengthened to uniform equivalence as well as to 
its relativized forms.

\begin{theorem}\label{theo-shift-equivu-1}
For any head-cycle free program $P$, 
and any set of atoms $A$,
it holds that $P \equivu^A P^\ra$.
\end{theorem}

\begin{proof}
For any set of facts $F\subseteq A$, it holds that $(P\cup F)^\ra = P^\ra \cup
F$ and that this program is head-cycle free
iff $P$ is head-cycle free.
Thus, 
$P \cup F \equiv (P\cup F)^\ra \equiv P^\ra \cup
F$. Hence, $P \equivu^A P^\ra$. 
\end{proof}

We emphasize that a similar result for strong equivalence fails, as
shown by the canonical counterexample in
Example~\ref{exa:hcf}. Moreover, the program $P = \{ a \lor b \la.\}$
is not strongly equivalent to any $\NLP$. Thus, we can not conclude
without further consideration that a simple disjunctive ``guessing
clause'' like the one in $P$ (such that $a$ and $b$ do not occur in
other rule heads) can be replaced in a more complex program by the
unstratified clauses $a \la \naf b$ and $b \la\naf a$ (the addition of a
further constraint $~\la a,b$ is required). However, we can conclude
this under uniform equivalence taking standard program splitting
results into account~\cite{lifs-turn-94,eite-etal-97f}.  

The following result 
provides a
characterization of arbitrary programs which are 
relativized 
strongly 
equivalent
to their shift variant. A more detailed  discussion of eliminating disjunction
under different notions of equivalences was recently published in~\cite{Eiter04a}.

First, we state a simple technical result.

\begin{lemma}\label{lemma:subset}
For any rule $r$, $\SE(r)\subseteq \SE(r^\ra)$.
\end{lemma}

\begin{proof}
Indirect. 
Suppose $(X,Y) \in \SE(r)$ and $(X,Y)\notin \SE(r^\ra)$. Then, 
$Y\models r$ and either 
$Y\cap B^-(r)\neq\emptyset$, $X\not \models B^+(r)$, or 
$X\cap H(r)\neq\emptyset$.
By classical logic, $Y\models r$ iff $Y\models r^\ra$.
By assumption $(X,Y)\notin\SE(r^\ra)$, 
there exists a rule in $r^\ra$ with 
$a$ as the only atom in its head, such 
that $a\notin X$, 
$Y\cap B^-(r)=\emptyset$, $X\models B^+(r)$, and 
$Y\cap (H(r)\setminus \{a\})=\emptyset$.
Hence, from the above conditions for $(X,Y)\in\SE(r)$, only
$X\cap H(r)\neq\emptyset$ applies. 
Then, some $b$ from $H(r)$ is contained
in $X$. 
If $a=b$ we get a contradiction to $a\notin X$; otherwise
we get a contradiction to $Y\cap (H(r)\setminus \{a\})=\emptyset$, since
$Y\supseteq X$ and thus $b\in Y$.
\end{proof}

Next, we define the following set, which characterizes the exact 
difference between $r$ and $r^\ra$ in terms of SE-models.

\begin{definition}
For any rule $r$, 
define 
$$
S_r = \{ (X,Y) \mid X\subseteq Y,\; X \models B^+(r),\; Y\cap B^-(r)=\emptyset,\;
\card{H(r)\cap Y}\geq 2,\; H(r)\cap X = \emptyset\}.
$$
\end{definition}

\begin{proposition}\label{prop:kr}
For any disjunctive rule $r$, $\SE(r^\ra)\setminus\SE(r)=S_r$.
\end{proposition}

A proof for 
this result can be found in~\cite{Eiter04a}.
Hence, together 
with Lemma~\ref{lemma:subset}, we get that, for any disjunctive 
rule $r$, $S_r$ characterizes exactly the difference between 
$r$ and $r^\ra$ in terms of SE-models.

\begin{theorem}\label{theo:shift}
Let $P$ be a program, and $r\in P$. Then,
$P\equivs^A P^\ra_r$ iff
for each SE-model $(X,Y)\in \SE(P^\ra_r) \cap S_r$,
exists a $X'\subset Y$, with $X'\neq X$ and $(X'\cap A)=(X\cap A)$,
such that $(X',Y) 
\in\SE(P)$.
\end{theorem}
\begin{proof}
Suppose $P\not\equivs^A P^\ra_r$. 
First, assume there exists an $A$-SE-interpretation
$(Z,Y)\in\SE^A(P)$ such that $(Z,Y)\not\in\SE^A(P^\ra_r)$.
By definition of 
$A$-SE-models, Lemma~\ref{lemma:subset} and the fact that 
$Y\models P$ iff $Y\models P^\ra_r$, we 
get that $Z=Y$.
Since $(Y,Y)\not\in\SE^A(P^\ra_r)$ but $Y\models P^\ra_r$, there 
exists an $X$ such that $(X,Y)$ is SE-model of $P^\ra_r$. Moreover, 
by Proposition~\ref{prop:kr}, $(X,Y)\in S_r$.
On the other hand, from $(Y,Y)\in\SE^A(P)$, we get that,
for each $X'\subset Y$ with $(X'\cap A)=(Y\cap A)=(X\cap A)$, 
$(X',Y)$ is not SE-model of $P$. 
Second, assume there exists an $A$-SE-interpretation
$(Z,Y)\in\SE^A(P^\ra_r)$, such that $(Z,Y)\not\in\SE^A(P)$.
One can verify that using Lemma~\ref{lemma:subset} this implies
$Z\subset Y$. Hence, there exists some $X\subseteq Y$ with 
$(X\cap A)=(Z\cap A)$ such that $(X,Y)$ is SE-model of $P^\ra$ but
no $X'$ with $(X'\cap A)=(X\cap A)$ is SE-model of $P$.
Moreover, $(X,Y)\in S_r$. This shows the claim.
The converse direction is by exactly the same arguments.
\end{proof}
 
As an immediate consequence of this result, we obtain the following 
characterization for general strong equivalence.

\begin{corollary}\label{cor:seshift}
Let $P$ be any $\DLP$. Then, $P\equivs P^\ra$ if and only if for every
disjunctive rule $r \in P$ it holds that $P^\ra$ has no SE-model
$(X,Y)\in S_r$ $($\iec $\SE(P^\ra)\cap S_r=\emptyset$$)$.
\end{corollary}

\begin{example}
Reconsider $P=\{ a\vee b\la  \}$. Then $P^\ra = \{ a\la \naf b, b\la
\naf a \}$ has the SE-model 
$(\emptyset,ab)$
which satisfies the
conditions 
for $S_{a\vee b\la}$.
Note that also the extended program $P'=\{a\vee b\la , a\la b, b\la
a\}$ is not strongly equivalent to its shifted program ${P'}^\ra$. 
Indeed, 
$(\emptyset,ab)$
is also an SE-model of ${P'}^\ra$.
Furthermore, $P'$ is also not uniformly equivalent to ${P'}^\ra$, since 
$(\emptyset,ab)$
is moreover a UE-model of ${P'}^\ra$, but $P'$ has
the single SE-model (and thus UE-model) 
$(ab,ab)$.
\end{example}

We already have seen that shifting is possible if the disjunction is made
exclusive with an additional constraint (see also Example~\ref{exa:excl}).

\begin{example}\label{exa:exclse}
Let $P$ be a program containing the two rules 
$r= a \lor b\la$ and $r'= \la a,b$. 
The rule $r'$ guarantees that no SE-model 
$(X,Y)$ of $P$ or of $P^\ra_r$ with $\{a,b\}\subseteq Y$ exists.
But then, $S_r$ does not contain an element from $\SE(P^\ra_r)$, 
and we get by Corollary~\ref{cor:seshift}, $P\equivs P^\ra_r$.
\end{example}

So far, we have presented a general semantic criterion for 
deciding whether shifting is invariant under~$\equivs^A$.
We close this section, 
with a syntactic criterion 
generalizing the 
concept of
head-cycle freeness.

\begin{definition}\label{def:ahcf}
For a set of atoms $A$,
a rule $r$ is $A$-head-cycle free ($A$-HCF) in a program $P$, iff
the dependency graph of $P$ augmented with the clique over $A$, 
does not contain a cycle going through two atoms from $H(r)$.
A program is $A$-HCF, iff all its rules are $A$-HCF.
\end{definition}

In other words, the considered augmented graph of $P$ 
as used in the definition is given by the pair
$(A\cup\atm{P},E)$ with 
$$
E = \bigcup_{r\in P} \{ (p,q) \mid p\in \bodyp{r}, q\in\head{r}, p\neq q\} \cup \{(p,q),\;(q,p)\mid p,q \in A,\, p\neq q \}
$$
and obviously coincides with the (ordinary) dependency graph of 
the program $P\cup R$, where $R$ is the set of all unary rules over 
$A$. Recall that following Corollary~\ref{cor:unary}, unary rules 
characterize 
relativized strong equivalence sufficiently.
From this observation, 
the forthcoming results follow in a straight-forward manner.

\begin{theorem}\label{thm:relHCF}
For any program $P$, $r\in P$, and a set of atoms $A$,
$P\equivs^A P^\ra_r$, whenever $r$ is $A$-HCF in $P$.
\end{theorem}

Note that if $r$ is $A$-HCF in $P$,  
then $r$ is HCF in $P\cup R$, where $R$ is the set of unary rules over $A$.
In turn, $r$ then is HCF in all programs $P\cup R'$, with
$R'\subseteq R$. Thus, $P\cup R'\equiv P^\ra_r\cup R'$ holds for
all $R'$ by known results. Consequently, $P\equivs^A P^\ra_r$.

\begin{corollary}
For any program $P$, and a set of atoms $A$,
$P\equivs^A P^\ra$ holds, whenever $P$ is $A$-HCF.
\end{corollary}

\section{Computational Complexity} 
\label{sec:complexity}

In this section, we address the computational complexity of checking
various notions of equivalence for logic programs.
We start with uniform equivalence also taking the associated consequence
operator into account. Then, we generalize these results and consider the 
complexity of relativized equivalence. Finally, we consider {\em bounded}
relativization, \iec the problem of deciding $P\equive^A Q$ ($e\in\{s,u\}$), such that the number of atoms 
{\em missing} in $A$ is bounded
by a constant k, 
denoted $P\,{^k}\!\!\equive^A Q$.
For all three groups of problems we provide a fine-grained
picture of their complexity by taking different classes of programs into account.

Recall that $\PiP{2}=\coNP^{\NP}$ is the class of problems such that
the complementary problem is nondeterministically decidable in
polynomial time with the help of an $\NP$ oracle, i.e., in $\SigmaP{2}=\NP^{\NP}$. As well, 
the class $D^P$ 
consists of all problems expressible as the conjunction
of a problem in 
$\NP$ and a problem in $\CONP$.
Moreover, any problem in $D^P$ can
be solved with 
a fixed number of $\NP$-oracle calls,
and is thus intuitively easier than a
problem complete for $\DeltaP{2}$.

Our results are summarized in Table~\ref{tab:newR}.
More precisely, the table shows the complexity of the considered problems 
$P\equivs^A Q$ and $P\equivu^A Q$ in the general case; as well
as in the bounded case ($P\,{^k}\!\!\equivs^A Q$ and 
$P\,{^k}\!\!\equivu^A Q$). Moreover, we explicitly
list the problem of uniform equivalence, $P\equivu Q$.
Depending on the program classes $P$ and $Q$ belong to, the 
corresponding entry shows the complexity 
(in terms of a completness result)
for all five equivalence problems
with respect to these classes. In fact, 
the table has to be read as follows.
For instance, the complexity of equivalence checking for 
$\DLP$s in general is given by the entry in the last line and the
first column 
of Table~\ref{tab:newR}.
The entry's first line refers to the problems 
$P\equivs^A Q$ and $P\equivu^A Q$
(which are both $\PiP{2}$-complete), and the entry's second line
refers to the problems $P\,{^k}\!\!\equivs^A Q$, $P\,{^k}\!\!\equivu^A Q$,
and $P\equivu Q$, respectively. The latter two show $\PiP{2}$-completenes
while 
$P\,{^k}\!\!\equivs^A Q$ 
is $\CONP$-complete.
As another example,
the complexity of deciding equivalence of a head-cycle free program
and a normal program is reported by the entry in the 
second line of the third column. 

We now highlight the most interesting entries of Table~\ref{tab:newR}.

\renewcommand{\arraystretch}{1.2}
\begin{table}
\begin{center}
\begin{tabular}{|l|lllll|}
\hline
$P\equivs^A Q$ / $P \equivu^A Q$ / & & & & &\\
$P\, ^k\!{\equivs^A}\, Q$ / $P\,{^k}\!{\equivu^A}\, Q$ / $P \equivu Q$ 
& $\DLP$ & positive  & HCF & normal & Horn 
\\
\hline
Horn & 
	$\PiP{2}$  & 
	$\CONP$ & 
	$\CONP$ & $\CONP$ & $\CONP$ 
\\
&	$\CONP$ & 
        $\CONP$ & $\CONP$ & $\CONP$ & $\Pol$ \\
\hline
normal & 	
	$\PiP{2}$ 
	& $\PiP{2}$ 
	& $\PiP{2}$/$\CONP$ 
	& $\CONP$ 
& \\
&	$\CONP$ & $\CONP$ &
	$\CONP$ & $\CONP$ & \\
\hline
HCF & 	
	$\PiP{2}$ 
	& $\PiP{2}$ 
	& $\PiP{2}$/$\CONP$ 
& & \\
&	$\CONP$ & $\CONP$ & $\CONP$ & & \\
\hline
positive & 
	$\PiP{2}$ 
	& $\PiP{2}$ 
	 & & &  \\
&	$\CONP$/$\PiP{2}$/$\PiP{2}$ & $\CONP$
	& & &  \\
\hline
$\DLP$ & 
	$\PiP{2}$ & & & &  \\
& $\CONP$/$\PiP{2}$/$\PiP{2}$ &  & & & 
\\
\hline
\end{tabular}
\end{center}
\caption{Complexity of Equivalence Checking in Terms of Completeness Results.}\label{tab:newR}
\end{table}

\begin{itemize}
\item 
(Unrelativized)
uniform equivalence is {\em harder} than (unrelativized) 
strong equivalence; and this result carries over to the case of bounded relativization. This difference
in complexity is only obtained if both programs involved 
contain head-cycles and at least one of them contains default negation.
\item For the case of relativization, 
uniform equivalence is in
some cases {\em easier} to decide than relativized strong equivalence. 
This effect occurs only,
if both programs are head-cycle free, whereby one of them 
may be normal (but not Horn).
\item 
Another interesting case amounts
if two Horn programs are involved. 
Hereby, relativized equivalence is harder than 
in the bounded case, but it
is also harder than 
\emph{ordinary}
equivalence (see Theorem~\ref{thm:horn} in Section~\ref{sec:cre} below).
In each other case, relativization is
never harder than ordinary equivalence. 
\item Finally, we list those cases where bounded relativizations 
decreases
the complexity: As already mentioned for both RSE and RUE, this holds
for comparing Horn programs. 
Additionally, in the case of RSE, there is a proper 
decrease whenever 
one program is disjunctive and the other is not Horn, 
or $P$  contains negation as well as head-cycles and $Q$ is Horn.
In the latter situation, we also observe a descrease in the case 
of RUE.
Additionaly, such a decrease for RUE is present, if 
$P$ is normal or HCF and $Q$ is disjunctive and contains headcycles, or
if two positive $\DLP$s containing headcycles are compared.
\end{itemize}

Some of the effects can be explained by inspecting the underlying
decision problem of model checking. 
For a set of atoms $A$, 
the problem of $A$-SE-model checking (resp. $A$-UE-model checking) is
defined as follows: 
Given sets of atoms $X$, $Y$, and a program
$P$, decide whether $(X,Y)\in\SE^A(P)$ (resp.\ $(X,Y)\in\UE^A(P)$).
We compactly summarize our results on $A$-SE-model checking, 
resp.~$A$-UE-model checking, in Table~\ref{tab:mcnew}. 
This table has to be read as follows. 
The lines determine the class of programs dealed with and the columns
refer to model checking problems in different settings.
From left to right we have: 
(i) bounded $A$-SE-model checking of a program $P$, \iec it is assumed
that $\var{P}\setminus A$ contains a fixed number of atoms;
(ii) the general $A$-SE-model checking problem;
(iii) the special case of $\card{A}=1$, where $A$-SE-model checking and
$A$-UE-model checking coincide;
(iv) the special case of $\card{A}=0$, where both $A$-SE-model checking and
$A$-UE-model checking coincide with answer set checking;
(v) the general $A$-UE-model checking problem;
(vi) bounded $A$-UE-model checking (analogously to bounded $A$-SE-model checking); 
and finally, 
we explicitly list the results for (vii) UE-model checking.
All results from Table~\ref{tab:mcnew}
are proven in detail in the subsequent sections, as well.
All entries except 
the ones
in the first column
are completeness results.
Some interesting observations,  which also intuitively
explain the different results for 
$\equivs^A$ and $\equivu^A$ include:
(1)
$A$-SE-model checking is easier than $A$-UE-model checking in the case
of $\DLP$s and bounded $A$; 
Roughly spoken, in this case the additional test for maximality 
in $A$-UE-model checking is responsible for the higher complexity;
(2)
for the case of head-cycle free programs,
$A$-SE-model checking
is harder than $A$-UE-model checking, viz. $\NP$-complete. 
This result is a consequence of Theorem~\ref{theo:shift}, which 
guarantees that in terms of uniform equivalence, shifted HCF
(and thus normal) programs can be employed; recall that this
simplification is not possible in the context of strong equivalence. 

\begin{table}[t!]
\begin{center}
\begin{tabular}{|l||l|l|l|l|l|l|l|}
\hline
& \multicolumn{2}{|c|}{$A$-SE-models} & 
\multicolumn{2}{}{} &
\multicolumn{2}{|c|}{$A$-UE-models}
& UE-models
\\
& 
$A$ bounded
& general &
$\card{A}=1$ & $A=\emptyset$ & general 
& 
$A$ bounded &
\\
\hline
\DLP/positive 	& in $\Pol$ 	& $D^P$ 	& $D^P$ 	& $\CONP$ 	& $D^P$ 	& $\CONP$ & \CONP \\
HCF 		& in $\Pol$ 	& $\NP$ 	& $\Pol$ 		& $\Pol$ 		& $\Pol$ 		& $\Pol$ & $\Pol$ \\
normal/Horn & in $\Pol$ & $\Pol$ & $\Pol$ & $\Pol$ & $\Pol$ & $\Pol$ & $\Pol$ \\
\hline
\end{tabular}
\end{center}
\caption{Complexity of Model Checking.}\label{tab:mcnew}
\end{table}

Towards showing all results in detail, 
we introduce the following notions used throughout this section.
We often reduce propositional formulas to logic programs using, for a set 
of propositional atoms $V$, an additional set
of atoms $\bar{V}=\{\bar{v}\mid v\in V\}$ within the programs to 
refer to negative literals. Consequently,
we associate to each interpretation $I\subseteq V$, an extended interpretation
$\sigma_V(I) = I\cup \{ \bar{v} \mid v\in V\setminus I\}$, usually dropping 
subscript $V$ if clear from the context. 
The classical models of a formula $\phi$ are denoted by $M_\phi$.
Furthermore, we have a mapping $(\cdot)^*$ defined as 
$v^*=v$, $(\neg v)^*=\bar{v}$, and $(\phi\circ\psi)^*=\phi^*\circ\psi^*$, 
with $v$ an atom, $\phi$ and $\psi$ formulas, and $\circ\in\{\vee,\wedge\}$.
A further mapping $(\overline{\cdot})$ is defined as 
$\overline{v}=\bar{v}$, $\overline{\neg v}=v$, 
$\overline{\phi\vee\psi}=\overline{\phi}\wedge\overline{\psi}$, and 
$\overline{(\phi\wedge\psi)}=\overline{\phi}\vee\overline{\psi}$. 
To use these mappings in logic programs, we 
denote
rules also by 
$a_1\vee \dots \vee a_l \la a_{l+1}\wedge\dots\wedge a_m\wedge\naf
a_{m+1}\wedge\dots\wedge\naf a_n$.

Finally, we define, for a set of atoms $Y\subseteq U$, the following sets of Horn rules.
\begin{eqnarray*}
Y^U_\subseteq & = & \{ \la y \mid y\in U\setminus Y\} \\
Y^U_\subset & = & Y^U_\subseteq \cup \{ \la y_1 \commadots y_n \} \\
Y^U_= & = & Y^U_\subseteq \cup Y
\end{eqnarray*}
Sometimes we do not write the superscript $U$ which refers to the universe.
We assume that, unless stated otherwise, $U$ refers all the atoms
occurring in the programs under consideration.

\subsection{Complexity of Uniform Equivalence}

\begin{table}[t!]
\begin{center}
\begin{tabular}{|l|lllll|}
\hline
$P \equivu Q$ & $\DLP$ & positive & HCF & normal & Horn \\
\hline
Horn & $\CONP$ & $\CONP$ & $\CONP$ & $\CONP$ & $\Pol$ \\
normal & $\CONP$ & $\CONP$ & $\CONP$ & $\CONP$ & \\
HCF & $\CONP$  & $\CONP$ & $\CONP$ & & \\
positive & $\PiP{2}$ & $\CONP$ & & & \\
$\DLP$ & $\PiP{2}$ & & & &\\
\hline
\end{tabular}
\end{center}
\caption{Complexity of Uniform Equivalence in Terms of Completeness Results.}\label{tab:resequivu}
\end{table}

In this section, we address the computational complexity of uniform
equivalence.  While our main interest is with 
the problem of deciding uniform equivalence of two given programs, we 
also consider the related problems of UE-model checking and UE-consequence.
Our complexity results for deciding uniform equivalence of two given 
programs are 
collected from Table~\ref{tab:newR} 
into Table~\ref{tab:resequivu}, for the matter of presentation.
The table has to be read as Table~\ref{tab:newR}. 
Note that in general, uniform equivalence is complete for class
$\PiP{2}$, and therefore 
more complex than deciding strong equivalence, which is
in $\coNP$~\cite{Pearce01,Lin02,Turner03}.  
Thus, the more liberal notion of uniform equivalence
comes at higher computational cost in general. However, for important
classes of programs, it has the same complexity as strong equivalence.

In what follows, we prove all the results in Table~\ref{tab:resequivu}.
Towards these results, we start with the problem of UE-model checking. 
Let $\|\alpha\|$ denote 
the size of an object $\alpha$.

\begin{theorem}
\label{theo-complex-3}
Given a pair of sets $(X,Y)$ and a program  $P$, the problem of 
deciding whether 
$(X,Y)\in\UE(P)$ is ($i$)~$\coNP$-complete in general, and ($ii$)
feasible in polynomial time 
with respect to~$\|P\|+\|X\|+\|Y\|$,  
if $P$ is head-cycle free. Hardness in Case~($i$) 
holds even for positive programs. 
\end{theorem}

\begin{proof}
Testing 
$Y\models P$ and $X\models P^Y$, \iec $(X,Y)\in\SE(P)$, 
for given interpretations $X$, $Y$,
is possible in polynomial time. 
If $X\subset Y$ it remains to check that no $X'$, $X'\models P^Y$, 
exists such that $X\subset X' \subset Y$. 
This can be done via checking
\begin{equation}\label{eq:hcf}
P^Y \cup X \cup Y_\subset \quad \models \quad 
X_=. 
\end{equation}
In fact,
each model, $X'$, of $P^Y \cup X \cup Y_\subset$ gives
a non-total SE-model $(X',Y)$ of $P$ with $X\subseteq X'\subset Y$.
On the other hand, the only model of $X_=$ is $X$ itself. 
Hence, (\ref{eq:hcf}) holds iff no $X'$ with $X\subset X'\subset Y$ exists
such that $(X',Y)\in\SE(P)$, \iec iff $(X,Y)\in\UE(P)$.  
In general, deciding (\ref{eq:hcf}) is in $\CONP$ witnessed by  
the membership part of~(i).

If $P$ is normal then 
the involved 
programs in (\ref{eq:hcf}) 
are Horn and, since 
classical 
consequence
can be decided in 
polynomial time for Horn programs, 
the overall check 
proceeds in polynomial time. 
Finally, if $P$ is head-cycle free, then also $P^Y$ is.
Moreover, by Theorem~\ref{theo-shift-equivu-1} we have $P\equivu P^\ra$.
Hence, in this case, (\ref{eq:hcf}) holds iff 
$(P^\ra)^Y \cup X \cup Y_\subset \models X_=$. 
Since $P^\ra$ is normal, the latter test can be done in polynomial time 
(with respect to $\|P\|+\|X\|+\|Y\|$).
This shows (ii).

It remains to show $\coNP$-hardness of UE-model checking for positive programs.
We show this by a reduction from tautology checking. Let 
$F=\bigvee_{k=1}^m D_k$ be a propositional formula in DNF containing 
literals over 
atoms $X=\{x_1\commadots x_n\}$, and 
consider the following program $P$: 
$$
\begin{array}{rrr@{\,}lr@{\,}ll@{\,}l}
P= & \big\{ & \ x_i\vee \bar{x}_i & \la x_j.     &  x_i\vee \bar{x}_i  & \la  \bar{x}_j.   & \quad \mid & 1\leq i\neq j\leq n\ \big\}\,\cup \\ 
& \big\{ & \ x_i & \la x_j,\bar{x}_j.     &  \bar{x}_i  & \la  x_j, \bar{x}_j.   & \quad \mid & 1\leq i\neq j\leq n\ \big\}\,\cup \\ 
    & \big\{  &              x_i & \la D^{*}_k. &           \bar{x}_i & \la
                    D^{*}_k. & \quad \mid \quad  & 1\leq k\leq m, 1\leq i\leq n\ \big\},
\end{array}
$$
where $D^{*}_k$ results from $D_k$ by replacing 
literals $\neg x_i$ by $\bar{x}_i$. 

Since $P$ is positive, the SE-models of $P$ are determined by its classical models, which are given by $\emptyset$, $X\cup \bar{X}$, and 
$\sigma(I)$, for each interpretation  $I\subseteq X$ making $F$ false.
Hence,
$(\emptyset,X\cup \bar{X})$ 
is an SE-model of 
$P$ and $(\emptyset,X\cup \bar{X})\in \UE(P)$ iff $F$ is a tautology. 
This proves $\coNP$-hardness.
\end{proof}

In fact, also those UE-model checking problems which are feasible
in polynomial time, are hard for the class $\Pol$.

\begin{theorem}\label{thm:phard}
Given a pair of sets $(X,Y)$ and a head-cycle free 
program $P$, the problem of 
deciding whether 
$(X,Y)\in\UE(P)$ is $\Pol$-complete.
Hardness holds, even if $P$ is definite Horn.
\end{theorem}
\begin{proof}
Membership has already been shown in Theorem~\ref{theo-complex-3}.
We show hardness via a reduction from the $\Pol$-complete problem 
HORNSAT to UE-model checking for Horn programs.
Hence, let $\phi=\phi_f\AND\phi_r\AND\phi_c$ a Horn formula over atoms $V$, 
where
$\phi_f =  a_1 \wedge\cdots\wedge a_n$; 
$\phi_r = \bigwedge_{i=1}^m (b_{i,1}\wedge\cdots\wedge b_{i,k_i} \rightarrow b_i)$; 
and
$\phi_c = \bigwedge_{i=1}^l \neg (c_{i,1}\wedge\cdots\wedge c_{i,k_i})$.
Wlog suppose $n\geq 1$ (otherwise $\phi$ would be trivially satisfiable by
the empty interpretation).
Let $u,w$ be new atoms, and take the program
\begin{eqnarray*}
P & = & \{ a_i \la u \mid 1\leq i \leq n \} \cup \\
& & \{ b_i \la b_{i,1}\commadots b_{i,k_i} \mid 1\leq i\leq m \}\cup \\
& & \{ w \la c_{i,1}\commadots c_{i,k_i} \mid 1\leq i \leq l \}\cup \\
& & \{ u \la v;\; v\la w \mid v\in V\} \cup \{u\la w\}. 
\end{eqnarray*}
We show that $\phi$ is unsatisfiable iff $(\emptyset,V\cup\{u,w\})$ is 
UE-model of $P$. Note that both $\emptyset$ and 
$V\cup\{u,w\}$ are classical models of $P$ for any $\phi$.
Since $P$ is positive, it is sufficient to show that $\phi$ is satisfiable
iff a model $M$ of $P$ exists, 
such that $\emptyset\subset M \subset (V\cup\{u,w\})$.

Suppose $\phi$ is satisfiable, and $M$ is a model of $\phi$; then 
it is easily checked that
$M\cup \{u\}$ is a model of $P$.
So suppose $\phi$ is unsatisfiable, and towards a contradiction let 
some $M$ with $\emptyset\subset M \subset (V\cup\{u,w\})$ be a model of $P$.
From the rules $\{v\la w \mid v\in V\} \cup \{u\la w\}$, we get $w\notin M$.
Hence, the constraints $\phi_c$ are true under $M$. 
Since $M$ is not empty, either $u\in M$ or some $v\in V$ is in $M$.
However, the latter implies that $u\in M$ as well 
(by rules $\{ u \la v\mid  v\in V\}$).
Recall that $\phi_f$ is not
empty by assumption, hence all $a_i$'s from $\phi_f$ are in $M$.
Then, it is easy to see that $M\setminus\{u\}$ satisfies $\phi$, 
which contradicts our assumption that $\phi$ is unsatisfiable. 
\end{proof}

We now consider the problem of our main interest, namely deciding
uniform equivalence. By the previous theorem, the following upper
bound on the complexity of this problem is obtained. 

\begin{lemma}\label{lemma-complex-1}
Given two $\DLP$s $P$ and $Q$, deciding whether $P \equivu Q$ is 
in the class $\PiP{2}$.
\end{lemma}

\begin{proof}
To show that two $\DLP$s $P$ and $Q$ are not uniformly equivalent, we
can by Theorem~\ref{theo-equivu-4} guess an SE-model $(X,Y)$ such that
$(X,Y)$ is an UE-model of exactly one of the programs $P$ and $Q$. By
Theorem~\ref{theo-complex-3}, the guess for $(X,Y)$ can be verified in
polynomial time with the help of an $\NP$ oracle. This proves
$\PiP{2}$-membership 
of $P \equivu Q$.
\end{proof}

This upper bound has a complementary lower bound 
proved in the following result. 

\begin{theorem}\label{theo-complex-2}
Given two $\DLP$s $P$ and $Q$, deciding whether $P \equivu Q$ is $\PiP{2}$-complete. Hardness holds even if one of the programs is positive.
\end{theorem}

\begin{proof}
Membership in $\PiP{2}$ has already been established in Lemma~\ref{lemma-complex-1}. 
To show $\PiP{2}$-hardness, we provide a polynomial reduction of
evaluating a quantified Boolean formula (QBF) from a fragment which is
known $\PiP{2}$-complete to deciding uniform equivalence of two
$\DLP$s $P$ and $Q$. 

Consider a $\mathit{QBF}_{2,\forall}$  of form 
$F=\forall X\exists Y \phi$ 
with $\phi= \bigwedge_{i=1}^{i=m} C_i$, where each 
$C_i$ is a 
disjunction of literals over the boolean variables in $X\cup Y$.
Deciding whether a given such $F$ is true is well known to be 
$\PiP{2}$-complete. 

For the moment, let us assume that $X=\emptyset$, i.e., the QBF
amounts to a SAT-instance $F$ over $Y$. 
More precisely,
in what follows we reduce the satisfiability problem of 
the quantifier-free formula $\phi$
to the problem of deciding uniform equivalence of
two programs $P$ and $Q$.
Afterwards, we take the entire QBF $F$ into account.

Let $a$ and $b$ 
be fresh atoms
and define 
\begin{eqnarray}
P & = & \{ 
y \lor \bar{y} \la \mid y\in Y \} \cup \label{prog:1} \\
& & \{ b \la y, \bar{y};\;
y \la b;\; \bar{y} \la b \mid y\in Y \} \cup \label{prog:3}\\
& & \{ b \la \overline{C}_i \mid 1\leq i \leq m \} \cup \label{prog:4} \\
& & \{ a \la \}. \label{prog:5}
\end{eqnarray}
Note that $P$ is positive.
The second program is defined as follows:
\begin{eqnarray}
Q & = & \{ 
y \lor \bar{y} \la z \mid y\in Y;\; z\in Y\cup\bar{Y}\cup\{a\} \} 
\cup \label{prog:1q} \\
& & \{ b \la y, \bar{y};\; 
y \la b;\; \bar{y} \la b \mid y\in Y \} 
\cup \label{prog:3q}\\
& & \{ b \la \overline{C}_i \mid 1\leq i \leq m \} \cup \label{prog:4q} \\
& & \{ a \la b;\; a\la\naf b;\; a\la \naf a \}. \label{prog:5q}
\end{eqnarray}
The only differences 
between the two programs $P$ and $Q$ are located in the 
rules (\ref{prog:1}) compared to (\ref{prog:1q}) 
as well as (\ref{prog:5}) compared to (\ref{prog:5q}).
Note that (\ref{prog:5q}) also contains default negation.

Let us first compute the SE-models of $P$. Since $P$ is positive
it is sufficient to consider classical models.  Let $\at=Y\cup \bar{Y}\cup\{a,b\}$.
First, $\at$ is clearly a classical model of $P$, and so
is 
$\sigma(I)\cup \{a\}$, for each classical model $I\in M_\phi$.
In fact, these are the only models of $P$.
This can be seen as follows. 
By rules (\ref{prog:1}), at least one $y$ or $\bar{y}$ must be
contained in a model, for each $y\in Y$.
By (\ref{prog:3}), if both $y$ and $\bar{y}$ are contained
in a candidate-model for some $y\in Y$ or $b$ is contained in the candidate,
then the candidate is spoiled up to $Y\cup \bar{Y}\cup \{b\}$.
Hence the classical models of (\ref{prog:1}--\ref{prog:3}) are 
given by $\{ \sigma(I) \mid I\subseteq Y\}$ and $Y\cup \bar{Y}\cup \{b\}$.
Now, (\ref{prog:4}) 
eliminates those candidates which make $\phi$ false by ``lifting'' them 
to $Y\cup \bar{Y}\cup \{b\}$.
By (\ref{prog:5}) we finally have to add $a$ to the remaining candidates.

Hence, the SE-models of $P$ are given by
$$
\{ (\sigma(I)\cup\{a\},\sigma(I)\cup\{a\}) \mid I\in M_\phi\}\, \cup\,
\{ (\sigma(I)\cup \{a\},\at) \mid I\in M_\phi\} \,\cup \,
(\at,\at).
$$
Obviously, each SE-model of $P$ is also UE-model 
of $P$.

We now analyze $Q$. First observe that the classical models of $P$ and
$Q$ coincide. 
This is due the fact that 
(\ref{prog:5}) is classically equivalent to (\ref{prog:5q}) and thus classically
derives $a$, making 
(\ref{prog:1q}) and (\ref{prog:1}) do the same job in this context.
However, since $Q$ is not positive we have to consider
the respective reducts of $Q$ to compute the SE-models.
We start with SE-models of the form $(X,\at)$. 
In fact, $(X,\at)\in\SE(Q)$ iff $X\in \{\emptyset,\at\} \cup 
\{ \sigma(I)\mid I\in M_\phi\} \cup \{ \sigma(I)\cup\{a\}\mid I\in M_\phi\}$.
The remaining SE-models of $Q$ are all total and, as for $P$, given by 
$\{(\sigma(I)\cup\{a\},\sigma(I)\cup\{a\})\mid I\in M_\phi\}$.

Hence, the set of all SE-models of $Q$ is
\begin{eqnarray*}
& & \{ (\sigma(I)\cup\{a\},\sigma(I)\cup\{a\}) \mid I\in M_\phi\} \;\cup\;
\{ (\sigma(I)\cup \{a\},\at) \mid I\in M_\phi\} \;\cup\;  (\at,\at)\;\cup \\
& & \{ (\sigma(I),\at) \mid I\in M_\phi\} \;\cup\; (\emptyset,\at);
\end{eqnarray*}
having additional SE-models compared to $P$, namely $(\emptyset,\at)$ and 
$\{ (\sigma(I),\at) \mid I\in M_\phi\}$. Note however, that the 
latter SE-models 
never are UE-models of $Q$, since clearly $\sigma(I)\subset(\sigma(I)\cup \{a\})$, for all $I\in M_\phi$.

Thus, if $M_\phi$ is not empty, 
the UE-models of $P$ and $Q$ coincide; 
otherwise there is a single non-total UE-model of $Q$,
namely $(\emptyset,\at)$. 
Note that the latter is not UE-model of $Q$ in the
case $M_\phi\neq\emptyset$ since, 
for each $I\in M_\phi$, $\sigma(I)\neq \emptyset$. 
Consequently, 
the UE-models of $P$ and $Q$ coincide iff $M_\phi$ is not empty, \iec
iff $\phi$ is satisfiable.

So far we have shown how to construct  programs $P$ and $Q$,
such that uniform equivalence 
encodes SAT. 
To complete the reduction for the QBF, we now also take $X$ into
account.

We add in both $P$ and $Q$ the set of rules
$$
\{ x \lor \bar{x} \la;\; \la x, \bar{x}  \mid  x\in X\}
$$
where the $\bar{x}$'s are fresh atoms.
The set $\at$ remains as before, \iec without any 
atom of the form $x$ or $\bar{x}$.

This has the following effects. 
First the classical models of both $P$ and $Q$ are now given by
$\sigma_{X\cup Y}(I)\cup \{a\}$, for each $I\in M_\phi$, 
and 
$(\sigma_{X\cup Y}(J) \cup \at)=(\sigma_{X}(J)\cup\at)$, for each $J\subseteq X$. 
Therefore, the SE-models of 
$P$ are given by 
\begin{eqnarray}
& & \{ (\sigma_{X\cup Y}(I)\cup\{a\}, \sigma_{X\cup Y}(I)\cup\{a\}) \mid I\in M_\phi\} \cup \label{set:1} \\
& & \{ (\sigma_{X\cup Y}(I)\cup\{a\}, \sigma_{X}(I)\cup\at) \mid I\in M_\phi \} \cup \label{set:2} \\
& & \{ (\sigma_{X}(J)\cup\at, \sigma_{X}(J)\cup\at) \mid J\subseteq X\}. \label{set:3}
\end{eqnarray}
Again, each SE-model of $P$ is also UE-model of $P$.
For $Q$ the argumentation from above is used analogously. 
In particular, for each $J\subseteq X$, we get an additional SE-model
$\{ (\sigma_X(J), \sigma_X(J)\cup\at) \}$ for $Q$.
Thus, the UE-models of $P$ and $Q$ coincide iff, none of these additional 
SE-models $\{ (\sigma_X(J), \sigma_X(J)\cup\at) \}$ of $Q$ is an UE-model 
of $Q$, as well.
This is the case iff, for each $J\subseteq X$,  
there exists a truth assignment to $Y$ making $\phi$ true, \iec
iff the QBF $\forall X\exists Y\phi$ is true.

Since $P$ and
$Q$ are obviously constructible in polynomial time, our result follows.
\end{proof}

For the construction of $P$ and $Q$ in above proof we used---for matters
of presentation---two additional atoms $a$ and $b$. 
However, one can resign on $b$; 
by replacing  rules
(\ref{prog:3}) and (\ref{prog:4}) in both programs by 
$\{ y\la\overline{C};\; \bar{y}\la\overline{C}_i \mid y\in Y;\; 1\leq i \leq m\}$;
and additionally rules (\ref{prog:5q}) in $Q$ by
$\{ a\la\overline{C};\; \bar{a}\la\overline{C}_i \mid 1\leq i \leq m\} \cup 
\{\la\naf a\}$.
Hence, already a single occurrence of default negation in one 
of the compared programs makes the
problem harder. Note that
equivalence of two positive disjunctive programs is among the
$\CONP$-problems discussed in the following.

\begin{theorem}\label{theo-pos-new}
Let $P$ and $Q$ be positive $\DLP$s.
Then, deciding whether $P \equivu Q$ is $\coNP$- complete, 
where $\coNP$-hardness holds even if one of the programs is Horn.
\end{theorem}
\begin{proof}
By Theorem~\ref{thm:uiss}, 
uniform equivalence and strong equivalence are the same concepts 
for positive programs. 
Since strong equivalence is in $\CONP$ in general, 
the membership part of the theorem follows immediately. 

We show $\coNP$-hardness for a positive $\DLP$ $P$ 
and a Horn program $Q$ by a reduction from UNSAT. 
Given a propositional
formula in CNF $F=\bigwedge_{i=1}^{m}C_i$ 
over 
atoms $X$, 
let 
\begin{eqnarray*}
P 
& = & 
\{ C^*_i \vee a\la \; \mid 1\leq i\leq m\}
\cup 
\{ \la x,\bar{x} \mid x\in X\};\quad\mbox{and}
\\
Q  &  =  & 
\{ a\la \} \cup \{ \la x,\bar{x} \mid x \in X\}.
\end{eqnarray*}
By Theorem~\ref{theo:equivu-pos}, 
$P\equivu Q$ iff $P$ and $Q$ have the same classical models. 
The latter holds iff 
each model of $P$ contains the atom $a$. 
But then, 
$F$ is unsatisfiable.
\end{proof}

We now turn to head-cycle free programs.

\begin{theorem}\label{theo-I}
Let $P$ and $Q$ be $\DLP$s, 
and $P$ head-cycle free. 
Then, deciding 
$P \equivu Q$ is $\coNP$- complete, 
where $\coNP$-hardness holds even if $P$ is normal and $Q$ is Horn.
\end{theorem}

\begin{proof}
For the membership part, by Theorem~\ref{prop-equivu-prog}, 
$P \equivu Q$ iff $P\modelsu Q$ and $Q\modelsu P$. 
Both tasks are in $\CONP$ (see Theorem~\ref{thm:consneu} below).
Since the
class $\coNP$ is closed under conjunction, it follows that deciding
$P \equivu Q$ is in $\coNP$.

To show $\CONP$-hardness consider the programs from the proof
of Theorem~\ref{theo-pos-new}. 
Indeed, $P$ is HCF and, therefore, $P\equivu P^\ra$ by
Theorem~\ref{theo-shift-equivu-1}.
Using the same argumentation as above,
yields $P^\ra \equivu Q$ iff $F$ is unsatisfiable. 
This shows the $\CONP$-hardness result for comparing normal and Horn programs.
\end{proof}

Note that Sagiv showed~\cite{sagi-88} that deciding $P \equivu Q$ for
given definite Horn programs $P$ and $Q$ is polynomial, which easily follows
from his result that the property of uniform
containment (whether the least model of $P\cup R$ is always a
subset of $Q\cup R$) can be decided  in polynomial time. 
As pointed out by Maher \cite{mahe-88}, Buntine \cite{bunt-88} has
like Sagiv  provided an algorithm for deciding uniform containment.

Sagiv's result clearly generalizes to arbitrary Horn programs, since
by Theorem~\ref{theo:equivu-pos}, deciding $P \equivu Q$ reduces to
checking classical equivalence of Horn theories, which is known to be
$\Pol$-complete.

\begin{corollary}
Deciding uniform equivalence of Horn programs is 
$\Pol$-complete.
\end{corollary}

This concludes our analysis on the complexity of checking uniform 
equivalence. Our results cover all possible combinations of the classes
of programs considered, \iec 
$\DLP$s, positive programs, normal programs, head-cycle free programs, 
as well as 
Horn programs, as already highlighted in Table~\ref{tab:resequivu}.

Finally, we complement the results on uniform equivalence and UE-model
checking with 
addressing the complexity of UE-consequence.
The proofs of these results can be found in the Appendix.

\begin{theorem}
\label{theo:consequence}
Given a $\DLP$ $P$ and a rule $r$, deciding $P \modelsu r$ is ($i$)
$\PiP{2}$-complete in general, ($ii$) $\coNP$-complete if $P$ is
either positive or head-cycle free, and ($iii$) polynomial 
if $P$ is Horn. 
\end{theorem}

\begin{theorem}
\label{thm:consneu}
Let $P$, $Q$ be $\DLP$s. Then, $P\modelsu Q$ is $\CONP$-complete, whenever
one of the programs is head-cycle free. 
$\CONP$-hardness holds, even if $P$ is normal and $Q$ is Horn.  
\end{theorem}

\subsection{Complexity of Relativized Equivalence}\label{sec:cre}

\begin{table}[t!]
\begin{center}
\begin{tabular}{|l|lllll|}
\hline
$P \equivs^A Q$ / $P \equivu^A Q$ & $\DLP$ & positive & HCF & normal & Horn \\
\hline
Horn & $\PiP{2}$ & $\CONP$ & $\CONP$ & $\CONP$ & $\CONP$ 
\\
normal & $\PiP{2}$  & $\PiP{2}$ & $\PiP{2}$/$\CONP$ & $\CONP$ & \\
HCF & $\PiP{2}$ & $\PiP{2}$ & $\PiP{2}$/$\CONP$ & & \\
positive & $\PiP{2}$ & $\PiP{2}$ & & & \\
$\DLP$ & $\PiP{2}$ & & & &\\
\hline
\end{tabular}
\end{center}
\caption{Complexity of Relativized 
Equivalences 
in Terms of Completeness Results.}\label{tab:resnew}
\end{table}

We now generalize the complexity results to 
relativized forms of equivalence.
In particular, we investigate the complexity of 
$A$-SE/UE-model checking as well as of the equivalence 
problems $\equivs^A$ and $\equivu^A$, respectively.
Like in the previous section, we also consider different classes of
programs. Our results are summarized in Table~\ref{tab:resnew} for both
RSE and RUE  
at a glance by just highlighting where the complexity differs.
Note that the only differences between 
RSE and RUE
stem from the entries $\PiP{2}/\CONP$ in the column for head-cycle
free programs.
Here we have that in the cases HCF/HCF and HCF/normal, checking $\equivs^A$ is 
in general
harder for RSE than for RUE.
Another issue to mention is that already for uniform equivalence, the 
concept of relativization make things more difficult. One just needs
to compare the first two columns of Tables~\ref{tab:resequivu} 
and~\ref{tab:resnew}, respectively. Even worse for strong equivalence, 
which is in $\CONP$ in its unrelativized version and now jumps up to 
$\PiP{2}$-completeness in several cases.
Finally,
also the comparison of two Horn programs
becomes intractable, viz.\ $\CONP$-complete, compared to the 
polynomial-time result in the cases of unrelativized 
strong and uniform equivalence.

To summarize, RSE and RUE 
(i) are harder to decide than in their unrelativized versions in several 
cases, and 
(ii)
both
are generally of the same complexity except 
head-cycle free programs are involved.
Note that Observation~(ii),
on the one hand, contrasts the current view that
notions of strong equivalence have milder complexity than 
notions like uniform equivalence.
On the other hand, the intuition behind this gap becomes apparent if one 
takes into account that for HCF programs $P$, 
$P\equivu^A P^\ra$ holds, while $P\equivs^A P^\ra$ does not.

For an even more fine-grained picture, note that problems associated
with equivalence tests 
relative to an atom set $A$ 
call for further
distinctions between several cases concerning the concrete
instance $A$.  We identify the following ones:
\begin{itemize}
\item $\card{A}=0$: In this case, both $A$-SE and $A$-UE-model checking 
collapse to 
answer set checking; correspondingly, 
RSE and RUE collapse to ordinary equivalence;
\item $\card{A}<2$: By Proposition~\ref{prop:a1} and Corollary~\ref{cor:uiss},
$A$-SE-models and $A$-UE-models coincide, and thus, 
RSE and RUE are the same concepts.
\end{itemize}

Our results 
for $\equive^A$, $e\in\{s,u\}$,
given in the following, 
consider arbitrary 
fixed $A$ unless stated otherwise.
Moreover, we consider that $A$ contains only atoms which also occur
in the programs under consideration.
In some cases the hardness-part of the complexity results is 
obtained only if $\card{A}>k$ for some constant $k$. We shall make these 
cases explicit. 

Another special case for $A$ is 
to consider {\em bounded relativization}. 
This denotes the class of problems 
where the cardinality of $(V\setminus A)$ is less or equal than a fixed
constant $k$, with $V$ being the atoms occurring 
in the two programs compared. 
Note that this concepts contains strong and uniform equivalence, respectively, 
as special cases, \iec if $(V\setminus A)=\emptyset$. 
We deal with bounded relativization explicitly in the subsequent section.

Towards deriving the results from Table~\ref{tab:resnew}, 
we first consider
model checking problems.
Formally, 
for a set of atoms $A$, 
the problem of $A$-SE-model checking (resp. $A$-UE-model checking) is 
defined as follows: Given sets of atoms $X$, $Y$, and a program
$P$, decide whether $(X,Y)\in\SE^A(P)$ (resp.\ $(X,Y)\in\UE^A(P)$).
We start with the following tractable cases. 

\begin{theorem}\label{thm:reltr}
Given a pair of sets $(X,Y)$, a set of atoms $A$,
and a program $P$,
the problem of
deciding whether
$(X,Y)\in\SE^A(P)$ $($resp. $(X,Y)\in\UE^A(P))$ is 
feasible in polynomial time 
with respect to~$\|P\|+\|X\|+\|Y\|$,  
whenever $P$ is normal $($resp. whenever $P$ is HCF$)$.
\end{theorem}

\begin{proof}
We start with the test whether
$(X,Y)$ is $A$-SE-model of a normal program $P$.
Note that $P^Y$ is Horn, 
and that $Y$ is a model of $P^Y$ iff $Y$ is a model of $P$.
Consider the following algorithm
\begin{enumerate}
\item Check whether $Y$ is a model of $P^Y$.
\item Check whether 
$P_Y=
P^Y \cup (Y\cap A) \cup Y_\subset
$
is unsatisfiable. 
\item 
If $X\subset Y$,
check whether $P_X= 
P^Y \cup (X\cap A) \cup 
\{ \la x \mid x\in (A\setminus X) \} 
\cup Y_\subseteq$
is satisfiable.
\end{enumerate}
Note that each step is feasible in polynomial time, especially since 
both $P_X$ and $P_Y$ are Horn. Hence, it remains to 
proof that 
above algorithm holds, exactly if $(X,Y)$ is $A$-SE-model of $P$.
This is seen as follows:
each step exactly coincides with 
one of the conditions of checking whether $(X,Y)$ is an $A$-SE-model, \iec 
(1) $Y\models P$;  
(2) for all $Y'\subset Y$ with $(Y'\cap A)=(Y\cap A)$, 
$Y'\not\models P^Y$; 
and (3) $X\subset Y$ implies existence of a 
$X'\subseteq Y$ with $(X'\cap A)=X$, such that $X'\models P^Y$.

For the result on $A$-UE-model checking we use a similar argumentation. 
First suppose that $P$ is normal and consider the algorithm from above
but replacing the second step by
\begin{enumerate}
\item[2a.] Check whether 
$P^Y \cup (X\cap A) \cup Y_\subset
\models (X\cap A) \cup \{ \la x \mid x\in (A\setminus X) \}$.
\end{enumerate}
The desired algorithm then corresponds to 
the respective conditions for $A$-UE-model checking following
Proposition~\ref{prop:rel}.
To be more specific, the models of $P^Y \cup (X\cap A) \cup Y_\subset$ 
are those $X'$ with $(X\cap A) \subseteq X' \subset Y$ such that 
$X'\models P^Y$. The set of models of the right-hand side is 
given by $\{ Z \mid (Z\cap A) = (X\cap A)\}$. Hence, 
the test in [2a.] is violated iff there exists an 
$X'$ with $(X\cap A) \subset (X'\cap A)$ and $X' \subset Y$ such that
$X'\models P^Y$, \iec iff $(X,Y)\notin\UE^A(P)$.
Moreover,
for HCF programs, $P^\ra$ is $A$-UE-equivalent to $P$, following Theorem~\ref{theo-shift-equivu-1}, \iec the $A$-UE-models for $P$ and $P^\ra$ coincide.
Applying $P^\ra$ to the presented
procedure 
thus shows that
$A$-UE-model checking
is feasible in polynomial time 
also for HCF programs.
\end{proof}

Without a formal proof, we mention that these tractable model checking
problems are complete for the class $\Pol$. Indeed, one can 
re-use the argumentation from the proof of Theorem~\ref{thm:phard} and
take, for instance, $A=\{u\}$. 
Then, $(\emptyset,V\cup\{u,w\})\in\SE^A(P)=\UE^A(P)$ iff
the encoded Horn formula is satisfiable. Note that $\Pol$-hardness holds
also for answer set checking (\iec $A=\emptyset$) by the straightforward
observation that a Horn program $P$ has an answer set iff $P$ is 
satisfiable.

Next, we consider the case of $A$-SE-model checking for head-cycle free programs.
Recall that for $\card{A}< 2$, $A$-SE-model checking 
coincides with $A$-UE-model checking, and thus in these cases
$A$-SE-model checking is feasible in polynomial time, as well. 
However, in general, $A$-SE-model checking is harder than $A$-UE-model checking
for head-cycle free programs.

\begin{theorem}\label{thm:rsehcf}
Let $(X,Y)$ be
a pair of interpretations, and $P$ a head-cycle free program.
Deciding whether
$(X,Y)\in\SE^A(P)$ is $\NP$-complete.
Hardness holds 
for any fixed $A$ with $\card{A}\geq 2$.
\end{theorem}

\begin{proof}
For the membership result we argue as follows.
First we check whether $(Y,Y)\in\SE^A(P)$. Note that 
$(Y,Y)\in\SE^A(P)$ iff $(Y,Y)\in\UE^A(P)$. By Theorem~\ref{thm:reltr} the
latter test
is feasible in polynomial time. 
It remains to check whether there exists a $X'\subseteq Y$ with $(X'\cap A)=X$, 
such that $X'\models P^Y$. This task is in $\NP$, and therefore, 
the entire test is in $\NP$.

For the corresponding $\NP$-hardness, consider the  
problem of checking satisfiability of a 
formula $\psi=\bigwedge_{j=1}^m C_j$ in CNF 
given over a set of atoms $V$. 
This problem is $\NP$-complete.
We reduce 
it to $A$-SE-model checking for a HCF program.
Consider the following program with additional atoms $a_1$, $a_2$, 
$\bar{V}= \{\bar{v}\mid v\in V\}$, and let $A=\{a_1, a_2\}$. 
\begin{eqnarray}
P & = & \{ v \OR \bar{v} \la \mid v\in V \} 
\label{eq:NP1}
\\
& & \{ v \la a_1;\; \bar{v}\la a_1\mid v\in V\}
\label{eq:NP2} \\
& & \{ a_2 \la \overline{C}_j \mid 1\leq j \leq m \} \label{eq:NP3}\\
& & \{ a_2 \la  v, \bar{v}  \mid v\in V \}\label{eq:NP4}.
\end{eqnarray}
Note that $P$ is HCF.
Let $Y=V\cup \bar{V}\cup A$.
We show that $(\emptyset,Y)\in\SE^A(P)$
iff $\psi$ is satisfiable.
It is clear that $Y\models P$ and no 
$Y'\subset Y$ with 
$(Y'\cap A)=(Y\cap A)$ satisfies $Y'\models P^Y=P$ due to 
Rules~(\ref{eq:NP2}). This shows $(Y,Y)\in\SE^A(P)$.
Now, $(\emptyset,Y)\in\SE^A(P)$ iff there exists a 
$X\subseteq (V\cup \bar{V})$ such that $X\models P^Y=P$. 
Suppose $X\models P$.
Since $a_2\notin X$, $X$ must represent a consistent guess
due to Rules~(\ref{eq:NP1}) and~(\ref{eq:NP4}). 
Moreover, $X$ has to represent a model of $\psi$ due to Rules~(\ref{eq:NP3}).
Finally,
$X\models$~(\ref{eq:NP2}) holds by trivial means, \iec since $a_1\notin X$.
The converse direction is by analogous arguments.
Hence, $(\emptyset,Y)\in\SE^A(P)$ iff there exists 
a model of $\psi$, \iec iff $\psi$ is satisfiable.

This shows hardness for $\card{A}=2$. To obtain $\CONP$-hardness
for any $A$ with $k=\card{A}>2$ and, such that all $a\in A$ are also
occurring in the program, 
consider $P$ as above augmented by rules 
$\{ a_{i+1} \la a_i \mid 2\leq i < k \}$ and 
$A=\{ a_i \mid 1\leq i \leq k\}$.
By analogous arguments as above,
one can show that then 
$(\emptyset,(V\cup \bar{V}\cup A))\in\SE^A(P)$ iff 
$\psi$ is satisfiable.
\end{proof}

The next result concerns $A$-SE-model checking and $A$-UE-model checking 
of disjunctive logic programs in general and positive $\DLP$s.
For $A=\emptyset$, these tasks coincide with 
answer set checking which
is known to be $\CONP$-complete (see, for instance, \cite{eite-gott-95}).
Already 
a single element in $A$ yields a mild increase of complexity.

\begin{theorem}\label{thm:re}
Let
$(X,Y)$ be a pair of interpretations, and $P$ a $\DLP$.
Deciding whether $(X,Y)\in\SE^A(P)$ 
$($resp. $(X,Y)\in\UE^A(P))$
is $D^P$-complete.
Hardness 
holds for any fixed $A$ with $\card{A}\geq 1$ 
even
for positive programs.
\end{theorem}

\begin{proof}
We first show $D^P$-membership.
By Definition~\ref{def:relSE}, a pair of interpretations $(X,Y)$
is an $A$-SE-model of $P$ iff (1) $(X,Y)$ is a valid $A$-SE-interpretation;
(2) $Y\models P$;  
(3) for all $Y'\subset Y$ with $(Y'\cap A)=(Y\cap A)$, 
$Y'\not\models P^Y$; and (4)
$X\subset Y$ implies
existence of a $X'\subseteq Y$ with $(X'\cap A)=X$,
such that $X'\models P^Y$ holds.
Obviously, (1) and (2) can be verified in polynomial time.
The complementary problem of (3) can be verified by a guess 
for $Y'$ and a derivability check. As well, (4) can be verified
by a guess for $X'$ and a derivability check. Hence, (3) is in $\CONP$ and
(4) is in $\NP$, which shows $D^P$-membership. 
Similar in the case of $A$-UE-models.
By
Proposition~\ref{prop:rel},
$(X,Y)\in\UE^A(P)$ iff (1) $(X,Y)$ is a valid $A$-SE-interpretation;
(2) $Y\models P$; 
(3) for each $X''\subset Y$ with 
$(X\cap A)\subset (X''\cap A)$ or $X''=(Y\cap A)$,
$X''\not\models P^Y$ holds; and (4) $X\subset Y$ implies that
there exists a $X'\subseteq Y$ with $(X'\cap A)=X$,
such that $X'\models P^Y$.
Similar as before one can verify that the first two conditions are
feasible in polynomial time, whereas checking 
(3) is a $\CONP$-test, and 
checking (4) a $\NP$-test.

For the matching lower bound, we consider the case $\card{A}=1$. 
Therefore, 
$D^P$-hardness of 
both $A$-SE-model checking and $A$-UE-model checking are captured at once.
We 
consider the problem of jointly checking whether
\begin{enumerate}
\item[(a)] a formula $\phi=\bigvee_{i=1}^n D_i$ in DNF is a tautology; and
\item[(b)] a formula $\psi=\bigwedge_{j=1}^m C_j$ in CNF is satisfiable.
\end{enumerate}
This problem is $D^P$-complete, even if both formulas are given over
the same set of atoms 
$V$.
Consider the following positive program 
\begin{eqnarray}
P & = & \{ v \OR \bar{v} \la \mid v\in V \} 
\label{eq:DP1}
\\
& & \{ v \la a_1,D^*_i;\; \bar{v} \la a_1,D^*_i\mid 
v\in V,\;
1\leq i \leq n \}\label{eq:DP2} \\
& & \{ a_1 \la \overline{C}_j \mid 1\leq j \leq m \} \label{eq:DP3}\\
& & \{ a_1 \la  v, \bar{v}  \mid 
v\in V\};\label{eq:DP4} 
\end{eqnarray}
where $a_1$ is a fresh atom.
Let $Y=\{a_1\}\cup V\cup \bar{V}$ and $A=\{a_1\}$. 
We show that $(\emptyset,Y)$ is $A$-SE-model of $P$ 
iff (a) and (b) jointly hold.
Since $P$ is positive, we can argue via classical models (over $Y$).
Rules (\ref{eq:DP1}) have classical models 
$\{ X \mid \sigma(I)\subseteq X \subseteq Y, I\subseteq V\}$.
By (\ref{eq:DP2}) this set splits into 
$S=\{ X \mid \sigma(I)\subseteq X \subseteq (Y\setminus\{a_1\})\}$ and
$T=\{ \sigma(I)\cup\{a_1\} \mid I\notin M_\phi \} \cup \{ Y \}$.
By (\ref{eq:DP4}), $S$ reduces to $\{ \sigma(I) \mid I\subseteq V\}$,
and by (\ref{eq:DP3}) only those elements $\sigma(I)$ survive 
with $I\in M_\psi$.
To summarize, the models of $P$ are given 
by
$$
\{ \sigma(I) \mid I\subseteq V,I\in M_\psi \} \cup \{ \sigma(I)\cup\{a_1\} \mid I\subseteq V,I\notin M_\phi\} \cup \{ Y \}.
$$
From this the $A$-SE-models are easily obtained. 
We want to check whether $(\emptyset,Y)\in\SE^A(P)$.
We have $Y\models P$. 
Further we have that no $Y'\subset Y$ with $a_1\in Y'$ exists such that 
$Y'\models P=P^Y$
iff 
there exists no $I\subseteq V$ 
making $\phi$ false, \iec iff $\phi$ is a tautology.
Finally, to show that $(\emptyset,Y)\in\SE^A(P)$, 
there has to exist an $X\subseteq (V\cup \bar{V})$, such that $X\models P=P^Y$. 
This holds exactly if $\psi$ is satisfiable.
Since $P$ is always polynomial in size of $\phi$ plus $\psi$, 
we derive $D^P$-hardness.

This shows the claim for $\card{A}=1$. 
For $\card{A}>1$, we apply a similar technique as in the proof of
Theorem~\ref{thm:rsehcf}.
However, since we deal here with both $A$-SE-models and $A$-UE-models we
have to be a bit more strict.
Let $k=\card{A}>1$. We add to $P$ the following rules
$\{ a_{i+1}\la a_{i};\; a_{i}\la a_{i+1}\mid 1\leq i < k\}$ and
set $A=\{ a_i \mid 1\leq i \leq k\}$. 
One can show that then, for $Y=A\cup V\cup \bar{V}$,
$(\emptyset,Y)\in\SE^A(P)$ iff $(\emptyset,Y)\in\UE^A(P)$ iff 
(a) and (b) jointly hold.
\end{proof}

With these results for model checking at hand, we obtain numerous 
complexity
results
for deciding relativized equivalence.

\begin{theorem}\label{thm:resmember}
For programs $P$, $Q$, a set of atoms $A$, and $e\in\{s,u\}$, $P\equive^A Q$ is
in $\PiP{2}$. 
\end{theorem}
\begin{proof}
We guess an $A$-SE-interpretation $(X,Y)$. Then, by virtue of 
Theorem~\ref{thm:re}, we can verify that $(X,Y)$ is $A$-SE-model 
(resp. $A$-UE-model) of exactly one of the programs $P$, $Q$ 
in polynomial time with four calls to an $\NP$-oracle 
(since the two model-checking tasks are in $D^P$).
Hence, 
the complementary problem of 
deciding relativized
equivalence is in $\Sigma^P_2$. 
This shows $\PiP{2}$-membership.
\end{proof}

\begin{theorem}
Let $P$, $Q$ be $\DLP$s, $A$ a set of atoms,
and $e\in\{s,u\}$. Then, $P\equive^A Q$ 
is $\PiP{2}$-complete. 
$\PiP{2}$-hardness holds even if $Q$ is Horn.
\end{theorem}

\begin{proof}
Membership is already shown in Theorem~\ref{thm:resmember}.

For the hardness part,
we reduce the $\SigmaP{2}$-complete 
problem of deciding truth of a QBF $\exists X\forall Y  \phi$ with 
$\phi=\bigvee_{i=1}^n D_i$ a DNF to
the complementary problem $P\not\equivs^A Q$.
We define 
\begin{eqnarray*}
P & = & \{ 
x\OR \bar{x}\la ;\; \la x,\bar{x}
\mid x\in X\} \cup\\
& & \{ y\OR\bar{y}\la;\; y\la a;\; \bar{y}\la a;\;
a \la y,\bar{y} 
\mid y\in Y\} \cup\\
& &  \{ a\la D^*_i \mid 1\leq i \leq n \}\cup \\
& &  \{ \la\naf a\};
\end{eqnarray*} 
and take $Q=\{\bot\}$. Note that 
$\{\bot\}$ has no $A$-SE-model, for any $A$.
It thus remains to show that $P$ has an $A$-SE-model iff
the QBF $\exists X\forall Y  \phi$ is true.

$P$ has 
an answer set (\iec an $\emptyset$-SE-model) iff
$\exists X\forall Y  \phi$  is true (see the
$\SigmaP{2}$-hardness proof for 
the program consistency problem
in~\cite{eite-gott-95}). From this we get that ordinary equivalence
is $\PiP{2}$-hard. 
This shows the claim for $\card{A}=0$. 
For $A$ of arbitrary cardinality $k$ it is sufficient to 
add ``dummy'' rules $a_i\la a_i$, for each $1\leq i \leq k$, to 
$P$. These rules do not have any effect on our argumentation. 
Whence, for any fixed $A$,
$\equivs^A$ and $\equivu^A$ are 
$\PiP{2}$-hard as well.
\end{proof}
A slight modification (see Appendix for details) 
of this proof gives us the following result.

\begin{theorem}\label{thm:pospos}
Let $P$ be a positive program, $A$ a set of atoms, and $e\in\{s,u\}$.
Then, deciding whether $P \equive^A Q$ is $\PiP{2}$-complete, 
where $\PiP{2}$-hardness holds even if 
$Q$ is either positive or normal.
\end{theorem}

For head-cycle free programs, RSE and RUE have different
complexities. We first consider RSE. 

\begin{theorem}
\label{theo:hcf-rse}
Let $P$ and $Q$ be head-cycle free programs, 
and $A$ be a set of atoms.
Then, deciding whether $P \equivs^A Q$ is $\PiP{2}$-complete, 
where $\PiP{2}$-hardness holds 
even if $Q$ is normal, and fixed $A$ with $\card{A}\geq 2$.
\end{theorem}

\begin{proof}
As before,
we reduce the problem 
of deciding truth of
a QBF of the form 
$\exists X \forall Y \phi$, with $\phi$ a DNF, 
to the complementary problem of $P\equivs^A Q$ 
using for $P$ a head-cycle free program and for $Q$  a normal program.
We use similar building blocks as in the proofs of the previous 
results, but the argumentation is more complex here. 
We need a 
further new atom $b$, and define
\begin{eqnarray*}
P & = & \{ 
x\OR \bar{x}\la ;\; \la x,\bar{x}
\mid x\in X\} \cup\\
& & \{ y\OR\bar{y}\la;\; y\la a;\; \bar{y}\la a
\mid y\in Y\} \cup\\
& &  \{ b\la D^*_i \mid 1\leq i \leq n \}
\cup \\
& &  \{ b \la y,\bar{y} \mid y\in  Y\}  \cup \{ b\la a \}.
\end{eqnarray*}
Note that $P$ is head-cycle free.
For the matter of presentation, suppose first $X=\emptyset$. 
We show that $\phi$ is valid 
iff $P\not\equivs^A P^\ra$ holds, for $A=\{a,b\}$. 
Afterwards, we 
generalize the claim to arbitrary $X$ and 
show that $P\not\equivs^A P^\ra$ iff $\exists X\forall Y\phi$ is true holds,
for any 
$A$ of the form $\{a,b\}\subseteq A\subseteq (X\cup\bar{X}\cup \{a,b\})$.

Let us first compute the $A$-SE-models of $P$
under the assumption that $X=\emptyset$.
Since $P$ is positive,
this is best accomplished by first considering the 
classical models of $P$.
These are given as follows:
\begin{itemize}
\item[(a)] $\sigma(I)$ for each $I\subseteq Y$ making $\phi$ false;
\item[(b)] $\sigma(I)\cup \{b\}$ for each $I\subseteq Y$;
\item[(c)] all $M$ satisfying $(\sigma(I)\cup \{b\})\subset M \subseteq (Y\cup\bar{Y}\cup\{b\})$ for some $I\subseteq Y$; and
\item[(d)] $\at = Y\cup \bar{Y}\cup \{a,b\}$.
\end{itemize}
Note that 
(a), (b), and (d) become total $A$-SE-models of $P$; 
while the elements in (c) do not.
In fact, for each element $M$ in (c) there exists 
a corresponding element $M'$ from (b), such 
that $M'\subset M$ and $(M'\cap A)=(M\cap A)=\{b\}$. 
It remains to consider non-total $A$-SE-models of $P$, 
by combining the elements from (a), (c), (d).
If there exists an element in (a) (\iec $\phi$ is not valid), 
then we get 
$(\emptyset,\sigma(I)\cup \{b\})\in\SE^A(P)$, for each $I\subseteq V$; 
as well we then have also $(\emptyset,\at)\in\SE^A(P)$. 
Combining (b) and (c), 
yields $(\{b\},\at)\in\SE^A(P)$. 
Hence, 
\begin{eqnarray*}
\SE^A(P) & = & \{ (\sigma(I),\sigma(I)) \mid I\subseteq V: \phi\mbox{\ is false under\ }I \}  \cup \\ 
& & \{ (\sigma(I)\cup \{b\}, \sigma(I)\cup\{b\}) \mid I\subseteq V \} \cup \\
& & \{ (\emptyset, \sigma(I)\cup \{b\}) \mid I\subseteq V\mbox{, if $\phi$ is not valid} \} \cup  \\
& & \{ (\emptyset, \at) \mid \mbox{if $\phi$ is not valid} \} \cup  \\
& & \{ ( \{b\},\at), (\at,\at) \}.
\end{eqnarray*}
For $P^\ra$ we get a (possibly) additional $A$-SE-model, viz.
$(\emptyset, \at)$, since $(\emptyset,\at)\in\SE(P^\ra)$ holds in any case, 
also if $\phi$ is valid.
Hence, the $A$-SE-models of $P$ and $P^\ra$ coincide iff $\phi$ is not valid.

The extension to $X\neq\emptyset$ and deciding truth of QBF 
$\exists X\forall Y\phi$ via the complementary problem $\equivs^A$ 
is similar to the argumentation in the proof of Theorem~\ref{theo-complex-2}.
In particular, we then can use any $A$ 
with $\{a,b\}\subseteq A\subseteq (X\cup\bar{X}\cup\{a,b\})$.
Recall that deciding $\exists X\forall Y\phi$ is $\SigmaP{2}$-complete, and 
thus we get that $P\equivs^A Q$ is $\PiP{2}$-hard for $P$ a HCF program, $Q$ normal. 
\end{proof}

This concludes the collection of problems which are located at the 
second level of the polynomial hierarchy.
Note that in the hardness part of the proof of Theorem~\ref{theo:hcf-rse}, we used at least two elements
in $A$. In fact, for HCF programs 
and $\card{A}\leq 1$ the complexity is different. 
Since for $\card{A}\leq 1$, $\equivs^A$ and $\equivu^A$ are the same
concepts, this special case is implicitly considered in the next
theorem.  Another issue is to decide $P\equivs^A Q$ if both $P$ and
$Q$ are $A$-HCF as introduced in Definition~\ref{def:ahcf}.  In this
case, we can employ $P^\ra \equivs^A Q^\ra$, and thus the complexity
coincides with the complexity of $\equivs^A$ for normal
programs. This is also part of the next theorem.

\begin{theorem}\label{thm:conpall}
Deciding $P\equive^A Q$ is $\CONP$-complete in the following 
settings:
\begin{itemize}
\item[(i)] $e\in\{s,u\}$, $P$ positive, $Q$ Horn;
\item[(ii)] $e=s$, $P$ head-cycle free and $Q$ Horn;
\item[(iii)] $e\in\{s,u\}$, $P$ and $Q$ normal;
\item[(iv)] $e=u$, $P$ and $Q$ head-cycle free.
\end{itemize}
$\CONP$-hardness of $P\equive^A Q$ $(e\in\{s,u\})$ holds 
even if $P$ is normal or positive and $Q$ is Horn.
\end{theorem}
\begin{proof}
We start with the $\CONP$-membership results.  The cases (iii) and
(iv) follow immediately from Theorem~\ref{thm:reltr}, since 
$A$-SE/UE-model checking for the programs involved is feasible in
polynomial time.  The more complicated cases ($i$) and ($ii$) 
are addressed in the Appendix. 

It remains to show the $\CONP$-hardness part of the theorem.
We use UNSAT of a formula $F=\bigwedge_{i=1}^n C_i$ in CNF over atoms $X$. 
Take 
$$
P=\{ x\vee \bar{x}\la;\; \la x,\bar{x} \mid x\in X\} \cup \{ \la \overline{C}_i \mid 1\leq i \leq n\}
$$
Note that this program is positive and HCF. 
The program has a classical model 
iff $F$ is satisfiable, \iec iff
it is not equivalent to the Horn program $Q=\{\bot\}$. 
In other words, $\SE^A(P)\neq\emptyset$ (or, resp.\ $\UE^A(P)\neq\emptyset$)
iff $\phi$ is satisfiable.
Note that $A$ can thus be of any form.
Since the rules $\la x,\bar{x}$ are present in $P$,
we have $P\equivs^A P^\ra$. This proves $\CONP$-hardness also for the 
case where one program is normal and the other is Horn.
\end{proof}

A final case remains open, namely 
checking relativized equivalence of Horn programs.
Unfortunately, this task is $\CONP$-complete. 
However, whenever the cardinality of $A$ is fixed by a constant the problem 
gets tractable. This is in contrast to the hardness results proved so far, 
which even hold in the case where $\card{A}$ is fixed. The proof of the theorem
is given in the Appendix.

\begin{theorem}\label{thm:horn}
Deciding $P\equive^A Q$, for $e\in\{s,u\}$, 
is $\CONP$-complete for Horn programs $P$, $Q$.
Hardness holds whenever $\card{A}$ is not  fixed  by a constant, 
and even for definite Horn programs.
\end{theorem}

Whenever the cardinality of $A$ is bounded, we can 
decide this problem in polynomial time.

\begin{theorem}\label{thm:hornfixed:1}
Let $P$, $Q$ be Horn programs and $A$ be a set of atoms such that 
$\card{A}\leq k$
with a fixed constant $k$. 
Then, deciding $P\equive^A Q$ is feasible in polynomial time with respect
to $\|P\| + \|Q\| + k$.
\end{theorem}
\begin{proof}
It is sufficient to show the claim for $e=u$. 
By explicitly checking
whether $(P\cup S) \equiv (Q\cup S)$ holds for any $S\subseteq A$.
we obtain a polynomial-time algorithm, 
since checking ordinary equivalence of Horn programs is 
polynomial and we 
need at most  
$2^k$ such checks.
\end{proof}

\subsection{Complexity of Bounded Relativization}
In this section,
we pay attention to the special case of
tests $\equivs^A$ and $\equivu^A$
where the number
of atoms from 
the considered programs 
{\em missing} in $A$, is bounded by some constant $k$ (in symbols
$P\,{^k}\!\!\equivs^A Q$, and resp., $P\,{^k}\!\!\equivu^A Q$). 
Hence,
the respective problem classes apply to programs $P$, $Q$, 
only if $\card{\var{P\cup Q}\setminus A} \leq k$.
Apparently,
this class of problems contains strong and uniform equivalence
in its unrelativized versions ($k=0$).
The complexity results are summarized in Table~\ref{tab:resbound}.
In particular, we 
get that in the case of RSE
all entries (except Horn/Horn) reduce to $\CONP$-completeness. 
This generalizes results on strong equivalence. 
Previous work 
reported some of these results but not in form of this exhaustive list.

\begin{table}[t!]
\begin{center}
\begin{tabular}{|l|lllll|}
\hline
$P\,{^k}\!\!\equivs^A Q$ / $P\,{^k}\!\!\equivu^A Q$
& $\DLP$ & positive & HCF & normal & Horn \\
\hline
Horn & $\CONP$ & $\CONP$ & $\CONP$ & $\CONP$ & $\Pol$ \\
normal & $\CONP$ & $\CONP$ & $\CONP$ & $\CONP$ & \\
HCF & $\CONP$  & $\CONP$ & $\CONP$ & & \\
positive & $\CONP$/$\PiP{2}$ & $\CONP$ & & & \\
\DLP & $\CONP$/$\PiP{2}$ & & & &
\\ \hline
\end{tabular}
\end{center}
\caption{Complexity of Equivalences with Bounded Relativization in Terms of Completeness Results.}\label{tab:resbound}
\end{table}

In what follows, we first give the respective results for
model checking, and then we prove the entries in Table~\ref{tab:resbound}.

\begin{lemma}\label{lemma:sebounded}
For a program $P$, and a set of atoms $A$, such 
that 
$\card{\var{P}\setminus A}\leq k$, with 
$k$ a fixed constant,
$A$-SE-model checking is feasible in polynomial time
with respect to to $\|P\| + k$.
\end{lemma}

\begin{proof} 
By the conditions in 
Definition~\ref{def:relSE}, 
deciding $(X,Y)\in\SE^A(P)$ can be done as follows:
(i) checking $Y\models P$; 
(ii) checking whether for all $Y'\subset Y$ with $(Y'\cap A)=(Y\cap A)$,
$Y\not\models P^Y$ holds; and
(iii) if $X\subset Y$, checking existence of a $X'\subseteq Y$ with
$(X'\cap A)=X$, such that $X'\models P^Y$ holds.
Test~(i) can be done in polynomial time; 
test~(ii) is a conjunction of at most $2^k-1$ independent polynomial tests (for 
each such $Y'$), while (iii) is a disjunction of at most $2^k$ polynomial tests (for each $X'$). Since we have fixed $k$ the entire test is feasible 
in polynomial time.
\end{proof}

Compared to the model checking problems discussed so far, 
the polynomial-time decidable problems of $A$-SE-model checking in the bounded
case do not belong to the class of $\Pol$-complete problems, but
are easier. This is best illustrated by SE-model checking, which 
obviously reduces to two (ordinary) independent 
model checking tests; which in turn
are in ALOGTIME~\cite{buss} (see also~\cite{barr-etal-90,imme-99}).
For bounded $A$-SE-model checking the situation is basically the same, 
since it is sufficient to employ a fixed number of independent 
model checking tests.

Concerning UE-model checking we already established some 
$\Pol$-hardness results in Theorem~\ref{thm:phard} which generalize to 
the relativized case for
arbitrary bound $A$.
In general, for
$A$-UE-model checking the decrease of 
complexity is in certain cases only moderate compared to 
the corresponding decrease in the case of
$A$-SE-model checking.

\begin{lemma}\label{lemma:checkconp}
For a program $P$ and a set of atoms $A$, such that 
$\card{\var{P}\setminus A}\leq k$, with 
$k$ a fixed constant,
$A$-UE-model checking is $\CONP$-complete. 
Hardness holds even for positive programs. 
\end{lemma}
\begin{proof}
We show $\NP$-membership for the complementary problem, \iec
checking whether a given pair $(X,Y)$ is not in $\UE^A(P)$.
We first check whether $(X,Y)$ is $A$-SE-model of $P$. This can 
be done in polynomial time, by Lemma~\ref{lemma:sebounded}.
If this is not the case we are done; otherwise,
we
guess an $X'$ with 
$X\subset X'\subset (Y\cap A)$ 
and check whether
$(X',Y)$ is $A$-SE-model of $P$. 
This guess for $(X',Y)$ can be verified in polynomial time using an 
$\NP$ oracle.
Therefore, the entire problem is in $\NP$.
The correctness of the procedure is 
given by its direct reflection of Definition~\ref{def:relUE}.
This yields $\CONP$-membership for bounded $A$-UE-model checking.

Hardness is obtained via 
the case $\card{\var{P}\setminus A}=0$, \iec
ordinary UE-model checking and the respective result 
in Theorem~\ref{theo-complex-3}.
\end{proof}

\begin{theorem}\label{thm:boundse}
For programs $P$, $Q$ 
and a set of atoms $A$,
such that 
$\card{\var{P\cup Q}\setminus A} \leq k$ with $k$ a fixed constant,
$P\equivs^A Q$ is $\CONP$-complete. 
Hardness holds provided $P$ and $Q$ are not Horn.
\end{theorem}
\begin{proof}
By Lemma~\ref{lemma:sebounded}, 
$A$-SE-model checking is feasible in polynomial time in the bounded case.
Hence, $\CONP$-membership for $P\equivs^A Q$ is an immediate consequence.
The hardness result is easily obtained by the hardness 
part from Theorem~\ref{thm:conpall}.
\end{proof}

For RUE some cases remain on the second level, however. 
This
is not a surprise, since as we have seen in 
Theorem~\ref{theo-complex-2}, 
(unrelativized)
uniform equivalence is $\PiP{2}$-complete in general.

\begin{theorem}
For programs $P$, $Q$ 
and a set of atoms $A$,
such that 
$\card{\var{P\cup Q}\setminus A} \leq k$ with $k$ a fixed constant,
$P\equivu^A Q$ is $\PiP{2}$-complete. $\PiP{2}$-hardness holds even if 
one of the programs is positive.
\end{theorem}
\begin{proof}
Membership is obtained 
by the fact that
$A$-UE-model checking with $A$ bounded is $\CONP$-complete 
(see Lemma~\ref{lemma:checkconp}). Hardness comes from 
the $\PiP{2}$-hardness of uniform equivalence.
\end{proof}

For all other cases, RUE for bounded $A$ is in $\CONP$.

\begin{theorem}
For programs $P$, $Q$ and a set of atoms $A$,
such that 
$\card{\var{P\cup Q}\setminus A} \leq k$ with $k$ a fixed constant,
$P\equivu^A Q$ is $\CONP$-complete, if 
either (i) 
both programs are positive; or (ii)
at least one program is head-cycle free.
Hardness holds, 
even if $P$ is normal or positive and $Q$ is Horn.
\end{theorem}

\begin{proof}
We start with $\CONP$-membership. For (i) this is an immediate consequence
of the fact that 
for positive programs, RSE and RUE are the same concepts and  since
RSE is $\CONP$-complete as shown in Theorem~\ref{thm:boundse}.

For~(ii) we argue as follows.
Consider $P$ is HCF.
By Theorem~\ref{thm:subsetUE} it is sufficient to 
check (a) $\UE^A(P)\subseteq \SE^A(Q)$ and
(b) $\UE^A(Q)\subseteq \SE^A(P)$.
We show that both
tasks are in $\CONP$.
ad (a): For the complementary problem we guess a 
pair $(X,Y)$ and check whether $(X,Y)\in\UE^A(P)$ and
$(X,Y)\notin\SE^A(Q)$. Both checks are already shown to be 
feasible in polynomial time. 
ad (b): 
We consider the complementary problem and show that this reduces
to the disjunction of two 
$\NP$ problems.
First, we consider total $A$-SE-interpretations.
By guessing $Y$ and check whether $(Y,Y)\in\SE^A(Q)$ and 
$(Y,Y)\notin\SE^A(P)$, we get obtain $\NP$-membership.
If this holds, we 
secondly we consider non-total $A$-SE-interpretations.
We claim that existence of a $(X,Y)\in\SE^A(Q)$, such that, 
for each $X\subseteq X'\subset (Y\cap A)$, $(X',Y)\notin\UE^A(P^\ra)$,
implies $\UEA(Q)\not\subseteq \SEA(P)$.
This can be seen as follows. 
Given $X$, $Y$, 
suppose no $X\subseteq X'\subset (Y\cap A)$ satisfies
$(X',Y)\in\UE^A(P^\ra)$. Then, no such $(X',Y)$ is $A$-UE-model
of the original $P$ (by Theorem~\ref{theo-shift-equivu-1}).
By definition,
no such $(X',Y)$ is $A$-SE-model of $P$.
On the other hand, either $(X,Y)\in\UE^A(Q)$ or for some such 
$X'$, $(X',Y)\in\UE^A(Q)$. Hence, 
$\UEA(Q)\not\subseteq \SEA(P)$.
Therefore, we guess a pair $(X,Y)$
and check $(X,Y)\in\SE^A(Q)$ 
and whether $T=(P^\ra)^Y\cup X \cup Y_\subset$
is unsatisfiable. Both can be done in polynomial time.
It remains to show that $T$ 
is unsatisfiable iff, 
for each $X\subseteq X'\subset (Y\cap A)$, $(X',Y)\notin\UE^A(P^\ra)$.
Suppose $T$ is satisfiable and let  $X'$ be a maximal interpretation
making $T$ true.
Then $(X'\cap A)\subset (Y\cap A)$ holds, 
since $(Y,Y)\in\UE^A(P)$ 
(and thus $(Y,Y)\in\SE^A(P^\ra)$)
by assumption that 
the total $A$-SE-models of $P$ and $Q$ coincide.  
But then $((X'\cap A),Y)\in\UE^A(P^\ra)$, since $X'$ is a maximal model of $T$.
On the other hand, if $T$ is unsatisfiable,
no $(X',Y)$ with $X\subseteq X'\subset (Y\cap A)$ can be $A$-SE-model of $P$,
and thus no such $(X',Y)$ is $A$-UE-model of $P$ and thus of $P^\ra$.
This gives membership for $\NP$.
Since $\NP$ is closed under disjunction, the entire complementary problem
is shown to be in $\NP$.

The matching lower bound is obtained from 
the hardness result in Theorem~\ref{thm:conpall}.
\end{proof}

One final case remains to be considered.

\begin{theorem}\label{thm:hornfixed:2}
Let $P$, $Q$ be Horn programs and let $A$ be a set of atoms such that 
$\card{\var{P\cup Q}\setminus A} \leq k$ with
a fixed constant $k$. 
Then, deciding $P\equive^A Q$ is 
feasible in polynomial time with respect
to $\|P\| + \|Q\| + k$.
\end{theorem}

\begin{proof}
We use the following characterization 
which can be derived from Theorem~\ref{thm:amin}:
For positive programs $P$, $Q$,  $P\equive^A Q$ holds, iff, 
for each model $Y$ of $P$, there exists a $X\subseteq Y$ with $(X\cap A)=(Y\cap A)$ being model of $Q$; 
and vice versa.
This can be done as follows.
We show one direction, \iec whether, for each interpretation $Y$, $Y\models P$ 
implies $X\models Q$ for some $X\subseteq Y$, such that $(X\cap A)=(Y\cap A)$.
Let $V=(\var{P\cup Q}\setminus A)$.
We test, 
for every 
$U\subseteq V$ 
and each 
$W\subseteq U$, 
whether
\begin{equation}\label{eq:horn}
P'_V \cup (U^V_=)' \cup (W^V_=)
\models Q;
\end{equation}
where $P'$ results from $P$ by replacing each $v\in V$ occurring in $P$
by $v'$ and $(U^V_=)'$ is the set $\{ v' \mid v\in U^V_=\}$ 
with $U^V_=$ as defined in the beginning of the section.
Observe that both sides in the derivability test (\ref{eq:horn}) are 
Horn programs.

$P'_V \cup (U^V_=)'$ has a model $R \cup S'$
iff there exists a $R\subseteq A$ and a $S'\subseteq V'$ 
such that $R\cup S'$ is a model of $P'_V$, \iec iff
$R\cup S$ is a model of $P$. Then, we check whether 
for one $W\subseteq U$, $R\cup W$ is model of $Q$. This 
matches the test whether for each model $R\cup S$ of $P$, there exists a $R\cup W$ with 
$((R\cup W)\cap A)=((R\cup S)\cap A)=R$, 
such that $R\cup W$ models $Q$, \iec the property to be tested.
This 
yields $O(2^k \times 2^k)=O(2^{k+1})$ Horn-derivability tests. 
The same procedure is done the other direction, \iec exchanging $P$ and $Q$.
Whenever $k$ is fixed, this gives us a polynomial time algorithm. 
(More efficient algorithms may be given, but we do not focus on this here.)
\end{proof}

\section{Language Variations}
\label{sec:extensions}

In this section, we briefly address how our results apply to 
variations of the language of logic programs. 
First, we consider modifications within the case of propositional programs, 
and then discuss the general DATALOG case.

\subsection{Extensions in the Propositional Case}

\paragraph{Adding Classical Negation.}

Our results easily carry over to extended logic programs, \iec programs where 
classical (also called strong) negation is allowed as well. If the 
inconsistent answer set is disregarded, \iec an inconsistent program has no 
models, then, as usual, the extension can be semantically captured by 
representing strongly negated atoms $\neg A$ by a positive atom $A'$ and 
adding constraints $\la A,A'$, for every atom $A$, to any program.

However, if in the extended setting the inconsistent answer set is
taken into account, then the given definitions have to be slightly
modified such that the characterizations of uniform equivalence
capture the extended case properly. The same holds true for the
characterization of strong equivalence by SE-models as illustrated by
the following example. Note that the redefinition of $\equivu$ and
$\equivs$ is straightforward.

Let $\extat = \{ A, \neg A \mid A \in \at\}$ denote the (inconsistent)
set of all literals using strong negation over $\at$.  Note that an
extended $\DLP$ $P$ has an inconsistent answer set iff $\extat$ is an
answer set of it; moreover, it is in the latter case the only answer
set of $P$. Call any $\DLP$ $P$ {\em contradiction-free}, if $\extat$ is
not an answer set of it, and {\em contradictory} otherwise.

\begin{example}\label{ex-ext}
Consider the extended logic programs 
$P=\{a\vee b\la\; ;\ \neg a \la a ;\  \neg b\la b\}$ and 
$Q=\{a\la \naf b ;\ b\la \naf a ;\ \neg a \la a ;\  \neg b\la b\}$.
They both have no SE-model; hence, by the criterion of Prop.~\ref{prop:turner},
$P\equivs Q$ would hold, which implies  $P\equivu Q$ 
and  $P\equiv Q$.
However, $P$ has the inconsistent answer set $\extat$, while $Q$ has no 
answer set. 
Thus formally, $P$ and $Q$ are not even 
equivalent if $\extat$ is admitted as answer set. 
\end{example}

Since~\cite{Turner01,Lifschitz01,Turner03} made no distinction between no 
answer set and inconsistent answer set, 
in \cite{eite-fink-03a} we adapted the definition of SE-models accordingly  
and got more general characterizations in terms of so-called SEE-models 
for extended programs. Many results easily carry over to the extended case: 
E.g., for positive programs, uniform and strong equivalence coincide also in 
this case and, as a consequence of previous complexity results, checking 
$P\equivu Q$ (resp.~$P\equivs Q$) for extended logic programs, $P$ and $Q$, 
is $\PiP{2}$-hard (resp.~$\coNP$-hard).

However, not all properties do carry over. As Example~\ref{ex-ext}
reveals, in general a head-cycle free extended $\DLP$ $P$ is no longer
equivalent, and hence not uniformly equivalent, to its shift variant
$P^\la$ (see \cite{eite-fink-03a} for a characterization of head-cycle 
and contradiction free programs for which this equivalence holds).

We expect a similar picture for relativized equivalences of extended 
logic programs but adapting corresponding proofs is still subject of 
future work.

\paragraph{Disallowing Constraints.}
Sometimes, it is desirable to consider constraints just as
abbreviations, in order to have core programs which are definite, \iec
without constraints.
The most direct approach is to replace each constraint 
$\la B$ by $w\la B, \naf w$; where $w$ is a designated atom not occurring
in the original program. 
Obviously, this does not influence ordinary equivalence tests, but
for notions as uniform and strong equivalence some more care is required.
Take the strongly equivalent programs $P=\{a\la\naf a\}$ 
and $Q=\{\la\naf a\}$. By above rewriting $Q$ 
becomes $Q'=\{w\la\naf a,\naf w\}$. Then, $(\cdot,w)\notin\SE(P)$ but
$(\cdot,w)\in\SE(Q')$. Hence, this rewriting is not sensitive 
under strong equivalence. However, if we disallow $w$ to appear in 
possible extensions, \iec employing $\equivs^A$ instead of $\equivs$ 
we can circumvent this problem. Simply take $A=U\setminus \{w\}$
where $U$ is the universe of atoms. Observe that this employs
bounded relativization, and in the light of Theorem~\ref{thm:boundse} 
this workaround does not result in a more complex problem.
For uniform equivalence the methodology can be applied in the same manner.

However, this approach requires (unstratified) negation. 
If we want to get rid off constraints for
comparing positive programs, an alternative method is 
to use a designated (spoiled) answer set to indicate that the original
program had
no answer set. 
The idea is to replace each constraint $\la B$ by $w\la B$, 
where $w$ is a designated atom as above; additionally we add
the collection of rules $v\la w$ for each atom $v$ of the universe 
to both programs (even if no constraint is present). 
This rewriting retains any equivalence notion, even
if $w$ is allowed to occur in the extensions. 

The problem of comparing, say, a positive program $P$ (with constraints)
and a normal program $Q$ is more subtle, if we require to replace the
constraints in $P$ by positive rules themselves. 
We leave this for further study, 
but refer to some results in~\cite{Eiter03a}, which suggest that
these settings may not be solved in an easy manner. To wit, 
\cite{Eiter03a} reports that the complexity for some problems of the form 
``Given a program $P$ from class $C$; does there exist a program 
$Q$ from class $C'$, such that $P\equive Q$?'' differs
with respect to allowing constraints.

\paragraph{Using Nested Expressions.}
Programs with nested expressions~\cite{Lifschitz99} (also called
nested logic programs) extend $\DLP$s in such a way that
arbitrarily nested
formulas, formed from literals using negation as failure, conjunction, and
disjunction, constitute the heads and bodies of rules.
Our characterizations for uniform equivalence are well suited for this
class as was shown in~\cite{Pearce04}. 
Since the proofs of our main results are generic in the
use of reducts, we expect that all results (including relativized notions 
of equivalence) can be carried over to nested logic programs without any 
problems. Note however, that the concrete definitions for 
subclasses (positive, normal, etc.) 
have to be extended
in the context  of  nested logic programs (see~\cite{Linke04a} for such
an extension of head-cycle free programs).
It remains for further work to 
apply our results to such classes.

\subsection{DATALOG programs}

The results in the previous sections on propositional logic programs 
provide an extensive basis for studying equivalences of DATALOG 
programs if, as usual, their semantics is given in terms of 
propositional programs. Basic notions and concepts for strong and uniform 
equivalence such as SE-models, UE-models, and the respective notions of 
consequence generalize naturally to this setting, using Herbrand interpretations 
over a relational alphabet and a set of constants in the usual way 
(see \cite{imix-d5.3-04}).
Furthermore, fundamental results can be lifted to DATALOG programs by reduction 
to the propositional case. In particular, the elementary characterizations 
$P\equive Q$ iff $M_e(P)=M_e(Q)$ iff $P\modelse Q$ and $Q\modelse P$ 
carry over to the DATALOG setting for $e\in\{s,u\}$ and $M_s(\cdot)=\SE(\cdot)$, 
respectively $M_u(\cdot)=\UE(\cdot)$ (see also \cite{imix-d5.3-04}). However, 
a detailed analysis of the DATALOG case including relativized notions of 
equivalence is subject of ongoing work.

Nevertheless, let us conclude this section with some remarks on the
complexity of programs with variables. For such programs, in case of a
given {\em finite} Herbrand universe the complexity of equivalence
checking, resp.~model checking, increases by an
exponential. Intuitively, this is explained by the exponential size of
a Herbrand interpretation, \iec the ground instance of a program over
the universe.  Note that \cite{Lin02} reported (without proof) that
checking strong equivalence for programs in this setting is in \coNP,
and thus would have the same complexity as in the propositional case;
however, for arbitrary programs, this is not correct.
Unsurprisingly, over {\em infinite} domains, in the light of the
results in \cite{Shmueli93,hale-etal-01}, decidability of equivalence
and inference problems for DATALOG programs is no longer
guaranteed. While strong equivalence and SE-inference remain decidable
(more precisely complete for co-NEXPTIME), this is not the case for
uniform equivalence (respectively inference) in general.  For positive
programs, however, the two notions coincide and are decidable (more
precisely complete for co-NEXPTIME); see \cite{imix-d5.3-04} for
details. It remains as an issue for future work to explore the
decidability versus undecidability frontier for classes of DATALOG
programs, possibly under restrictions as in \cite{hale-etal-01,chau-vard-92}.

\section{Conclusion and Further Work}
\label{sec:conclusion}

In this paper, we have extended the research about equivalence of
nonmonotonic logic programs under answer set semantics, 
in order 
to simplify parts (or modules) of a program, 
without analyzing the entire program.
Such local simplifications call for alternative notions of equivalence,
since a simple comparison of the answer sets 
does not provide information whether a program part can be replaced
by its simplification.
To wit, by the non-monotonicity of the answer set 
semantics, 
two (ordinary) equivalent (parts of) programs may lead to different 
answer sets if they are used in the same global program $R$.
Alternative notions of equivalence thus
require that the answer sets of the two programs coincide under
different 
$R$:
strong equivalence~\cite{Lifschitz01}, for instance, 
requires that the compared programs are equivalent under any 
extension $R$.

In this paper, we have considered further notions of 
equivalence, in which the actual form of 
$R$ is syntactically constrained: 

\begin{itemize}
\item
Uniform equivalence of logic programs, which has been considered
earlier for DATALOG and general Horn logic programs~\cite{sagi-88,mahe-88}.
Under 
answer set semantics
uniform equivalence can be exploited for optimization
of components in a logic program which is modularly structured. 
\item 
Relativized notions of both uniform and strong equivalence restrict
the alphabet of the extensions. This allows to specify which 
atoms may occur in the extensions, and which do not. This 
notion
of equivalence for answer set semantics
was originally suggested by Lin in~\cite{Lin02} but not further investigated. 
In practice,
relativization is a 
natural concept,
since it allows to specify internal atoms, which only
occur in the compared program parts, but it is guaranteed that they do 
not occur anywhere else. 
\end{itemize}

We have provided semantical characterizations of all
these notions of equivalence 
by adopting the concept of SE-models~\cite{Turner01}
(equivalently, HT-models \cite{Lifschitz01}), which capture the
essence of a program with respect to strong equivalence. Furthermore,
we have thoroughly analyzed the complexity of equivalence checking and
related problems  for the general case and several important fragments.
This collection of results gives 
a valuable 
theoretical underpinning for
advanced methods of program optimization and for enhanced ASP
application development, as well as a potential basis for the
development of ASP debugging tools.

Several issues remain for further work. One issue is a
characterization of uniform equivalence in terms of ``models'' for
arbitrary programs in the infinite case; as we have shown, no subset
of SE-models serves this purpose. In particular, a notion of models
which correspond to the UE-models in the case where the latter capture
uniform equivalence would be interesting.

We focused here on the propositional case, to which general programs
with variables reduce, and we just briefly mentioned a possible
extension to a DATALOG setting \cite{imix-d5.3-04}. Here,
undecidability of uniform equivalence arises if negation may be present 
in programs. A thorough study of cases under which uniform
equivalence and the other notions of equivalence are decidable is
needed, along with complexity characterizations. Given that in
addition to the syntactic conditions on propositional programs
considered here, further ones involving predicates might be taken into
account (cf.\ \cite{chau-vard-92,hale-etal-01}), quite a number of different
cases remains to be analyzed. 

Finally, an important issue is to explore the usage of uniform
equivalence and relativized equivalence in program replacement and
rewriting, and to develop optimization methods and tools for Answer
Set Programming; a first step in this direction, picking 
up some of the results of this paper, has been made in
\cite{Eiter03a}. However, much more remains to be done.


\subsubsection*{Acknowledgments.} 

The authors would like to thank David Pearce for interesting discussions and comments about this work and pointers to related
literature, as well as Katsumi Inoue and Chiaki Sakama for their 
valuable comments 
on relativizing equivalence. 
We are also grateful to the anonymous reviewers of ICLP
2003 and JELIA 2004. Their comments on submissions preliminary to this 
article helped to improve this work, as well.


{\small

\makebblfalse
\ifmakebbl

\bibliographystyle{abbrv}
\bibliography{rel}

\else

\fi 

}

\appendix
\section{Proofs}
\subsection{Proof of Lemma~\ref{theo-equivru-1}}
For the only-if direction, suppose $P\equivu^A Q$. If $(Y,Y)$ is neither 
$A$-SE-model of $P$, nor of $Q$, then 
$(X,Y)$ is not an $A$-SE-model of any of the programs
$P$ and $Q$.  
Without loss of generality, assume $(Y,Y)\in\SE^A(P)$ and
$(Y,Y)\notin\SE^A(Q)$. 
Let $F=(Y\cap A)$. 
We have the following situation by definition of $A$-SE-models. 
First, from $(Y,Y)\in\SE^A(P)$, we get $Y\models P$. Hence, 
$Y\models P\cup F$. 
Second, $(Y,Y)\in\SE^A(P)$ implies that for each $Y'\subset Y$ 
with $(Y'\cap A)=(Y\cap A)$, $Y'\not\models P^Y$. Hence, 
for each such $Y'$, $Y'\not\models (P\cup F)^Y$.
Finally, for each $X\subset Y$ with $(X\cap A)\subset (Y\cap A)$, 
$X\not\models F$ and thus $X\not\models P^Y\cup F$. 
To summarize, we arrive at $Y\in\SM(P\cup F)$. 
On the other hand, $Y\not\in\SM(Q\cup F)$. 
This can be seen as follows. By $(Y,Y)\not\in\SE^A(Q)$, either $Y\not\models Q$ or there exists an $Y'\subset Y$ with $(Y'\cap A)=(Y\cap A)$, such that
$Y'\models Q^Y$. But then, $Y'\models (Q\cup F)^Y$.
Hence, this contradicts our assumption $P\equivu^A Q$, since $F$ is a set of facts over $A$.
Item ($i$) must hold.

To show ($ii$), assume first that $(X,Y)$ is an $A$-SE-model of $P$ but
not of $Q$.  In view of ($i$), it is clear that $X\subset Y$ must
hold. Moreover, $X\subseteq A$.
Suppose now that for every set $X'$, $X\subset X'\subset Y$, it
holds that $(X',Y)$ is not an $A$-SE-model of $Q$. Then, since no subset of
$X$ models $Q^Y\cup X$, 
$(Y,Y)$ is the only $A$-SE-model of $Q\cup X$ of
form $(\cdot,Y)$. 
Thus, $Y\in\SM(Q\cup X)$ in this
case, while $Y\not\in\SM(P\cup X)$. 
This is seen as follows: Since $(X,Y)\in\SE^A(P)$, there exists an 
$X'\subseteq Y$ with $(X'\cap A)=(X\cap A)$, such that $X'\models P^Y$.
Moreover, $X'\models (P\cup X)^Y$. Thus, $Y\notin\SM(P\cup X)$.
This contradicts $P\equivu^A Q$, since $X\subseteq A$. 
Thus, it follows that for some $M$ such that 
$X \subset M \subset Y$, $(M,Y)$ is an $A$-SE-model of $Q$.
The argument in the case where $(X,Y)$ is an SE-model of $Q$ but
not of $P$ is analogous. This proves item ($ii$).

For the if direction, assume that ($i$) and ($ii$) hold for every
$A$-SE-interpretation $(X,Y)$ which is an 
$A$-SE-model of exactly one of $P$ and $Q$.
Suppose that there exist sets of atoms $F\subseteq A$ 
and $Z$, such that w.l.o.g., 
$Z\in\SM(P\cup F)$, but $Z\notin\SM(Q\cup F)$.
Since $Z\in\SM(P\cup F)$, we have 
that $F\subseteq Z$, $Z\models P$, 
and, for each $Z'\subset Z$ with 
$(Z'\cap A) = (Z \cap A)$, $Z'\not\models P^{Z}$.
Consequently, $(Z,Z)$ is an $A$-SE-model of $P$.  
Since $Z\not\in\SM(Q\cup F)$, either $Z\not\models (Q\cup F)$,
or there exists  a $Z'\subset Z$ such that $Z'\models (Q\cup F)^{Z}$.

Let us first assume $Z\not\models (Q\cup F)$. 
However, since $F\subseteq Z$, we get $Z\not\models Q$. We immediately 
get $(Z,Z)\notin \SE^A(Q)$, \iec $(Z,Z)$ violates ($i$).
It follows that $Z\models (Q\cup F)$ must hold, and that there must exist 
a $Z'\subset Z$ such that $Z'\models (Q\cup F)^{Z} = Q^{Z}\cup F$. 
We have two cases: If $(Z'\cap A)=(Z\cap A)$, then, by definition of
$A$-SE-models, 
$(Z,Z)\notin\SE^A(Q)$, as well. 
Hence, the following relations hold
$Z\models Q$; for each $Z'$ with 
$(Z'\cap A)=(Z\cap A)$, $Z'\not\models Q^{Z}$, and there exists an $Z''$ 
with $(Z''\cap A)\subset (Z\cap A)$, such that $Z''\models Q^{Z}$.
We immediately get that $((Z''\cap A),Z)\in\SE^A(Q)$.
But $(Z'',Z)\notin\SE^A(P)$.
To see the latter, note that $F\subseteq Z$ must hold. 
So, if $((Z''\cap A),Z)$ were an $A$-SE-model of $P$, 
then it would also be an $A$-SE-model 
of $P\cup F$, contradicting the assumption that $Z\in\SM(P\cup F)$. 
Again we get an $A$-SE-model, $((Z''\cap A),Z)$, of exactly one of the 
programs, $Q$ in this case. Hence, according to ($ii$), there exists an 
$A$-SE-model $(M,Z)$ of $P$, $Z''\subset M\subset Z$. However, because of 
$F\subseteq Z$, it follows that $(M,Z)$ is also an $A$-SE-model of 
$P\cup F$, contradicting our assumption that $Z\in\SM(P\cup F)$.

This proves that, given ($i$) and ($ii$) for every $A$-SE-model $(X,Y)$ 
such that $(X,Y)$ is an $A$-SE-model of exactly one of  $P$ and $Q$, 
no sets of atoms $F\subseteq A$ 
and $Z$ exists such that $Z$ is 
an answer set of 
exactly one of $P\cup F$ and $Q\cup F$. That is,
$P\equivu^A Q$ holds.\hfill\qed

\subsection{Proof of Theorem~\ref{thm:subsetUE}}

For $(a)$, 
by Theorem~\ref{thm:RUE}, $P\equivu^A Q$ implies $\UE^A(P)=\UE^A(Q)$. 
Each $A$-UE-model of a program is, by definition, an $A$-SE-model of that
program. We immediately get $\UE^A(P)=\UE^A(Q)\subseteq \SE^A(Q)$ and
$\UE^A(Q)=\UE^A(P)\subseteq \SE^A(P)$.

For $(b)$, suppose $P\not\equivu^A Q$, and either $P$, $Q$, or $A$ is finite. 
By Theorem~\ref{thm:RUE} we have $\UE^A(P)\neq\UE(Q)$.  
Wlog, assume interpretations $X$, $Y$, such
that $(X,Y)\in\UEA(P)$ and $(X,Y)\not\in\UEA(Q)$. We have two cases:
If $(X,Y)\notin\SEA(Q)$, we are done, since then $\UE^A(P)\subseteq
\SE^A(Q)$ cannot hold.  If $(X,Y)\in\SEA(Q)$, this implies existence
of an $X'$ with $X\subset X'\subset Y$, such that
$(X',Y)\in\UEA(Q)$. However, since $(X,Y)\in\UEA(P)$, for each such
$X'$, $(X',Y)\notin\SEA(P)$.  Hence, $\UE^A(Q)\subseteq \SE^A(P)$
cannot hold.
\hfill\qed

\subsection{Proof of Theorem~\ref{theo:consequence}}
The complementary problem, $P \not\modelsu r$,
is in $\SigmaP{2}$ for general $P$ and in $\NP$ for head-cycle free
$P$, since a guess for a UE-model $(X,Y)$ of $P$ which violates $r$
can, by Theorem~\ref{theo-complex-3} be verified with a call to a
$\NP$-oracle resp.\ in polynomial time. In case of a positive $P$, by
Theorem~\ref{theo-C}, $P\modelsu r$ iff $P\models r$, which is in
$\coNP$ for general $P$ and polynomial for Horn $P$. 

The $\PiP{2}$-hardness part for ($i$)  is easily obtained from the
reduction proving the $\PiP{2}$-hardness part of Theorem~\ref{theo-complex-2}. For
the program $Q$ constructed there, it holds 
$Q\modelsu a\la $ if and only if none of the SE-models 
$\{ (\sigma_X(J), \sigma_X(J)\cup\at) \}$ of Q is an UE-model of $Q$ as well,
\iec if and only if $P \equivu Q$ holds, which is $\PiP{2}$-hard to
decide. 

The $\coNP$-hardness in case of ($ii$) follows easily from the
reduction which proves the $\coNP$-hardness part of
Theorem~\ref{theo-pos-new}:
the positive program $P$ constructed there
satisfies, by Theorem~\ref{theo-pos-new}, $P \modelsu a\la~$ 
if and only if $P \equivu Q$ holds, which is
equivalent to unsatisfiability of the CNF
$F$ there. Since $P$ is HCF we can, as in the proof of Theorem~\ref{theo-I}, 
again use $P^\ra$ and Theorem~\ref{theo-shift-equivu-1} in order 
to  show $\coNP$-hardness for head-cycle free (non-positive) programs.
\hfill\qed

\subsection{Proof of Theorem~\ref{thm:consneu}}
First consider $P$ is HCF.
Then, $\CONP$-membership of $P\modelsu Q$ is an immediate consequence
of the result in 
Theorem~\ref{theo:consequence} by testing $P\modelsu r$, for each $r\in Q$.
Since the class $\CONP$ is closed under conjunction,
$\CONP$-membership for $P\modelsu Q$ follows. 

Next, suppose $Q$ is HCF.
We first show the claim for normal $Q$,
using the complementary problem $P\not\modelsu Q$. 
By inspecting the characterizations of 
uniform equivalence,
$P\not\modelsu Q$ iff 
(i) $P\not\models Q$, 
or (ii) there exists an SE-model $(X,Y)$ of $P$, such that 
no $(X',Y)$ with $X\subseteq X' \subset Y$ is SE-model of $Q$. 
Test (i) 
is obviously in $\NP$. 
For containment in $\NP$ of Test~(ii), we argue
as follows:
We guess a pair $(X,Y)$ and check in polynomial time
whether it is SE-model of $P$. 
In order to check that
no $(X',Y)$ with $X\subseteq X' \subset Y$ is SE-model of $Q$
we test unsatisfiability of 
the program
$Q^Y\cup X \cup Y_\subset$,
which is Horn, whenever $Q$ is normal. 
Therefore, this test is 
is feasible in polynomial time.
Hence, $P\modelsu Q$ is in $\CONP$ for normal $Q$.
Recall that for a HCF program $Q$, we have $Q\equivu Q^\ra$. 
This implies that $P\not\modelsu Q^\ra$ iff $P\not\modelsu Q$.
Therefore, the claim holds for HCF programs as well.

We proceed with the matching lower bound.
Let $P$ and $Q$ as in the proof of Theorem~\ref{theo-pos-new}, 
then $P^\ra$ is normal, $Q$ is Horn, and $P\modelsu Q$ iff 
$P^\ra \modelsu Q$ iff $P^\ra \modelsu a\la~$, which is $\CONP$-hard.
\hfill\qed

\subsection{Proof of Theorem~\ref{thm:pospos}}
Membership is due to Theorem~\ref{thm:resmember}.

The hardness part is by a similar construction as above, 
\iec consider a QBF of the form
$\exists X\forall Y  \phi$ with 
$\phi=\bigvee_{i=1}^n D_i$ a DNF.
We take here the following programs, viz. 
\begin{eqnarray*}
P & = & \{ 
x\OR \bar{x}\la ;\; \la x,\bar{x}
\mid x\in X\} \cup\\
& & \{ y\OR\bar{y}\la;\; y\la a;\; \bar{y}\la a;\;
a \la y,\bar{y} 
\mid y\in Y\} \cup\\
& &  \{ a\la D^*_i \mid 1\leq i \leq n \}
\end{eqnarray*}
which is the 
same program as above, but without $\la\naf a$, and thus positive.
For the second program take
\begin{eqnarray*}
Q & = & \{ 
x\OR \bar{x}\la;\; \la x,\bar{x}
\mid x\in X\} \cup\\
& & \{ y\OR\bar{y}\la;\;  \la  y,\bar{y}
\mid y\in Y\} 
\cup\\
& &  \{  \la D^*_i \mid 1\leq i \leq n \} \cup\\
& &  \{  \la a \}.
\end{eqnarray*}
We start computing the SE-models of the two programs. 
Let, for any $J\subseteq X$, 
$$
M[J]\quad =\quad 
\sigma_X(J) \cup Y\cup \bar{Y}\cup \{a\},
$$
and suppose $A\subseteq X\cup \bar{X}$. 
The set of classical models of $P$ is given by 
$\{M[J]\mid J\subseteq X\}$ 
and $\sigma(J\cup I)$, for each $I\subseteq Y$, 
such that $\phi$ is false under $J\cup I$.
Thus, we get:
\begin{eqnarray*}
\SE(P) & = & 
\{ \big(\sigma(J\cup I),\sigma(J\cup I)\big),\;\; 
\big(\sigma(J\cup I),M[J]\big) \mid J\subseteq X, I\subseteq Y: 
J\cup I\not\models \phi 
\} \cup \\
& & \{ \big(M[J], M[J]\big) \mid J\subseteq X \}; \\
\SE(Q) & = &
\{ \big(\sigma(J\cup I),\sigma(J\cup I)\big) \mid J\subseteq X, I\subseteq Y: 
J\cup I\not\models \phi
\}.
\end{eqnarray*} 
First, 
each pair $(\sigma(J\cup I),\sigma(J\cup I))\in\SE(P)$ is $A$-SE-model of both, 
$P$ and $Q$. 
Second, $P$ possesses additional $A$-SE-models, if there 
exists at least one $J\subseteq X$ 
with $(M[J],M[J])\in\SE^A(P)$. 
This is the case, if no $I\subseteq Y$ 
makes $\phi$ false under $J\cup I$, \iec if the QBF
$\exists X\forall Y  \phi$ is true.
This shows $\SigmaP{2}$-hardness of 
deciding 
$P\not\equivs^A Q$ with $P$ and $Q$ positive. 
Consequently, $P\equivs^A Q$ under this setting is $\PiP{2}$-hard.
Note that since the argumentation holds also for $\card{A}<2$, we captured
both $\equivs^A$ and $\equivu^A$.

It remains to show $\PiP{2}$-hardness for $P\equive^A Q$, for the case
where $P$ is positive and $Q$ is normal, $e\in\{s,u\}$. 
As a consequence of Corollary~\ref{cor:seshift} (see also 
Example~\ref{exa:exclse}), 
for a disjunctive rule $r 
=v\OR w\la$, 
$Q\equivs^A Q^\ra_r$ holds for any $A$ ,whenever $\la v,w\in Q$.  
Hence, we can shift each disjunctive rule in $Q$ and get
$Q\equivs^A Q^\ra$. 
This shows $\PiP{2}$-hardness for $P\equivs^A Q$, for the case
where $P$ is positive and $Q$ is normal.
Again, we immediately get the respective result for $P\equivu^A Q$, 
since 
the argumentation holds also for $\card{A}<2$. 
\hfill\qed

\subsection{Proof of Theorem~\ref{thm:conpall}}
It remains to show $\CONP$-membership for two cases, viz.\
($i$)~
$P\equive^A Q$ with $P$ positive and $Q$ Horn; and
$P\equivs^A Q$ with $P$ HCF and $Q$ Horn. 
Therefore, we first show the following additional result:
\begin{lemma}\label{lemma:poscons}
For positive programs $P$, $Q$, and a set of atoms $A$,
$P\equive^A Q$ holds iff
(i) each $A$-minimal model of $P$ is a classical model of $Q$; and
(ii) for each interpretation $Y$, $Y\models Q$ implies existence of a $Y'\subseteq Y$ with $(Y'\cap A) = (Y\cap A)$, such that 
$Y'\models P$. 
\end{lemma}
\begin{proof}
For the only-if direction, first suppose (i) does not hold. 
It is easily seen, that then the $A$-minimal models cannot coincide, 
and thus $P\not\equive^A Q$. So
suppose (ii) does not hold; \iec there exists an interpretation $Y$, such 
that $Y\models Q$ but no $Y'\subseteq Y$ with $(Y'\cap A) = (Y\cap A)$ is a model of $P$. Again, the $A$-minimal models of $P$ and $Q$ cannot coincide.

For the if direction, 
suppose $P\not\equive^A Q$. First let $Y$ be
$A$-minimal for $P$ but not for $Q$. 
If $Y\not\models Q$ we are done, since (i) is violated. 
Otherwise, there exists a $Y'\subset Y$ with $(Y'\cap A)=(Y\cap A)$ 
which is a model of $Q$ but not a classical model of $P$; (ii) is violated.
Second, suppose there exists an $Y$ which is $A$-minimal 
for $Q$ but not for $P$.
If, each $Y'\subset Y$ with $(Y'\cap A)=(Y\cap A)$ does not model $P$, 
(ii) is violated. 
Otherwise, if there exists a $Y'\subset Y$ with $(Y'\cap A)=(Y\cap A)$ and $Y'\models P$, 
then there exists a $Y''$ with $(Y''\cap A)=(Y\cap A)$, which is 
$A$-minimal for $P$ but not a classical model of $Q$; whence (i) is violated.
\end{proof}
We proceed by proving (i) and (ii).
\medskip

\noindent
($i$): 
Since both $P$ and $Q$ are positive, 
$e=s$ and $e=u$
are the same concepts.
$\CONP$-membership is obtained
by 
applying Theorem~\ref{lemma:poscons}. 
In fact, this result suggests 
the following algorithm:
\begin{enumerate}
\item Check whether each $A$-minimal model 
of $Q$ is model of $P$;
\item Check whether, for each model $Y$ of $P$, there exists a $Y'\subseteq Y$ with $(Y'\cap A) = Y$ being model of $Q$.
\end{enumerate}
We show that, for both steps, the complementary problem is in $\NP$.
For Step~1, we guess a $Y$ and check whether it is $A$-minimal for $Q$ 
but not a classical model of $P$. The latter test is feasible in polynomial
time. The former reduces to test
unsatisfiability of the Horn theory 
$Q\cup (Y\cap A)\cup Y_\subset$. 
For the second step the argumentation is similar. 
Again, we guess an interpretation $Y$, check whether it is a model of $P$, 
and additionally, 
whether all $Y'\subseteq Y$ with $Y'\cap A = Y$ are not model of $Q$. 
The latter reduces to test
unsatisfiability 
of the Horn program
$Q\cup (Y\cap A) \cup Y_\subseteq$.
\medskip

\noindent
($ii$)~
In this setting, $\CONP$-membership is obtained
by the following algorithm:
\begin{enumerate}
\item Check whether the total $A$-SE-models of $P$ and $Q$ coincide;
\item Check whether, for each $X\subset Y$,
$(X,Y)\in\SE^A(P)$ implies $(X,Y)\in\SE^A(Q)$.
\end{enumerate}
The correctness of this procedure is a consequence of Proposition~\ref{prop:postotal}, 
\iec that $\SE^A(Q)\subseteq\SE^A(P)$ holds for positive $Q$, whenever the
total $A$-SE-models of $P$ and $Q$ coincide. Since $Q$ is Horn and thus 
positive, it is sufficient to 
check $\SE^A(P)\subseteq\SE^A(Q)$ which is accomplished by Step~2, indeed.
The first step is clearly in $\CONP$, since for total $A$-SE-interpretations,
$A$-SE-model checking and $A$-UE-model checking is the same task. 
By Theorem~\ref{thm:reltr}, $A$-UE-model checking 
is polynomial for HCF programs.
For the second step, we show $\NP$-membership for the complementary task.
We guess some $X'$ and $Y$, and test whether 
$(Y,Y)\in\SE^A(P)$, 
$X'\models P^Y$, and
$((X'\cap A),Y)\notin\SE^A(Q)$.
All tests are feasible in polynomial time and
imply that 
$(X,Y)\in\SE^A(P)$ but $(X,Y)\not\in\SE^A(Q)$, with $X=(X'\cap A)$.
\hfill\qed

\subsection{Proof of Theorem~\ref{thm:horn}}
Membership has already been obtained in Theorem~\ref{thm:conpall}.
For the hardness-part 
we reduce UNSAT 
to $P\equive^A Q$, where $P$ and $Q$ are Horn. 
The case of definite programs is discussed below.%
\footnote{
Our proof closely follows concepts used 
in~\cite{Cadoli94} to establish $\CONP$-hardness results for
closed world reasoning over Horn theories.}
Hence, let $F= \bigwedge_{i=1}^n c_{i,1}\vee\cdots\vee c_{i,n_i}$ be
given over atoms $V$ and  
consider 
$G=\{g_1\commadots g_n\}$ 
as new atoms. 
Define 
$$
P = \{ \la v,\bar{v} \mid v\in V \}  \cup \{ g_i \la c^*_{i,j} \mid 1\leq i \leq n; 1 \leq j\leq n_i\};
$$
and let $A=V\cup\bar{V}$.
Then, $F$ is unsatisfiable iff
$$
P \quad \equive^A \quad (P\cup \{  \la g_1\commadots g_n \})
$$
with 
$e\in\{s,u\}$.
We show the claim for $e=s$. Recall that RSE and RUE are the same for Horn programs. 
For the only-if direction suppose $F$ is unsatisfiable. Then, there does not
exist an interpretation $I\subseteq V$, such that $\sigma(I)\cup G$
is $A$-minimal for $P$. To wit, there exists at least a $G'\subseteq G$ such 
that $\sigma(I)\cup G'$ is model of $P$ as well. It is easily verified
that under these conditions, 
$P \equive^A P\cup \{   \la g_1\commadots g_n \}$ holds.
On the other hand, if $F$ is satisfiable, there exists an interpretation
$I\subseteq V$ such that $\sigma(I)\cup G$ is an $A$-minimal model of $P$.
However, $\sigma(I)\cup G$ is not a model of $P\cup \{ \la g_1\commadots g_n \}$. 
This proves the claim.

We show that $\CONP$-hardness holds also for definite programs.
Therefore, we introduce further atoms $a,b$ and change $P$ 
to 
$$
P = \{ a \la v,\bar{v} \mid v\in V \}  \cup \{ g_i \la c^*_{i,j} \mid 1\leq i \leq n; 1 \leq j\leq n_i\} \cup \{ u \la a \mid u \in \at\}
$$
where $\at= \{b\} \cup V\cup \bar{V}\cup G$.
Then,
$F$ is unsatisfiable iff
$$
P \equive^A P\cup \{ b \la g_1\commadots g_n \}
$$
with $A=\{a,b\}\cup V\cup \bar{V}$.
The correctness of the claim is by analogous arguments as above.
\hfill\qed
\end{document}